\documentclass{article}

\usepackage{PRIMEarxiv}

\usepackage[utf8]{inputenc} % allow utf-8 input
\usepackage[T1]{fontenc}    % use 8-bit T1 fonts
\usepackage[hyperfootnotes=false]{hyperref}
\usepackage{url}            % simple URL typesetting
\usepackage{booktabs}       % professional-quality tables
\usepackage{amsfonts}       % blackboard math symbols
\usepackage{nicefrac}       % compact symbols for 1/2, etc.
\usepackage{microtype}      % microtypography
\usepackage{lipsum}
\usepackage{fancyhdr}       % header
\usepackage{graphicx}       % graphics
\usepackage{subcaption}

\graphicspath{{media/}}     % organize your images and other figures under media/ folder
\usepackage{comment}
\usepackage{amsmath}
\usepackage{MnSymbol}
\usepackage{appendix}
\usepackage{multirow}
\usepackage{booktabs}
\usepackage{tablefootnote} 
\newcommand\blfootnote[1]{%
  \begingroup
  \renewcommand\thefootnote{}\footnote{#1}%
  \addtocounter{footnote}{-1}%
  \endgroup
}

\newcommand{\x}{\boldsymbol{x}}
\newcommand{\z}{\boldsymbol{z}}
\newcommand{\param}{\boldsymbol{\theta}}
\newcommand{\phiparam}{\boldsymbol{\phi}}
\newcommand{\qphi}{q_{\phiparam}}

\newcommand{\vv}{\boldsymbol{v}}
\newcommand{\ppdf}{p_{\boldsymbol{\theta}}}

\usepackage{mathtools}
\DeclarePairedDelimiter\floor{\lfloor}{\rfloor}

\DeclareMathOperator*{\argmax}{arg\,max}
 
\title{Physics-informed Variational Autoencoders \\ for Improved Robustness to Environmental \\ Factors of Variation}

\author{
 Romain Thoreau \\
  \texttt{romain.thoreau@cnes.fr} \\
   \And
 Laurent Risser \\
  \texttt{lrisser@math.univ-toulouse.fr} \\
  \And
 Véronique Achard \\
  \texttt{veronique.achard@onera.fr} \\
  \And
  Béatrice Berthelot \\
  \texttt{beatrice.berthelot@magellium.fr} \\
  \And 
  Xavier Briottet \\
  \texttt{xavier.briottet@onera.fr}
}

\begin{document}
\maketitle
\begin{abstract}
The combination of machine learning models with physical models is a recent research path to learn robust data representations. 
In this paper, we introduce p$^3$VAE, a variational autoencoder that integrates prior physical knowledge about the latent factors of variation that are related to the data acquisition conditions. 
p$^3$VAE combines standard neural network layers with non-trainable physics layers in order to partially ground the latent space to physical variables.
We introduce a semi-supervised learning algorithm that strikes a balance between the machine learning part and the physics part.
Experiments on simulated and real data sets demonstrate the benefits of our framework against competing physics-informed and conventional machine learning models, in terms of extrapolation capabilities and interpretability. In particular, we show that p$^3$VAE naturally has interesting disentanglement capabilities. Our code and data have been made publicly available at \href{https://github.com/Romain3Ch216/p3VAE}{https://github.com/Romain3Ch216/p3VAE}.
\end{abstract}
\blfootnote{This paper is submitted to Springer Machine Learning.}

% keywords can be removed

\section{Introduction}\label{sec:intro}

Physics-informed machine learning, that is the combination of data-driven and theory-driven modeling, has recently raised a lot of attention. 
The integration of physical models in machine learning has indeed demonstrated promising properties such as improved interpolation and extrapolation capabilities, and increased interpretability \cite{raissi2019physics,  takeishi2021physics, yin2021augmenting, zerah2024physics}. 
%Machine learning models rely on assumptions, sometimes called inductive biases \cite{mitchell1980need, zhao2018bias}, made over the model architecture (\textit{e.g.} convolutional layers), the objective function (\textit{e.g.} L2 regularization), or the learning algorithm (\textit{e.g.} stochastic gradient descent).
Machine learning models rely on assumptions, sometimes called inductive biases \cite{mitchell1980need, zhao2018bias}, made over the model architecture, the objective function, or the learning algorithm, that are crucial for generalization.
Physics-informed machine learning models, in addition, integrate inductive biases derived from physical prior knowledge, which by their nature generalize to out-of-distribution data. 
Therefore, in various fields for which the data distribution is governed by physical laws, such as fluid dynamics, thermodynamics or solid mechanics, physics-informed machine learning has recently become a hot topic \cite{jacobsen2021disentangling, wei2020thermodynamic, trask2022unsupervised}.

\noindent
In the present paper, we introduce p$^3$VAE, a \textbf{p}hysics-informed variational autoencoder that integrates \textbf{p}artial \textbf{p}rior knowledge about the data generative process into representation learning. p$^3$VAE aims to decouple the variation factors that are intrinsic to the data from environmental factors related to acquisition conditions. On the hypothesis that acquisition conditions induce complex and non-linear data variations, we argue that a conventional generative model would hardly decorrelate intrinsic factors from environmental factors. Therefore, we introduce prior physical knowledge in the decoding part of a semi-supervised variational autoencoder. Variational AutoEncoders (VAEs) \cite{kingma2014auto} are powerful latent variable probabilistic models that have an autoencoder framework. The key to their success lies in their stochastic variational inference and learning algorithm that allows to leverage very large unlabeled data sets and to model complex posterior data distributions given latent variables \cite{kingma2014auto}. Besides, recent advances offer more and more control on the latent space distributions \cite{higgins2016early, higgins2016beta, figurnov2018implicit, joo2020dirichlet}. VAEs were used, for instance, in text modeling to capture topic information as a Dirichlet latent variable \cite{xiao2018dirichlet} or in image modeling where latent variables were both discrete and continuous \cite{dupont2018learning}. 

\noindent
Integrating physics in an autoencoder was first proposed by \cite{aragon2020self}. They used a fully physical model inplace the decoder to inverse a 2D exponential light galaxy profile model. In the same spirit, \cite{zerah2024physics} substituted the decoder of a VAE by a radiative transfer model. In \cite{takeishi2021physics}, Takeishi and Kalousis generalized the work of \cite{aragon2020self} by developing a mathematical formalism introduced as physics-integrated VAEs ($\phi$-VAE). $\phi$-VAEs  complement an incomplete physical model with a machine learning model in the decoder of a VAE. $\phi$-VAEs can be seen as a "VAE variant" of \cite{yin2021augmenting} that complements incomplete physical dynamics described by an ordinary differential equation with a neural network. To have a consistent use of physics despite the high representational power of machine learning models, $\phi$-VAEs employ  a regularization strategy that is central to their contribution. Their regularization heavily relies on the capacity of the physics model to reconstruct, by itself, the data from a subset of the latent space. In contrast, our model, p$^3$VAE, tackles problems for which prior physical knowledge only describes how environmental latent factors of variation alter the data. To that extent, physics is integrated in a different way into the decoder. In order to strike a balance between the physics part and the machine learning part, we introduce a new semi-supervised training algorithm that grounds a subset of the latent space to physical variables. 
The main contributions of the paper are as follows:

\begin{itemize}
\setlength{\itemindent}{.2in}
	\item[\textemdash] We introduce p$^3$VAE, a variational autoencoder that combines conventional neural network layers with physics-based layers for improved robustness to environmental factors of variation,
    \item[\textemdash] We enhance the semi-supervised training procedure introduced in \cite{kingma2014semi} to balance between physics and trainable parameters,
    \item[\textemdash] We introduce an inference scheme to fully leverage the physics part of p$^3$VAE, 
    \item[\textemdash] We assess the benefits of our model on several synthetic and real-world data sets, compared to physics-informed and standard machine learning models.
\end{itemize}

\noindent
Our paper is organized as follows. Section \ref{sec2} discusses the related work. Section \ref{sec3} presents our model in details. Section \ref{sec5} describes the numerical experiments. Section \ref{sec7} opens a discussion and brings conclusions to the paper. 

\section{Related work}\label{sec2}

Our method makes use, at the same time, of physical priors and of a semi-supervised learning technique, which is intrinsically linked to the architecture of p$^3$VAE. Therefore, we review related work on physics-informed machine learning and semi-supervised learning.

\subsection{Physics-informed machine learning}\label{subsec22}

Physics-informed machine learning (ML) can be divided into physics-based losses \cite{chen2020physics, wang2021understanding, wei2020thermodynamic, raissi2019physics} and physics-based models, in which our method fits. Physics-based models embed physical layers in their design. \cite{yildiz2019ode2vae} and \cite{linial2021generative} introduced VAEs which latent variables were grounded to physical quantities (such as position and velocity) and governed by Ordinary Differential Equations (ODEs). More related to our work, \cite{aragon2020self} and \cite{zerah2024physics} inverse fully physical models using the frameworks of autoencoders and variational autoencoders, respectively, as discussed in section \ref{sec:intro}. They demonstrated that decoding the latent space with physical models enforced disentanglement. Both methods can be seen as particular instances of physics-integrated VAEs ($\phi$-VAE) introduced by \cite{takeishi2021physics}. $\phi$-VAE  is the most closely related methodology to p$^3$VAE. Differences between our work and $\phi$-VAEs are discussed in details in section \ref{sec:comparison}. 

\noindent
Close but outside the scope of physics-informed ML are methods that disentangle the latent space to have semantically meaningful latent variables. 
For instance, \cite{kulkarni2015deep} introduced a stochastic gradient variational Bayes training procedure to encourage latent variables of a VAE to fit physical properties (rotations and lighting conditions of 3D rendering of objects). 
Other state-of-the-art disentanglement techniques such as \cite{ding2020guided, rodriguez2021disentanglement, chen2018isolating} can yield interpretable latent spaces, for instance on the MNIST data set \cite{lecun-mnisthandwrittendigit-2010}, where latent variables represent the rotations or the thickness of handwritten digits.
The seminal work of \cite{locatello2019challenging} on the unsupervised learning of disentangled representations offers a principled perspective on physics-informed ML.
Theoretical results in \cite{locatello2019challenging} demonstrate that unsupervised disentanglement learning is fundamentally impossible for arbitrary generative models
without inductive biases both on the models and on the data. 
This result motivates the integration of prior physical knowledge into generative models. 
In our work, we combine physical prior knowledge with the partial supervision of latent generative factors.

\subsection{Semi-supervised learning}\label{subsec21}

Semi-supervised techniques introduce inductive biases into learning by exploiting the structure of unlabeled data, in addition to scarce labeled data, in order to learn data manifolds. Van Engelen and Hoos provide an exhaustive survey on transductive and inductive semi-supervised methods in \cite{van2020survey}. As far as transductive methods optimize over the predictions themselves, we only focus on inductive methods that optimize over predictive models, to which we can integrate some prior knowledge. Common inductive techniques include pseudo-labeling, or self-labeled, approaches \cite{triguero2015self} (where the labeled data set is iteratively enlarged by the predictions of the model), unsupervised preprocessing (such as feature extraction with contractive autoencoders \cite{salah2011contractive} or pretraining \cite{erhan2010does}) and regularization techniques on the unlabeled data. Regularization includes (along side a classification loss) additional reconstruction losses on embedding spaces \cite{ranzato2008semi, weston2012deep}, manifold regularization \cite{rifai2011higher}, perturbation-based approaches such as virtual adversarial training \cite{miyato2018virtual} that decreases the predictions sensitivity to inputs noise or, in the context of semantic segmentation, relaxed k-means \cite{castillo2021semi}. Finally, generative models are, by nature, relevant to handle unlabeled data. State-of-the-art generative models such as Generative Antagonist Nets (GANs) \cite{goodfellow2014generative}, VAEs \cite{kingma2014auto} and normalizing flows \cite{rezende2015variational} were enhanced to handle both labeled and unlabeled data. For instance, \cite{spurr2017guiding} introduced a semi-supervised version of InfoGAN \cite{chen2016infogan}; \cite{kingma2014semi} and \cite{izmailov2020semi} developed a methodology to train VAEs and normalizing flows in semi-supervised settings, respectively.

\section{p$^3$VAE framework}\label{sec3}

\begin{figure*}[t]
    \centering
    \subfloat{\includegraphics[height=0.24\textwidth]{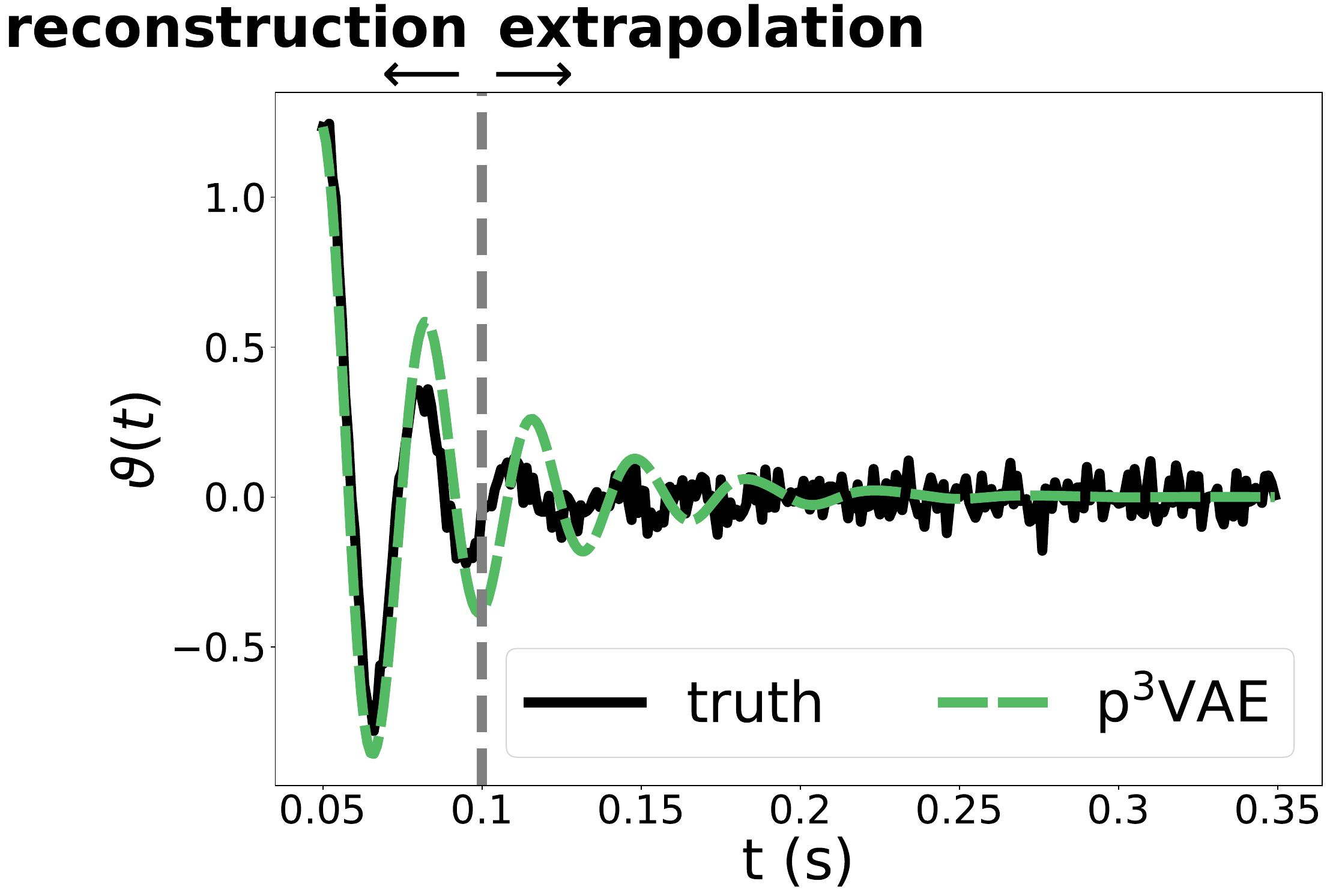}}
    \subfloat{\includegraphics[height=0.215\textwidth]{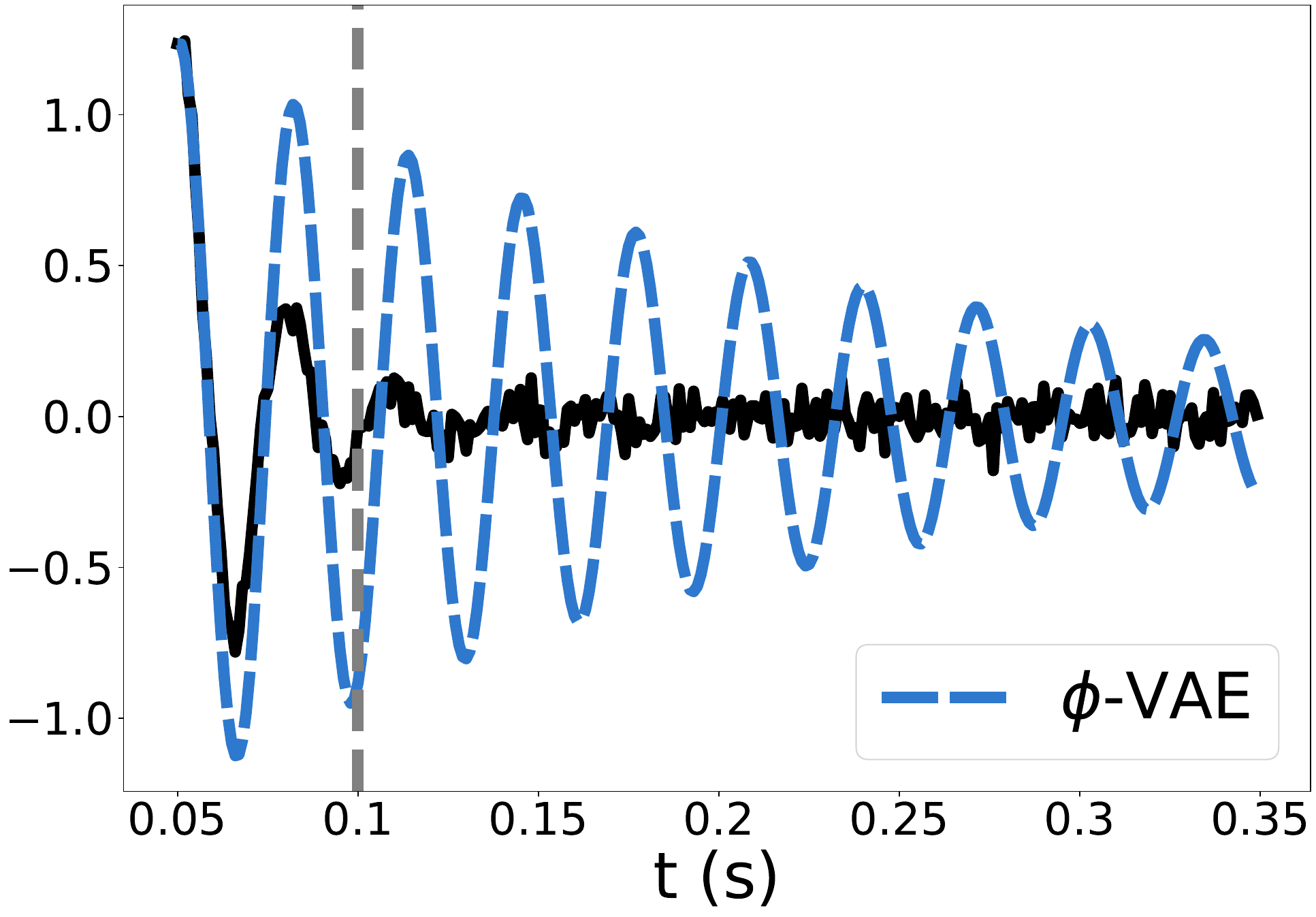}}
    \subfloat{\includegraphics[height=0.215\textwidth]{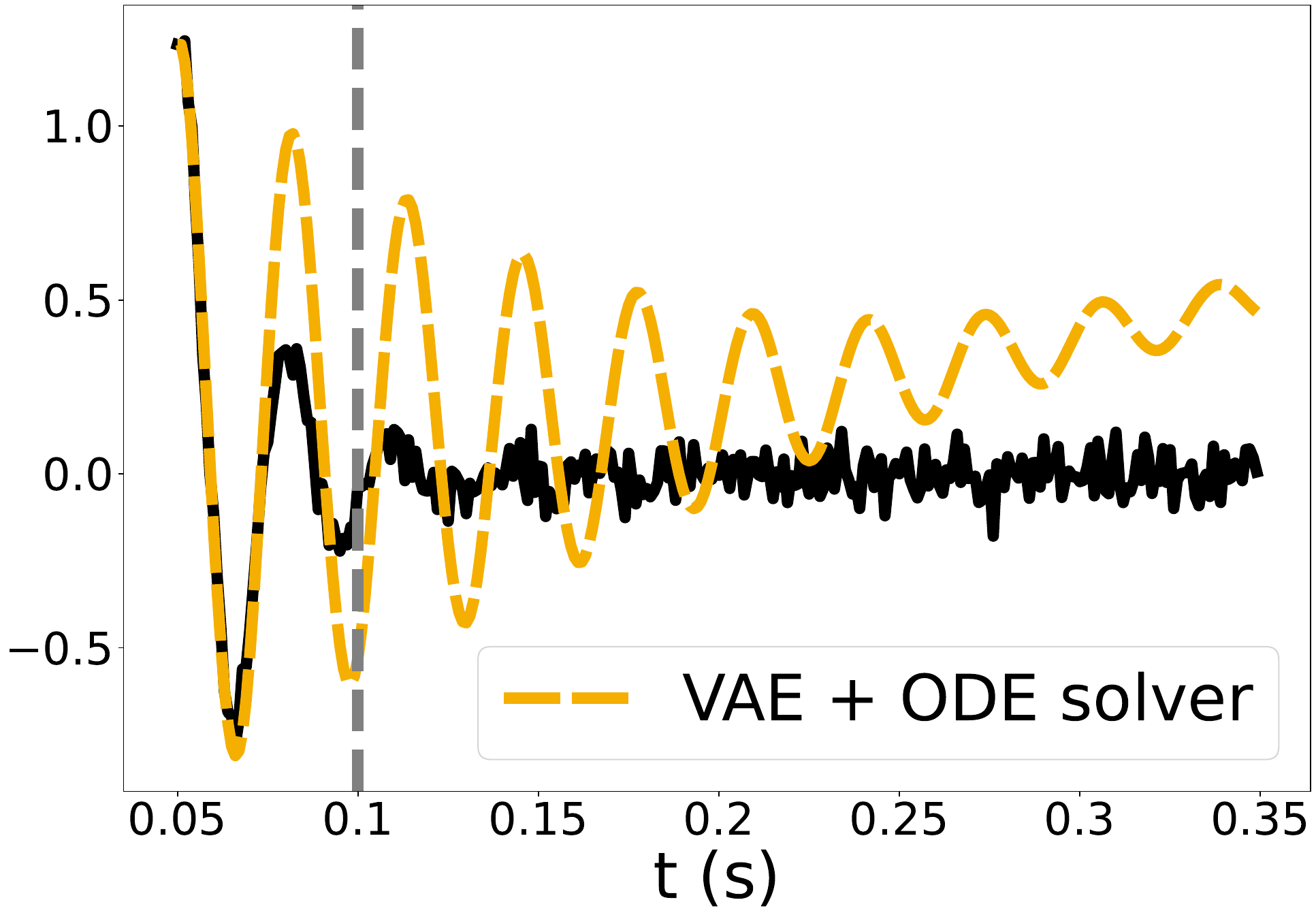}}
    \caption{Predicted dynamics of p$^3$VAE (ours), $\phi$-VAE \cite{takeishi2021physics} and VAE + ODE solver \cite{yildiz2019ode2vae, tothhamiltonian} for the damped pendulum vs. ground truth trajectory $\frac{d^2\vartheta}{dt^2}(t) + \xi \frac{d\vartheta}{dt}(t) + \omega^2 \mbox{ sin } \vartheta(t) = 0$ with noise. \label{fig:pendulum}}
\end{figure*}

\subsection{Motivating example \label{sec:example}}

In order to illustrate the main concepts of p$^3$VAE, we start with the damped pendulum example. The pendulum example is a common problem in physics-informed machine learning \cite{takeishi2021physics, yin2021augmenting}. Let us consider the trajectories of damped pendulums (cf. Fig \ref{fig:pendulum}), whose dynamics are governed by the following ordinary differential equation (ODE):
\begin{align}
	\frac{d^2\vartheta}{dt^2}(t) + \xi \frac{d\vartheta}{dt}(t) + \omega^2 \mbox{ sin } \vartheta(t) = 0 \label{eq:pendulum_ode}
\end{align}
where $\vartheta$ is a pendulum's angle, $\omega$ is the pendulum's angular frequency, and $\xi$ is the fluid damping coefficient. 
The pendulum's angular frequency and the fluid damping coefficient are the latent generative factors of variation. 

\noindent
Given a training data set comprising pendulum trajectories, we aim to infer the dynamics of the pendulum and the latent generative factors, with the help of prior physical knowledge. 
In this work, we assume that prior partial physical knowledge models the relation between the data and the the fluid damping coefficient $\xi$. While $\omega$ is intrinsic to the pendulum, we assume that the fluid damping coefficient $\xi$ only depends on the environment in which the data is observed, \textit{i.e.} what we call the acquisition conditions. Precisely, this partial prior knowledge enables to write a solution of the ODE as follows, for all $t \in \mathbb{R}_+$:
\begin{align}
	\vartheta(t) = f_E(\mathcal{F}_{\mbox{\tiny{ODE}}}[\ddot{\vartheta} + f_I(\vartheta; \omega) = 0] ; \xi)
\end{align}
where $\mathcal{F}_{\mbox{\tiny{ODE}}}$ denotes an ODE solver with regard to $\vartheta$, $f_E$ is a known function derived from prior physical knowledge, and $f_I$ describes the unknown intrinsic dynamics of the pendulum, that have to be learned by a neural network. The intrinsic dynamics described by $f_I$ depend on the intrinsic factor $\omega$ only, and not on the environmental factor $\xi$. 
In contrast, $f_E$  describes how the fluid damping coefficient $\xi$ induces an exponential decrease of the pendulum's angle, which is independent of the pendulum's angular frequency $\omega$. The analytical expression of $f_E$ is given in section \ref{sec:pend_exp}. 

\noindent
p$^3$VAE is a framework, based on the variational autoencoder \cite{kingma2014auto}, to concomitantly approximate the unknown dynamics $f_I$ and the latent factors of variation $\omega$ and $\xi$.
It takes as input sequences $\x^{(i)} = [\vartheta^{(i)}(0) \: \vartheta^{(i)}(\Delta t) \: \ldots \: \vartheta^{(i)}((\tau - 1) \Delta t)]^T$ of pendulums' angles, and encodes them into low-dimensional latent codes.
The latent space of p$^3$VAE is split in two parts: intrinsic latent variables $z_I$, meant to encode $\omega$, and environmental latent variables $\z_E$, meant to encode $\xi$. 
$\z_I$ and $\z_E$ are linked to $f_I^{\param}$ and $f_E$, respectively.
In order to ground $\z_E$ to $\xi$, we integrate the prior physical knowledge, $f_E$, into the decoder, in combination with a standard neural network  $f_I^{\param}$.
On the other hand, $z_I$ is grounded to $\omega$ with a partial supervision of the $\omega^{(i)}$s associated to the training data points $\x^{(i)}$s.
In summary, the decoder here is defined as follows: 
\begin{align}
	\mathbb{E}[\x] = f_E(\mathcal{F}_{\mbox{\tiny{ODE}}}[\ddot{\vartheta} + f_I^{\param}(\vartheta; \z_I) = 0] ; \z_E)
\end{align}
The encoder and the machine learning part of the decoder are jointly trained with a semi-supervised algorithm, introduced in section \ref{sec:general}.

\begin{figure*}[t]
    \centering
    \subfloat[Supervised with $\z_I^\ast$]{\includegraphics[height=0.09\textwidth]{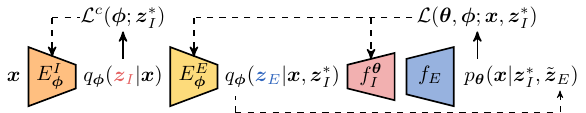}}
    \hfill
    \subfloat[Unsupervised]{\includegraphics[height=0.09\textwidth]{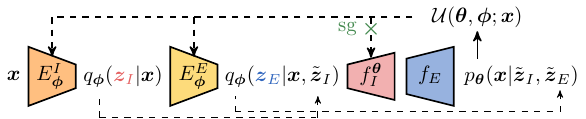}}

    \caption{Illustration of p$^3$VAE at training. (a) Supervised part: p$^3$VAE is supervised by the latent intrinsic factors, denoted as $\z_I^\ast$. The lower bound of the log-likelihood, $\mathcal{L}(\param, \boldsymbol{\phi}; \x, \z_I^\ast)$, is maximized while the cross-entropy (resp. mean squared error) is minimized, for discrete (resp. continuous) latent variables.  (b) Unlabeled part: The lower bound of the marginal log-likelihood, $\mathcal{U}(\param, \boldsymbol{\phi}; \x)$, is maximized with respect to $\boldsymbol{\phi}$ only, via a stop-gradient operator. Thus, the unlabeled step decreases the tightness of the bound, \textit{i.e.} the gap between the lower bound $\mathcal{U}(\param, \boldsymbol{\phi}; \x)$ and the marginal log-likelihood $\mbox{log } \ppdf(\x)$.\label{fig:p3vae_training}}
\end{figure*}

\subsection{General formulation \label{sec:general}}

We now present the framework of p$^3$VAE in general. p$^3$VAE is based on four main elements:
\begin{itemize}
\setlength{\itemindent}{.2in}
	\item[\textemdash] The integration of prior physical knowledge encoded into $f_E$, that describes how environmental factors related to acquisition conditions alter the data,
	\item[\textemdash] The partial supervision of the intrinsic factors of variation, encoded into the latent variables $\z_I$,
	\item[\textemdash] A balance between the physics-based model $f_E$ and the neural network $f_I^{\param}$ through a stop-gradient operator,
	 \item[\textemdash] The estimation of the true posterior distribution at inference through importance sampling.
\end{itemize}

\subsubsection{p$^3$VAE architecture \label{sec:arch_p3VAE}}

We assume that a data point $\x \in \mathcal{X}$ is generated by a random process that involves intrinsic and environmental latent variables $\z_I \in \mathcal{Z}_I$ and $\z_E \in \mathcal{Z}_E$, respectively. Part of the intrinsic latent variables may sometimes be observed, and sometimes unobserved. We assume that the environmental latent variables $\z_E$ are grounded to some physical properties related to the acquisition conditions of the data. $\z_E$ could be, for instance, the fluid damping coefficient mentioned in section \ref{sec:example}. The latent variables can be continuous, discrete, or both. $\mathcal{X}$, $\mathcal{Z}_I$ and $\mathcal{Z}_E$ are subsets of the euclidean space. 

The generative process consists of two steps: (1) values of $\z_I$ and $\z_E$ are generated from a prior distribution $p(\z_I, \z_E)$; (2) a value $\x$ is generated from the conditional distribution $\ppdf(\x \vert \z_I, \z_E)$ parameterized by $\param$. We assume that latent variables $\z_I$ and $\z_E$ are independent factors of variations, \textit{i.e.} that they are mutually independent:
\begin{equation}
    p(\z_I, \z_E) := p(\z_I) p(\z_E)
\end{equation}
Besides, we assume a Gaussian observation noise $\Sigma$ and define the likelihood as follows: 
\begin{equation}
    \ppdf(\x \vert \z_I, \z_E) := \mathcal{N}\big(\x \vert f_E(\mathcal{F}[f_I^{\param}, \z_I], \z_E), \Sigma \big)
\end{equation}
where $f_I^{\param}: \mathcal{Z}_I \longrightarrow \mathcal{Z}$ is a neural network and $f_E: \mathcal{Z} \times \mathcal{Z}_E \longrightarrow \mathcal{X}$ is a physics model differentiable with regard to its inputs. $\mathcal{F}$ is a functional that either solves an ODE, specified by $f_I^{\param}$ and $\z_I$, with a differentiable ODE solver, or simply evaluates $f_I^{\param}$. 

\noindent
We would like to maximize the marginal likelihood $\ppdf(\x, \z_I)$ when $\z_I$ is observed and $\ppdf(\x)$ otherwise, which are unfortunately intractable. Therefore, we follow the state-of-the-art variational optimization technique introduced in \cite{kingma2014auto} by approximating the true posterior $\ppdf(\z_I, \z_E \vert \x)$ by a recognition model $\qphi(\z_I, \z_E \vert \x)$, computed by neural networks. 
Precisely, one neural network computes $\qphi(\z_I \vert \x)$, and another one computes $\qphi(\z_E \vert \x, \z_I)$.
The encoding part $\qphi(\z_I, \z_E \vert \x)$ and the decoding part $\ppdf(\x \vert \z_I, \z_E)$ form a variational autoencoder (with parameters $\param$ and $\boldsymbol{\phi}$). 

\subsubsection{p$^3$VAE semi-supervised training}\label{sec:optimization}

We now explain how we enhanced the semi-supervised optimization algorithm introduced by \cite{kingma2014semi}  to the training of our model. Fig. \ref{fig:p3vae_training} illustrates how p$^3$VAE is trained. 

\noindent
\textbf{Semi-supervised Model Objective}. The objective function derived by \cite{kingma2014semi} for an observation $x$ is twofold. First (with our notations), when the latent variables $\z_I$ are observed, the evidence lower bound of the log-likelihood $\mbox{log} \: \ppdf(\x, \z_I)$ is:
\begin{align}
\begin{split}\label{eq:supervised_objective}
    \mathcal{L}(\x, \z_I) & = \mathbb{E}_{\qphi(\z_E \vert \x, \z_I)}[\mbox{log} \: \ppdf(\x \vert \z_I, \z_E) + \mbox{log} \: p(\z_I) + \mbox{log} \: p(\z_E) - \mbox{log} \: \qphi(\z_E \vert \x, \z_I)] \leq \mbox{log} \: \ppdf(\x, \z_I) 
\end{split}
\end{align}
Second, when $\z_I$ is not observed, the evidence lower bound of the marginal log-likelihood $\mbox{log} \: \ppdf(\x)$ is:
\begin{align}
\begin{split}\label{eq:unsupervised_objective}
    \mathcal{U}(\x) & = \mathbb{E}_{\qphi(\z_I, \z_E \vert \x)}[\mbox{log} \: \ppdf(\x \vert \z_I, \z_E) + \mbox{log} \: p(\z_I) + \mbox{log} \: p(z_E) - \mbox{log} \: \qphi(\z_I, \z_E \vert \x)] \\
    & = \mathbb{E}_{\z_I \sim \qphi(\z_I \vert \x)} \big(\mathcal{L}(\x, \z_I)\big) + H\big(\qphi(\z_I \vert \x)\big) \leq \mbox{log} \: \ppdf(\x) 
\end{split}
\end{align}
where $H\big(\qphi(\z_I \vert \x)\big)$ denotes the entropy of $\z_I \vert \x$. Besides, we can note that the term $\mathbb{E}_{\qphi(\z_E \vert \x, \z_I)}[\mbox{log} \: p(\z_E) - \mbox{log} \: \qphi(\z_E \vert \x, \z_I)]$ in equation (\ref{eq:supervised_objective}) is the negative Kullback-Leibler divergence between $\qphi(\z_E \vert \x, \z_I)$ and $p(z_E)$. \\

\noindent
The predictive distribution $\qphi(\z_I \vert \x)$ only contributes in the second objective function (\ref{eq:unsupervised_objective}). To remedy this, \cite{kingma2014semi} adds a supervised loss $\mathcal{L}^c(\phiparam; \z_I)$ to the total objective function, weighted by a coefficient $\lambda > 0$, so that $\qphi(\z_I \vert \x)$ also learns from labeled data. 

\noindent 
\textbf{Stop-gradient operator}. The main contribution to the semi-supervised optimization scheme is what we refer as the stop-gradient operator. The neural network $f_I^{\param}$ generates data features from the latent intrinsic variables $\z_I$. Because $f_I^{\param}$ has a high representational power, \textit{i.e.} high flexibility, some inconsistent values of $(\z_I, \z_E)$ (\textit{e.g.} physically unrealistic values) can lead to a good reconstruction of the training data. In order to mitigate this issue, we do not back-propagate the gradients with regard to $f_I^{\param}$ parameters when $\z_I$ is not observed, via a stop-gradient operator. Another perspective on the stop-gradient operator is to note that it makes the optimization of p$^3$VAE a two-step algorithm in an expectation-maximization fashion:
\begin{itemize}
\setlength{\itemindent}{.2in}
	\item[\textemdash] Supervised part = step 1:% $\approx$ EM steps: 
	\begin{align*}
		\max_{\param, \boldsymbol{\phi}} \:\: \mathcal{L}(\param, \boldsymbol{\phi}; \x, \z_I) - \lambda \mathcal{L}^c(\phiparam; \z_I)
	\end{align*}
	\item[\textemdash] Unsupervised part = step 2: % $\approx$ E step: 
	\begin{align*}
		\max_{\boldsymbol{\phi}} \:\: \mathcal{U}(\boldsymbol{\phi}; \param, \x)
	\end{align*}
\end{itemize}
The second step leaves $\param$ unchanged, and only decreases the tightness of the bound, which is the gap between the lower bound $\mathcal{U}(\param, \boldsymbol{\phi}; \x)$ and the marginal log-likelihood $\mbox{log } \ppdf(\x)$, that does not depend on $\boldsymbol{\phi}$.

%===============================================================================================
\subsubsection{Inference}\label{sec:inference}

At inference, \cite{kingma2014semi} uses the approximate predictive distribution $\qphi(\z_I \vert \x)$ to make predictions. However, for discrete variables, we can compute $\argmax_{\z_I} \ppdf(\z_I \vert \x)$ although the true predictive distribution $\ppdf(\z_I \vert \x)$ is intractable. As a matter of fact, assuming that $p(\z_I)$ is uniform, we have from Bayes rule that:
\begin{equation}
    \ppdf(\z_I \vert \x) = \frac{\ppdf(\x \vert \z_I)p(\z_I)}{\ppdf(\x)} \propto \ppdf(\x \vert \z_I)
\end{equation}
Moreover, as $\z_E$ is independent from $\z_I$, we can write that:
\begin{align}
    \ppdf(\x \vert \z_I) & = \int \ppdf(\x, \z_E \vert \z_I) \: d\z_E = \int \ppdf(\x \vert \z_I, \z_E)p(\z_E) \: d\z_E = \mathbb{E}_{\z_E \sim p(\z_E)} \ppdf(\x \vert \z_I, \z_E)
\end{align}
Thus, we can perform Monte Carlo sampling to estimate $\ppdf(\x \vert \z_I)$. In order to decrease the variance of the estimation, we can sample $\z_E$ from $\qphi(\z_E \vert \x)$, performing importance sampling as follows:
\begin{align}
    \ppdf(\x \vert \z_I) & = \mathbb{E}_{\qphi(\z_E \vert \x)} \: w \: \ppdf(\x \vert \z_I, \z_E) = \mathbb{E}_{\tilde{\z}_I \sim \qphi(\z_I \vert \x)} \mathbb{E}_{\qphi(\z_E \vert \x, \tilde{\z}_I)} \: w \:  \ppdf(\x \vert \z_I, \z_E)
\end{align}
where $w = \frac{p(\z_E)}{\qphi(\z_E \vert \x)}$ is the likelihood ratio. To our knowledge, this derivation of $\argmax_{\z_I} \ppdf(\z_I \vert \x)$ has not yet been used in the context of semi-supervised VAEs, while it is close to the importance sampling technique introduced in \cite{rezende2014stochastic} to estimate the marginal likelihood of VAEs. The major advantage of our strategy is to explicitly use the physics model at inference, which has natural extrapolation capacities. Besides, this derivation offers a convenient way to estimate an uncertainty heuristic over the inferred latent variables. We can easily compute the empirical standard deviation of the $\z_E^i$s where $\z_E^i \sim \qphi(\z_E \vert \x, \tilde{\z}_I)$ and $\tilde{\z}_I \sim \qphi(\z_I \vert \x)$.

\subsection{Differences with $\phi$-VAE \cite{takeishi2021physics} \label{sec:comparison}}

In this section, we emphasize the fundamental differences between our framework, p$^3$VAE, and "physics-integrated VAEs" \cite{takeishi2021physics}, denoted as $\phi$-VAEs. These differences stem from the different natures of the problems we are tackling, and reflect into:
\begin{itemize}
	\setlength{\itemindent}{.2in}
	\item[\textemdash] the formulation of the decoder,
	\item[\textemdash] the training algorithm.
\end{itemize}

\noindent
\textbf{p$^{\boldsymbol{3}}$VAE decoder VS $\boldsymbol{\phi}$-VAE decoder} $\phi$-VAE tackles problems for which a physics model, denoted as $f_P$, provides a \textit{quite accurate} though \textit{imperfect} reconstruction of the data from latent variables $\z_P$. In order to complete the physics model, \cite{takeishi2021physics} augment the decoder with a neural network, denoted as $f_A$, and auxiliary latent variables $\z_A$. Depending on the problem, $f_A$ and $f_P$ can be either be composited additively or by composition (in one way only):
\begin{align}
	\mathbb{E}[\x]^{\phi\mbox{-VAE}} = \begin{cases} \mathcal{F}[f_A(f_P(\z_P), \z_A)] \\ 
	\mathcal{F}[f_A(\z_A) + f_P(\z_P)] \end{cases}
\end{align}
where $\mathcal{F}[f]$ is a functional that either solves an ODE described by $f$, or evaluates $f$. $\phi$-VAE cannot compose $f_P$ and $f_A$ in the other way (\textit{i.e.} $f_P(f_A(\z_A), \z_P)$). In contrast, p$^3$VAE tackles problems for which a physics model $f_E$ (equivalent to $f_P$) can \textit{not} provide a useful reconstruction of the data from the environmental latent variables $\z_E$. Instead, $f_E$ can only adjust the prediction of a neural network $f_I^{\param}$ (equivalent to $f_A$), through the following combination:
\begin{align}
	\mathbb{E}[\x]^{\mbox{p}^3\mbox{VAE}} = \mathcal{F}[f_E(f_I^{\param}(\z_I), \z_E)]
	 & \big(= \mathcal{F}[f_P(f_A(\z_A), \z_P)]
	 \neq \mathcal{F}[f_A(f_P(\z_P), \z_A)] 
	 = \mathbb{E}[\x]^{\phi\mbox{-VAE}} \\
	&  \mbox{ with the notations of $\phi$-VAE} \big) \nonumber
\end{align}

\noindent
\textbf{p$^{\boldsymbol{3}}$VAE training VS $\boldsymbol{\phi}$-VAE training} The difference between decoders is related to differences in the training algorithms. A key issue when combining physics models with neural networks is to regularize the flexibility of the neural networks. $\phi$-VAE relies on a regularization strategy based on the fact that the physics model $f_P$, by itself, can provide a reconstruction of the data from the latent variables $\z_P$. Therefore, the training algorithm of $\phi$-VAE can not, by nature, be applied to the optimization of p$^3$VAE. In contrast, p$^3$VAE relies on the partial supervision of the intrinsic latent factors and on the stop-gradient operator in order to regularize the flexibility of the neural network. 

\section{Experiments}\label{sec5}

In order to show the generic aspect of p$^3$VAE, we conducted numerical experiments on four data sets and three tasks. First, we considered synthetic time series data of damped pendulum angles, that have already been discussed in section \ref{sec2}. 
Second, we considered two real-world problems facing concrete issues: hyperspectral image classification (synthetic and real data) and methane plume inversion from hyperspectral data (semi-real data). We compared p$^3$VAE with standard baselines and state-of-the-art methods, selected according to the data and tasks. Besides qualitative comparisons, we reported mean absolute errors (MAE) for regression tasks, F1 score for classification tasks, and the MIG metric \cite{NEURIPS2018_1ee3dfcd}, when possible, that measures disentanglement. Further details on the MIG metric are given in Appendix \ref{sec:appendix_mig}.

\subsection{Damped pendulum \label{sec:pend_exp}}

\begin{table}[t]
\begin{center}
%\begin{minipage}{0.48\textwidth}
\caption{Mean absolute error of reconstruction (in-MAE), mean absolute error of extrapolation (out-MAE) and mutual information gap (MIG) on the test data, averaged over 5 runs. \label{tab:pendulum_metrics}}%
%\begin{tabular*}{\textwidth}{@{}ccccccc@{}}
\begin{tabular*}{0.5\textwidth}{lcccc}
\toprule
& & & \multicolumn{2}{@{}c@{}}{\textbf{MIG} $\uparrow$}\\ \cmidrule{4-5}%
\textbf{Model}  & \textbf{in-MAE} $\downarrow$ & \textbf{out-MAE} $\downarrow$ & $\omega$ & $\xi$ \\ \cmidrule{1-5}

VAE & \textbf{5.2} & 1.4E2 & 1.6E-2 & 1.6E-2 \\

$\phi$-VAE & 8.4 & 9.2E1 & 3.8E-1 & \textbf{1.5E-1} \\

p$^3$VAE & 5.3 & \textbf{3.9E1} & \textbf{1.1} & 9.6E-2 \\ \toprule
\end{tabular*}
%\end{minipage}
\end{center}
\end{table}

In this section, we address the problem described in section \ref{sec:example}.

\subsubsection{Data set}

We generated data from eq. \ref{eq:pendulum_ode}. Each data point $\x$ is a sequence $\x := [\vartheta(0) \: \vartheta(\Delta t) \: \ldots \: \vartheta((\tau - 1) \Delta t)]^T \in \mathbb{R}^\tau$ for $\Delta t = 0.05$ and $\tau = 100$, where $\vartheta(t)$ is the pendulum's angle at time $t$. The pendulum's angular frequency $\omega$, the fluid damping coefficient $\xi$ and the initial condition $\vartheta(0)$ ($\dot{\vartheta}(0)$ is fixed) were randomly drawn for each data point. We generated 2,500 sequences and separated them  into a training, validation, and test sets with 1,000, 500, and 1,000 sequences, respectively. 

\subsubsection{Prior physical knowledge $f_E$}

As explained in section \ref{sec:example}, $f_E : \mathbb{R}^\tau \times \mathbb{R}_+ \longrightarrow \mathbb{R}^\tau$ describes how the fluid damping coefficient involves an exponential decrease of the pendulum's angle:
\begin{align*}
	f_E(\x, \xi) = \x \: \mbox{exp}(-\xi \boldsymbol{t})
\end{align*}
where $\boldsymbol{t} = [0 \: \Delta t \: \ldots \: (\tau - 1) \Delta t]^T$. 
%While we assume that $f_I$ is unknown, we make the assumption that $\int_0^\vartheta f_I(u)du \geq 0$ in order to guarantee the stability of the ODE. 
The neural network $f_I^{\param}$ is meant to recover the unknown dynamics described by the term $\omega^2 \mbox{ sin } \vartheta(t)$ in eq. \ref{eq:pendulum_ode}. Additional details about the derivation of $f_E$ are deferred to Appendix \ref{sec:appendix_pendulum}.

\subsubsection{Competing methods}

\noindent
\textbf{VAE + ODE solver} \cite{yildiz2019ode2vae, tothhamiltonian} A VAE whose decoder is $\mathbb{E}[\x] = \mathcal{F}_{\tiny{\mbox{ODE}}}[\ddot{\vartheta}  + \mbox{NN}_{\param}(\z) = 0]$, where $\mathcal{F}$ is a differentiable solver of an ODE with respect to $\vartheta$. It was used as a baseline in \cite{takeishi2021physics}. 

\noindent
\textbf{$\boldsymbol{\phi}$-VAE} \cite{takeishi2021physics} The damped pendulum case is the only problem for which we could compare with ${\phi}$-VAE. 
For this specific case, prior physical knowledge can be integrated both with a combination in the form $f_E \circ f_I^{\param}$ (p$^3$VAE) and additively in the form $f_P + f_A$ (${\phi}$-VAE). 
The decoder of ${\phi}$-VAE and the physics model are defined as follows: 
\begin{align}
	\mathbb{E}[\x] & = \mathcal{F}_{\tiny{\mbox{ODE}}}[\ddot{\vartheta} + f_A(\z_A) + f_P(\z_P) = 0] \\
	f_P(\z_P) & = \z_P \frac{d\vartheta}{dt}(t)
\end{align}
where $\z_P$ is meant to encode $\xi$, and $f_A$ is meant to recover the term $\omega^2 \mbox{ sin } \vartheta(t)$ in eq. \ref{eq:pendulum_ode}.

\subsubsection{Optimization}

Every models were trained with the Adam stochastic descent gradient algorithm \cite{kingma2014adam} for 2000 epochs. We used the original and reproduced implementations of \cite{takeishi2021physics} for $\phi$-VAE and VAE + ODE solver. 

\subsubsection{Results}

Tab. \ref{tab:pendulum_metrics} shows the mean absolute error for reconstruction (in-MAE) and extrapolation (out-MAE), and the mutual information gap (MIG) for the generative  factors $\omega$ and $\xi$. While the VAE + ODE solver baseline reached a low in-MAE, it had significantly worse performance for extrapolation than $\phi$-VAE and p$^3$VAE. $\phi$-VAE had also much worse performance for extrapolation than p$^3$VAE. Those gaps in extrapolation performance are illustrated in Fig. \ref{fig:pendulum}, and in Fig. \ref{fig:appendix_pendulum}. In terms of disentanglement, p$^3$VAE significantly outperformed baselines for the pendulum's angular frequency $\omega$. However, $\phi$-VAE reached a higher MIG than p$^3$VAE for the fluid damping coefficient $\xi$. p$^3$VAE low MIG score is likely due to the highly non-linear relation between $\xi$ and the observed data $\x$, modeled by the physics prior $f_E$.

\subsection{Hyperspectral image classification}

\begin{figure}[t]%
    \centering
    \begin{minipage}[c]{.35\textwidth}
    \subfloat[Observed reflectance spectra]{\includegraphics[width=\textwidth]{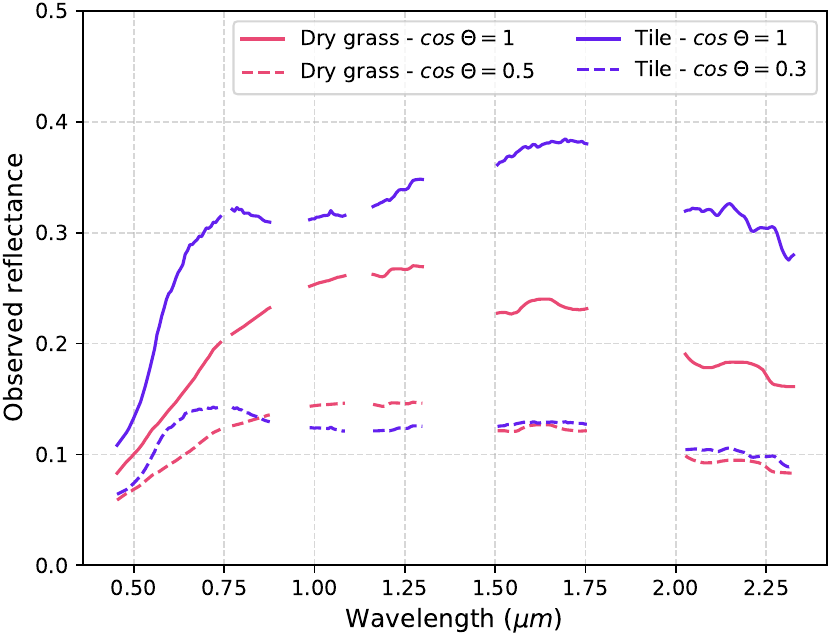}}
    \end{minipage}
    \hspace{0.5cm}
    \begin{minipage}[c]{.45\textwidth}
    \center
    \subfloat[False-color composition of the hyperspectral image]{\includegraphics[width=0.35\textwidth]{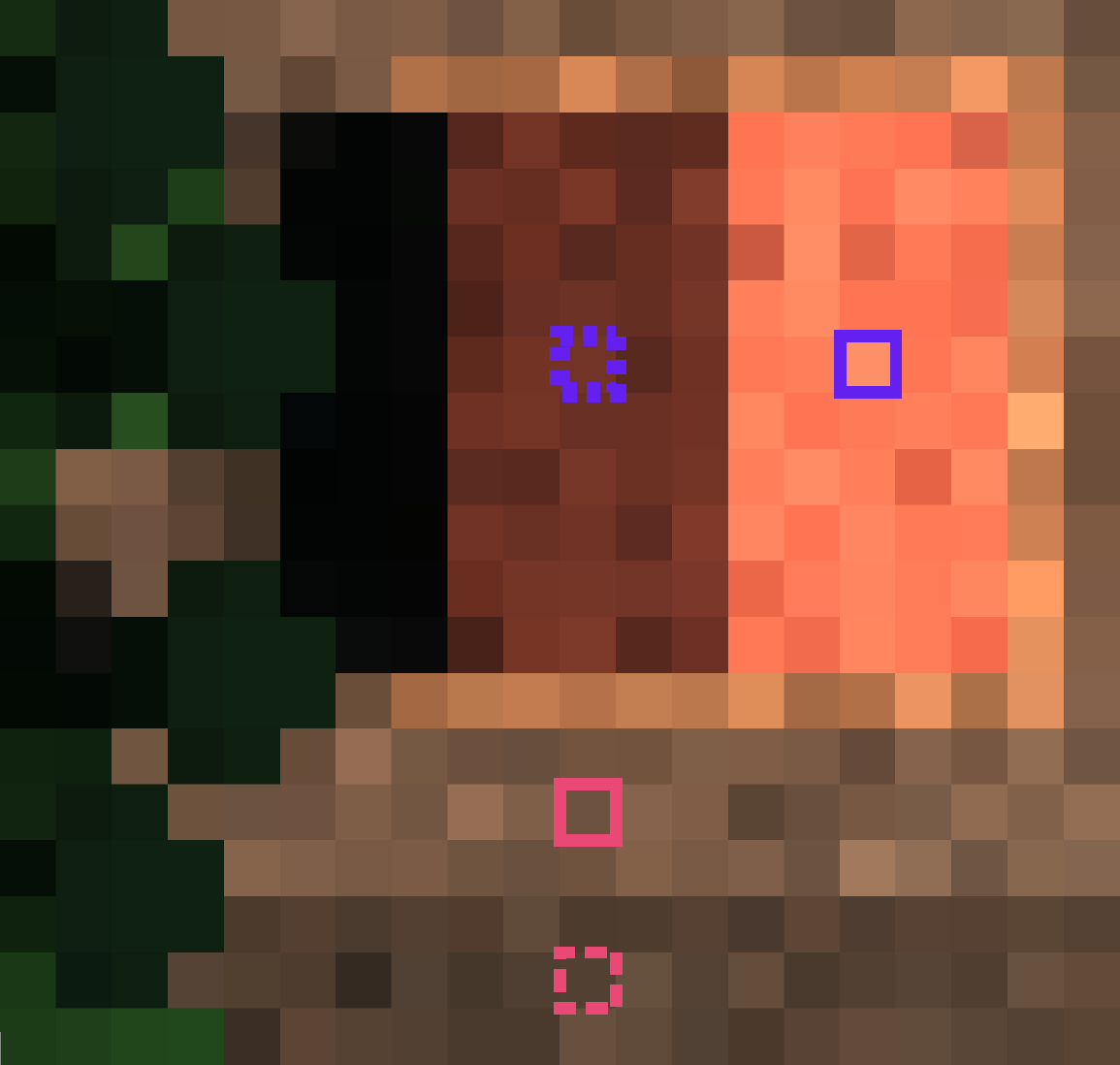}}
    \hspace{0.8cm}
    \subfloat[3D view of the scene]{\includegraphics[width=0.48\textwidth]{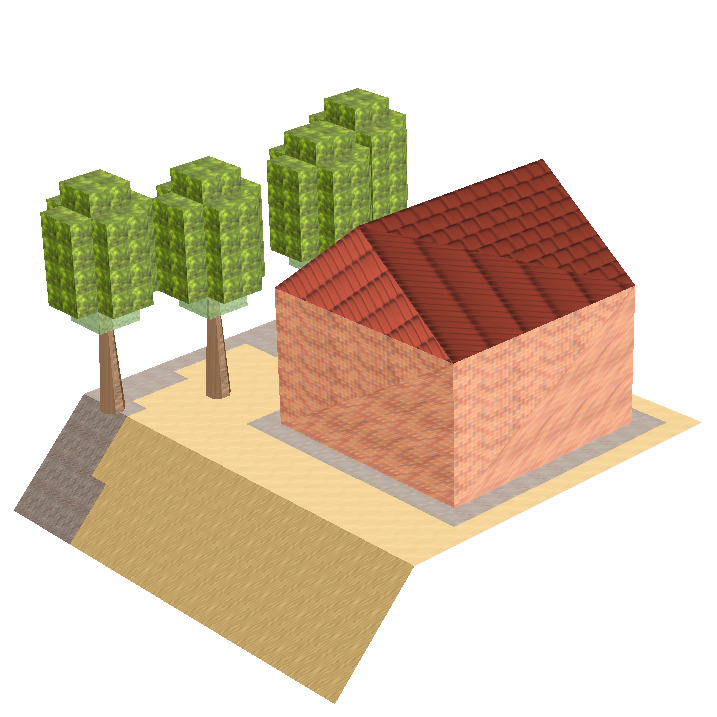}}
    \end{minipage}
    \caption{Illustration of intra-class variability induced by environmental factors of variation. The topography of the scene locally changes the solar incidence angle $\Theta$, resulting in variations of the local irradiance. Irradiance variations yield non-linear spectral shifts, as shown by the two spectra of tile (in purple) on different sides of the roof, and by the two spectra of bare soil (in pink) on the flat ground and the hillside. \label{fig:reflectance_estimates}}
\end{figure}

In this section, we address the pixel-wise classification of airborne hyperspectral images for land cover mapping. Unlike standard RGB images, hyperspectral images offer high spatial and spectral resolutions with hundreds of spectral channels, providing detailed ground-level reflectance data. 
Reflectance is a physical property intrinsic to matter and therefore highly informative of the land cover (\textit{e.g.} tile, bare soil). However, reflectance can vary due to the image acquisition conditions, impacting the local irradiance, and to intrinsic factors such as surface roughness, humidity, or aging. 
Fig. \ref{fig:reflectance_estimates} illustrates how variations of local irradiance modify hyperspectral data.
Meanwhile, annotated data are scarce, as labeling hyperspectral images requires expensive field campaigns and expert analysis. 
Therefore, we use p$^3$VAE to leverage prior physical knowledge in order to improve the robustness of learned latent representations to environmental-based intra-class variability.

\subsubsection{Data sets}

\noindent
\textbf{Synthetic data} We simulated an airborne hyperspectral image with the radiative transfer software DART \cite{gastellu2012dart} (300 spectral bands with a $6.5$ nm resolution from 450 nm to 2300 nm, and a $0.5$ m ground sampling distance). The scene comprises five materials (vegetation, sheet metal, sandy loam, tile and asphalt). In order to reproduce the scarcity of annotations in remote sensing, we only partially labeled the image. Additional details are deferred in Appendix \ref{sec:appendix_hyp_dart_data}. 

\noindent
\textbf{Real data} We used subsets of the recently published airborne hyperspectral data set of the CAMCATT-AI4GEO experiment\footnote{Data is publicly available here: \url{https://camcatt.sedoo.fr/}} in Toulouse, France \cite{ROUPIOZ2023109109}.
We labeled a ground truth through a field campaign and photo interpretation. We selected eight land cover classes (Tile, Asphalt, Vegetation, Painted sheet metal, Water, Gravels, Metal and Fiber cement) which were spatially split in a labeled training set, an unlabeled training set and a test set.
The labeled training set, unlabeled training set and test set approximately contain 4,000 pixels, 8,000 pixels and 10,000 pixels, respectively. Additional details are deferred in Appendix \ref{sec:appendix_hyp_real_data}.

\subsubsection{Prior physical knowledge $f_E$ \label{sec:prior_hyp_seg}}

Hyperspectral sensors measure spectral radiance, which depends, in the reflective domain, on sun irradiance, atmospheric composition, and land cover. Ground reflectance $\x \in [0,1]^B$, with $B$ the number of spectral bands, is estimated by atmospheric correction codes, such as COCHISE \cite{miesch2005direct}, from the sensor-level radiance. Most atmospheric correction codes assume a flat ground, hence uniform irradiance over every pixels, resulting in different reflectance estimates for the same land cover with different local irradiance (that depends on the scene 3D geometry, \textit{i.e.} what we call acquisition conditions).

\begin{figure}[h]
\begin{minipage}{0.3\textwidth}
\begin{center}
\includegraphics[width=\linewidth]{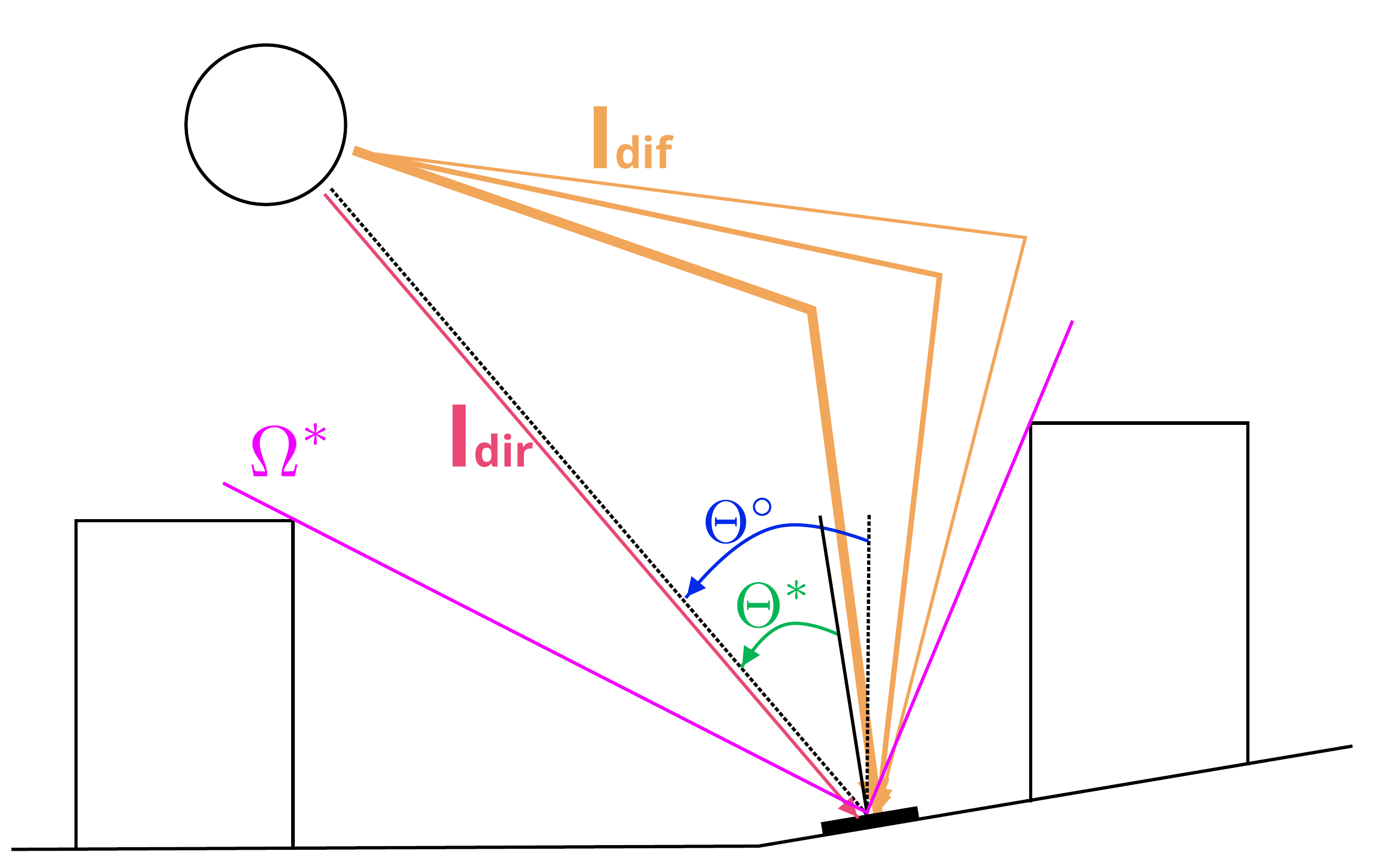}
\end{center}
\captionof{figure}{Illustration of irradiance terms. \label{fig:radiance}}
\end{minipage}
\begin{minipage}{0.6\linewidth}
\begin{minipage}{0.48\linewidth}
\begin{center}
\includegraphics[width=\linewidth]{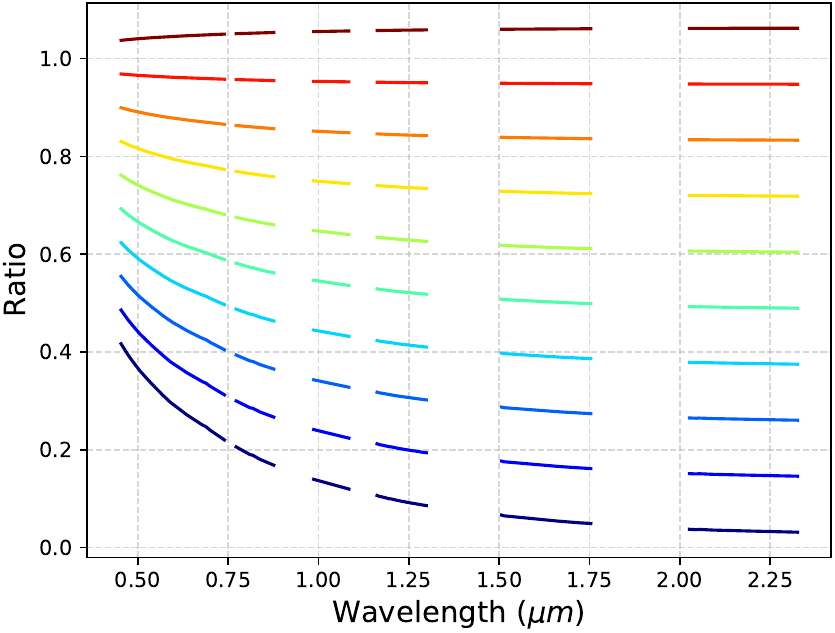}
\end{center}
\caption*{$I_{dif} = I_{dif}^o$, $\delta_{dir} \in \mbox{[}0,1\mbox{]}$}
\end{minipage}
\begin{minipage}{0.48\linewidth}
\begin{center}
\includegraphics[width=\linewidth]{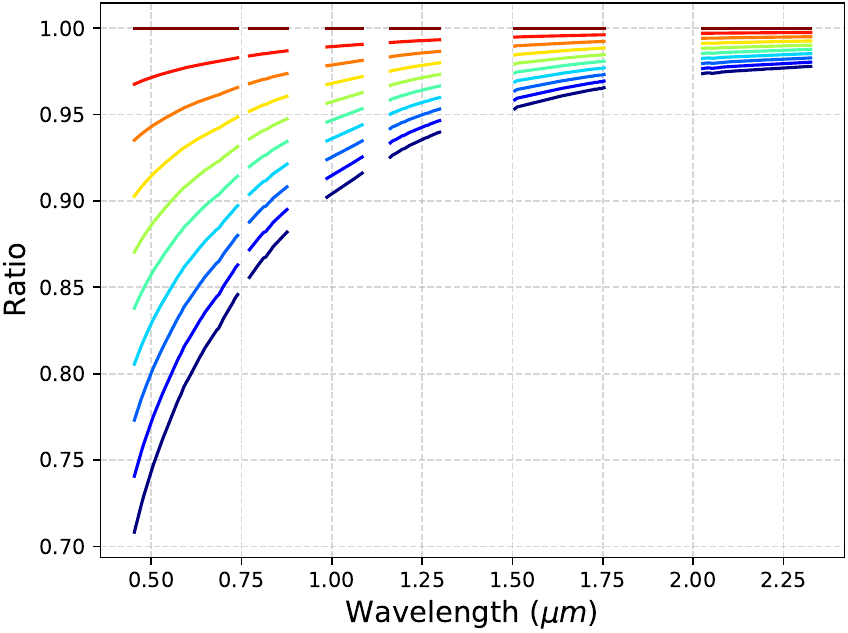}
\end{center}
\caption*{$I_{dir} = I_{dir}^o$, $\Omega \in \mbox{[}0.3,1\mbox{]}$}
\end{minipage}
\captionof{figure}{Ratio of the estimated reflectance over the true reflectance for varying irradiance conditions. \label{fig:ratio}}
\end{minipage}
\end{figure}

\noindent
Thus, we aim to integrate into the learning process a priori information defined by the ratio of the estimated reflectance $\x$ by an atmospheric
correction code over the true reflectance\footnote{The reflectance that would be computed by an atmospheric correction code with the exact knowledge of the local irradiance conditions.} $\x^\ast$. We approximate this quantity (see Appendix \ref{sec:appendix_radiative_modeling} for further details), that expresses how the signal varies at a wavelength $\lambda$ given variations of the direct irradiance, denoted as $I_{dir}$, and the diffuse irradiance (scattered by the atmosphere), denoted as $I_{dif}$:
\begin{equation}\label{eq:cochise_correction}
    \frac{\x}{\x^\ast}(\lambda) \approx \frac{I_{dir}^\ast(\lambda) + I_{dif}^\ast(\lambda)}{I_{dir}^{code}(\lambda) + I_{dif}^{code}(\lambda)} 
\end{equation} 
where the superscript $\ast$ is given for non-observed true physical quantities and the superscript $code$ is given for physical quantities assumed by the atmospheric correction code. The true direct irradiance $I_{dir}^\ast$ differs from the assumed direct irradiance $I_{dir}^{code}$ in shadows and in slopes. We model shadows with a quantity $\delta_{dir}^\ast \in [0, 1]$ corresponding to the portion of pixel directly lit by sun, and slopes with the local solar zenith angle, denoted as $\Theta^\ast$, in contrast to the solar zenith angle assumed by the atmospheric correction code $\Theta^{code}$. The true diffuse irradiance $I_{dif}^\ast$ mainly differs from the assumed diffuse irradiance $I_{dif}^{code}$ for pixels from where a small portion of the sky is visible. This quantity is commonly named the sky viewing angle factor that we denote as $\Omega^\ast \in [0, 1]$. Overall, we model the true direct and diffuse irradiance as follows:
\begin{align}
	I_{dir}^\ast(\lambda) & = \delta_{dir}^\ast \cdot \frac{cos \: \Theta^\ast}{cos \: \Theta^{code}} \cdot I_{dir}^{code}(\lambda) \\
	 I_{dif}^\ast(\lambda) & = \Omega^\ast \cdot p_\Theta^\ast \cdot I_{dif}^{code}(\lambda)
\end{align}
where $p_\Theta^\ast$ is a correction coefficient that accounts for the anisotropy of the diffuse irradiance. Fig. \ref{fig:radiance} illustrates the radiative terms used in our model and Fig. \ref{fig:ratio} illustrates the ratio we derived for varying irradiance conditions, showing how non-linear the spectral variations are with respect to direct and diffuse irradiance variations. 

\noindent
Therefore, $f_E$ (derived from eq. \ref{eq:cochise_correction}) computes the spectral shift induced by the variation of the local irradiance, encoded in $z_E$, given a reflectance spectrum, denoted as $\x^\ast$:
\begin{equation} \label{eq:fE_hyp}
	f_E(z_E, \x^\ast) = \frac{z_E \frac{I_{dir}^{code}}{cos \: \Theta^{code}} + g(z_E) I_{dif}^{code}}{I_{dir}^{code} + I_{dif}^{code}} \x^\ast 
\end{equation}
where $g$ is an affine function that empirically approximates the local diffuse irradiance (weight and bias of $g$ are hyperparameters). In eq. \ref{eq:fE_hyp}, $z_E$ is meant to encode the term $\delta_{dir}^\ast cos \: \Theta^\ast$ that describes the local direct irradiance. Concomitantly, the neural network $f_I^{\boldsymbol{\theta}}$ is meant to recover the true reflectance $\x^\ast$ given the land cover and the latent representation of the intrinsic intra-class variability $\z_I$. The intrinsic latent variable $\z_I$ models simultaneously the land cover class, denoted as $y$, and the intra-class variability through a vector denoted as $\vv$ that sums to one, and represents the contribution of $\mbox{dim}(\vv)$ sub-classes of $y$. Precisely, 
\begin{align*}
	f_I^{\param}(\z_I) = \sum_{k=1}^{\mbox{dim}(\vv)} \vv[k] \mbox{NN}_{\param}(y)[k]
\end{align*}
where $\mbox{NN}_{\param}$ computes $\mbox{dim}(\vv)$ $B$-dimensional latent reflectance spectra given the land cover $y$. Additional modeling choices about the latent variables are deferred in Appendix \ref{sec:appendix_latent_modeling_hyp}.

\subsubsection{Competing methods}

\noindent
\textbf{Spectral CNN} A standard CNN used in the hyperspectral literature \cite{audebert2019deep}. 

\noindent
\textbf{Semi-supervised VAE} \cite{kingma2014semi} A conventional semi-supervised gaussian VAE. The dimension of the latent space is equal to dim($z_E$) $+$ dim($\z_I$). 

\noindent
\textbf{ssInfoGAN} \cite{spurr2017guiding} A standard InfoGAN \cite{chen2016infogan} (GAN designed for disentanglement) trained with a semi-supervised algorithm. 

\noindent 
\textbf{Semi-supervised FG-UNET} \cite{stoian2019land} A semantic segmentation model designed to process remote sensing images, based on the U-net \cite{ronneberger2015u}. FG-UNET was compared on real images only, because synthetic test data are one dimensional (only the spectral dimension was simulated). 

\noindent
\textbf{Physics-guided VAE}. A VAE with prior and posterior distributions over the latent variables guided by physical considerations (same distributions than for our instance of p$^3$VAE, described in Appendix \ref{sec:appendix_latent_modeling_hyp}). It also has the same architecture than p$^3$VAE. The only, major, difference is that it does not have the physics model $f_E$.

\subsubsection{Optimization}\label{sec:exp_optim}

The CNNs and the VAE-like models were optimized with the Adam stochastic descent gradient algorithm \cite{kingma2014adam} for 100 and 50 epochs on the simulated and real data, respectively, and a batch size of 64. We optimized ssInfoGAN in the Wasserstein training fashion \cite{arjovsky2017wasserstein} with the gradient penalty loss introduced in \cite{gulrajani2017improved}, using the RMSProp algorithm for 200 and 150 epochs on the simulated and real data, respectively, and a batch size of 64. FG-UNET was trained with the Adam algorithm for 500 epochs and a batch size of 16. In \cite{stoian2019land}, FG-UNET is optimized by minimizing a standard cross-entropy between labels and predictions. In our experiments, we add an unsupervised reconstruction loss on the L1 norm like in \cite{castillo2021semi} to guide the training of FG-UNET with unlabeled pixels. We tuned the learning rate between $5\cdot10^{-5}$ and $1\cdot10^{-4}$ to reach loss convergence on the training and validation sets. We modeled p$^3$VAE empirical function $g : z_P \longmapsto \Omega \: p_\Theta $ by an affine transformation and tuned its parameters on the validation set. We retained $g(z_P) = z_P + 0.2$, which led to good spectral reconstruction under low direct irradiance conditions despite being very simplistic. We weighted the Kullback-Leibler divergence term in equations (\ref{eq:supervised_objective}) and (\ref{eq:unsupervised_objective}) by $\beta = 10^{-4}$ following \cite{higgins2016beta} technique. We also weighted the entropy term in equation (\ref{eq:unsupervised_objective}) to balance between high accuracy and high uncertainty with a $10^{-1}$ coefficient. Besides, we applied a Ridge regularization on the weights of the classifiers and encoders with a $10^{-2}$ penalty coefficient. 

\subsubsection{Results on simulated data}

\begin{table*}
\begin{center}
\begin{minipage}{\textwidth}
\caption{Mean F1 score per class over 10 runs on the simulated data}\label{tab1}%
%\begin{tabular*}{\textwidth}{@{}ccccccc@{}}
\scalebox{0.82}{
\begin{tabular*}{1.2\textwidth}{@{\extracolsep{\fill}}lccccccc@{\extracolsep{\fill}}}
\toprule
%Class & \multirow{2}{*}{Vegetation}  & \multirow{2}{*}{Sheet metal} & \multirow{2}{*}{Sandy loam} & \multirow{2}{*}{Tile} & \multirow{2}{*}{Asphalt} \\
& & \multicolumn{5}{@{}c@{}}{\textbf{Classes}} \\ \cmidrule{3-7}%
\multicolumn{2}{@{}l@{}}{\textbf{Model}} & Vegetation  & Sheet metal & Sandy loam & Tile & Asphalt & Average\\ \midrule
%\multicolumn{1}{r}{Inference model} & & & & & \\
CNN & $\qphi(y \vert \x)$ & 0.90 & 0.81 & 0.77 & 0.79 & 0.75 & 0.80 \\ 
& $\qphi(y \vert \x)$ (full annotations)\footnotemark[1]{} & 0.92 & 0.79 & \textit{0.90} & 0.87 & 0.86 & 0.86 \\ \cmidrule{1-8}

ssInfoGAN & $\qphi(y \vert \x)$ & 0.86 & 0.79 & 0.75 & 0.75 & 0.69 & 0.77 \\ \cmidrule{1-8}

\multirow{2}{*}{Gaussian VAE} & $\qphi(y \vert \x)$ & 0.93 & 0.80 & 0.87 & 0.86 & 0.74 & 0.84 \\ 
& $\ppdf(y \vert \x)$ & 0.94 & 0.88 & \textbf{0.90} & \textbf{0.92} & 0.85 & 0.90 \\ \cmidrule{1-8}

\multirow{2}{*}{Physics-guided VAE} & $\qphi(y \vert \x)$ & 0.93 & 0.81 & 0.86 & 0.86 & 0.76 & 0.84 \\
& $\ppdf(y \vert \x)$ & 0.86 & 0.85 & 0.83 & 0.75 & 0.77 & 0.81\\ \cmidrule{1-8}

\multirow{2}{*}{p$^3$VAE} & $\qphi(y \vert \x)$ & 0.92 & 0.82 & 0.88 & 0.87 & 0.81 & 0.86 \\
& $\ppdf(y \vert \x)$ & \textbf{0.96} & \textbf{0.97} & \textbf{0.90} & 0.90 & \textbf{0.93} & \textbf{0.93} \\
\toprule
\end{tabular*}
}
\end{minipage}
\end{center}
$^1$The CNN was optimized with every pixels labeled on the image (\textit{i.e.} the labeled and unlabeled sets showed on Fig. \ref{fig:traindataset}), in contrast with every other cases where the image is partially labeled.
\end{table*}

\begin{figure*}
	\center
	\captionsetup[subfigure]{labelformat=empty}
	\subfloat{\includegraphics[height=0.18\textwidth]{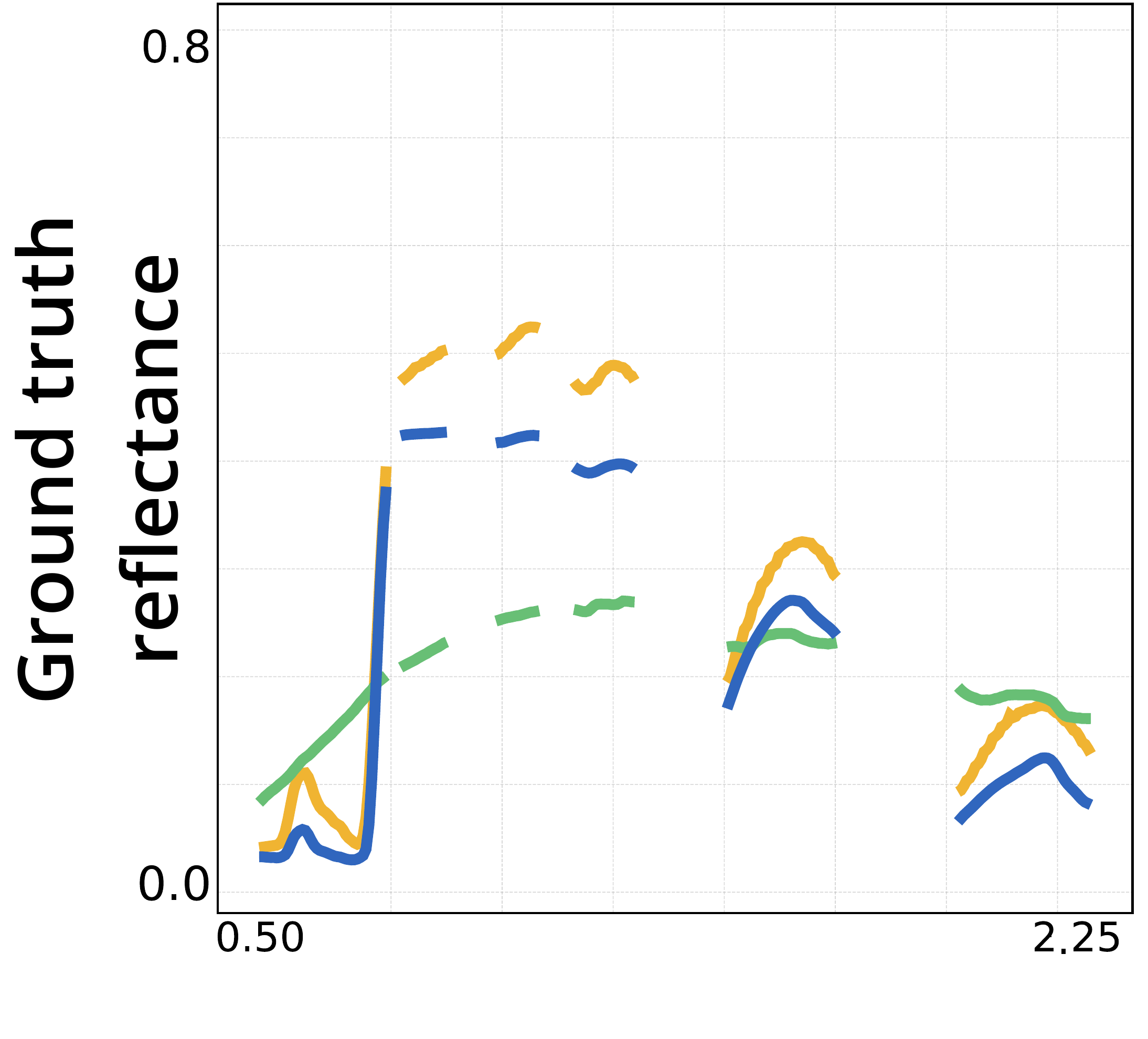}}
	\subfloat{\includegraphics[height=0.18\textwidth]{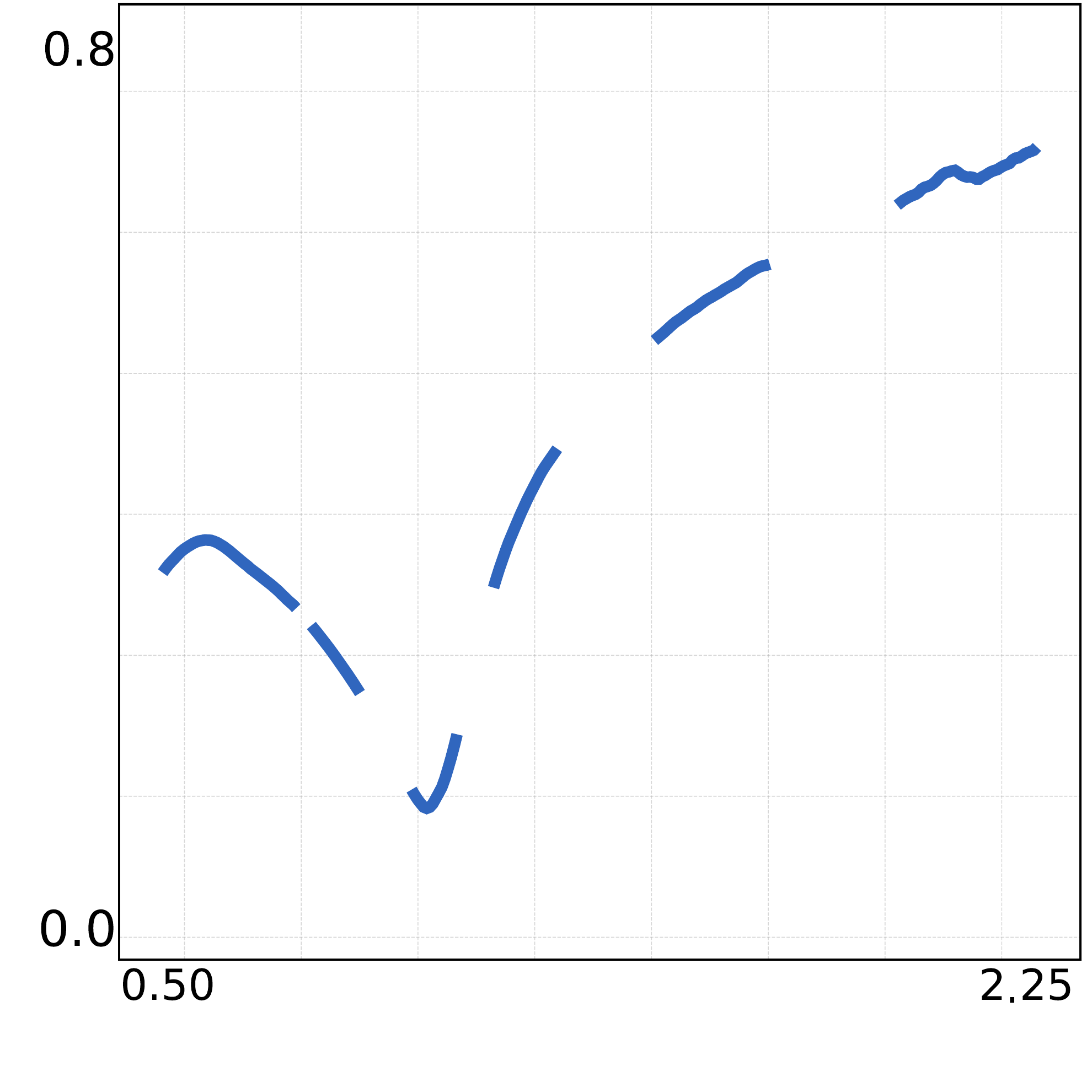}}
	\subfloat{\includegraphics[height=0.18\textwidth]{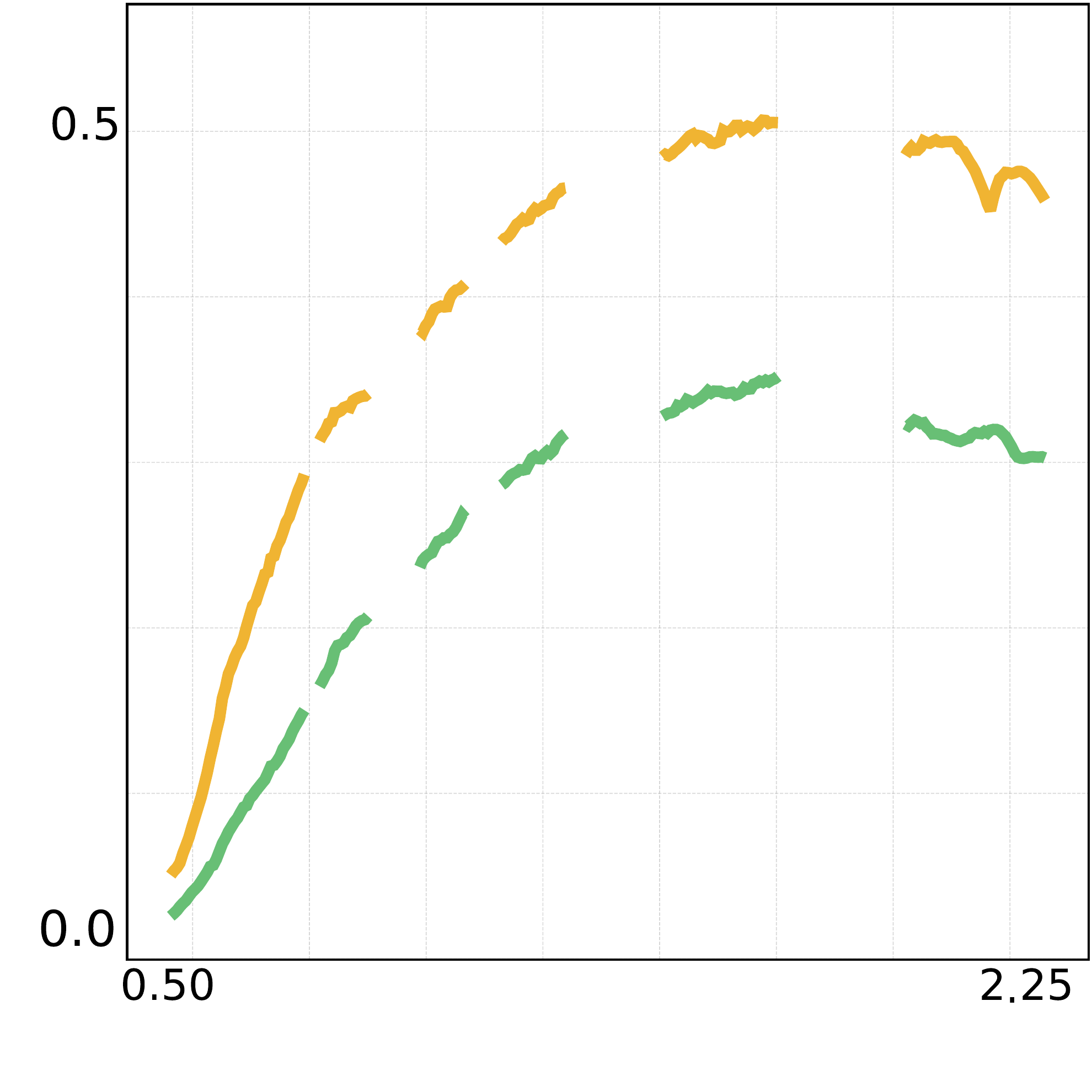}}
	\subfloat{\includegraphics[height=0.18\textwidth]{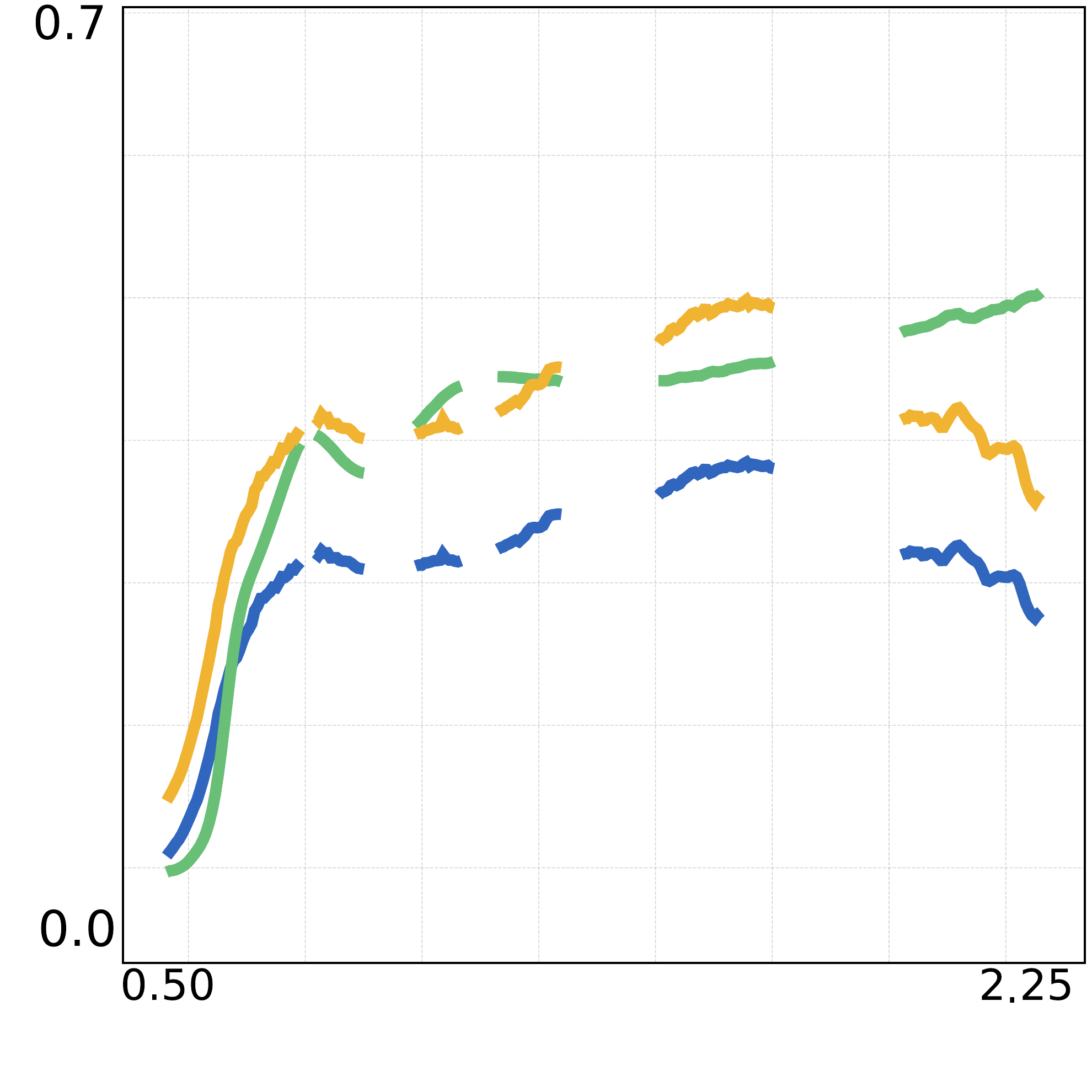}}
	\subfloat{\includegraphics[height=0.18\textwidth]{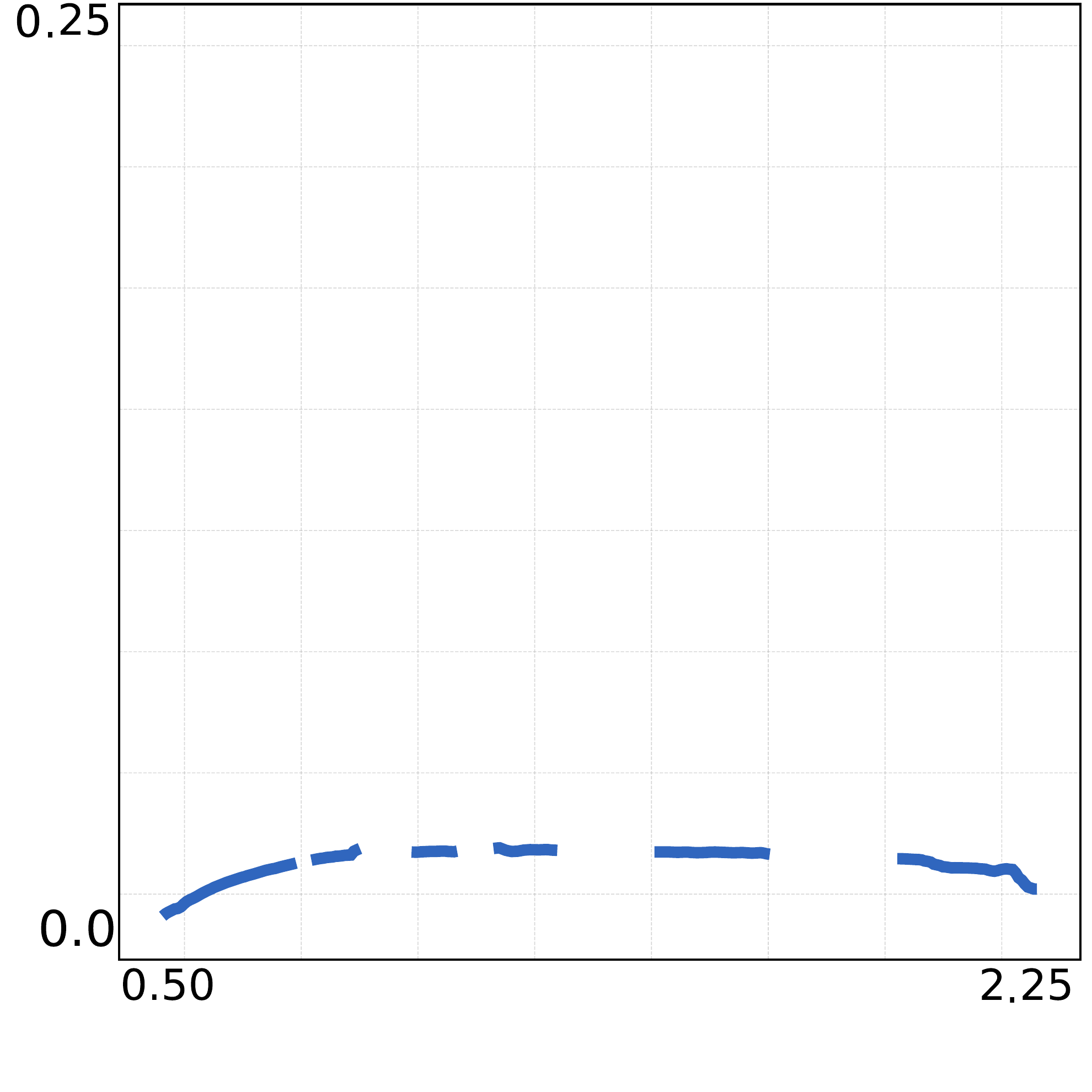}}
	
	\subfloat{\includegraphics[height=0.18\textwidth]{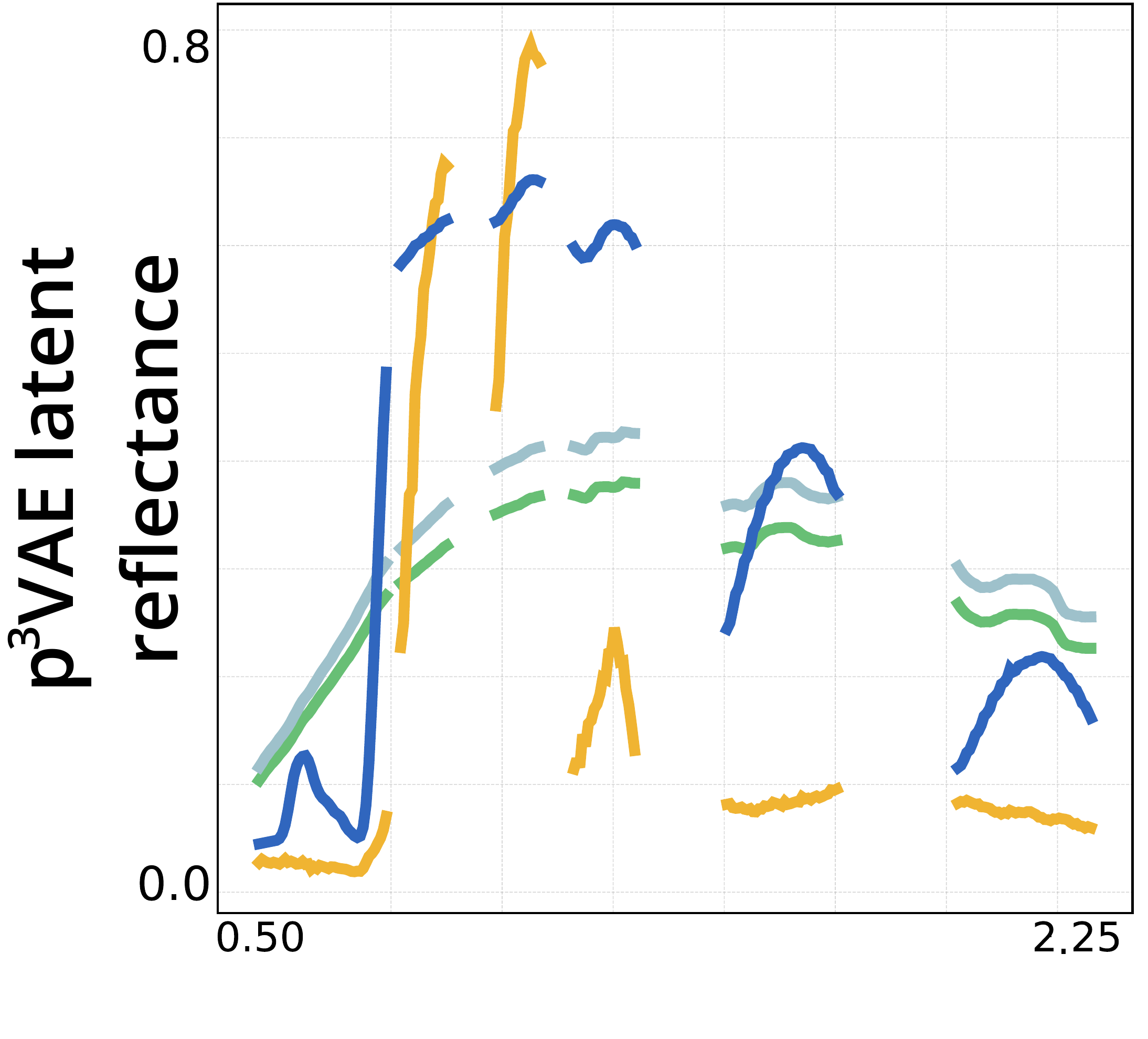}}
	\subfloat{\includegraphics[height=0.18\textwidth]{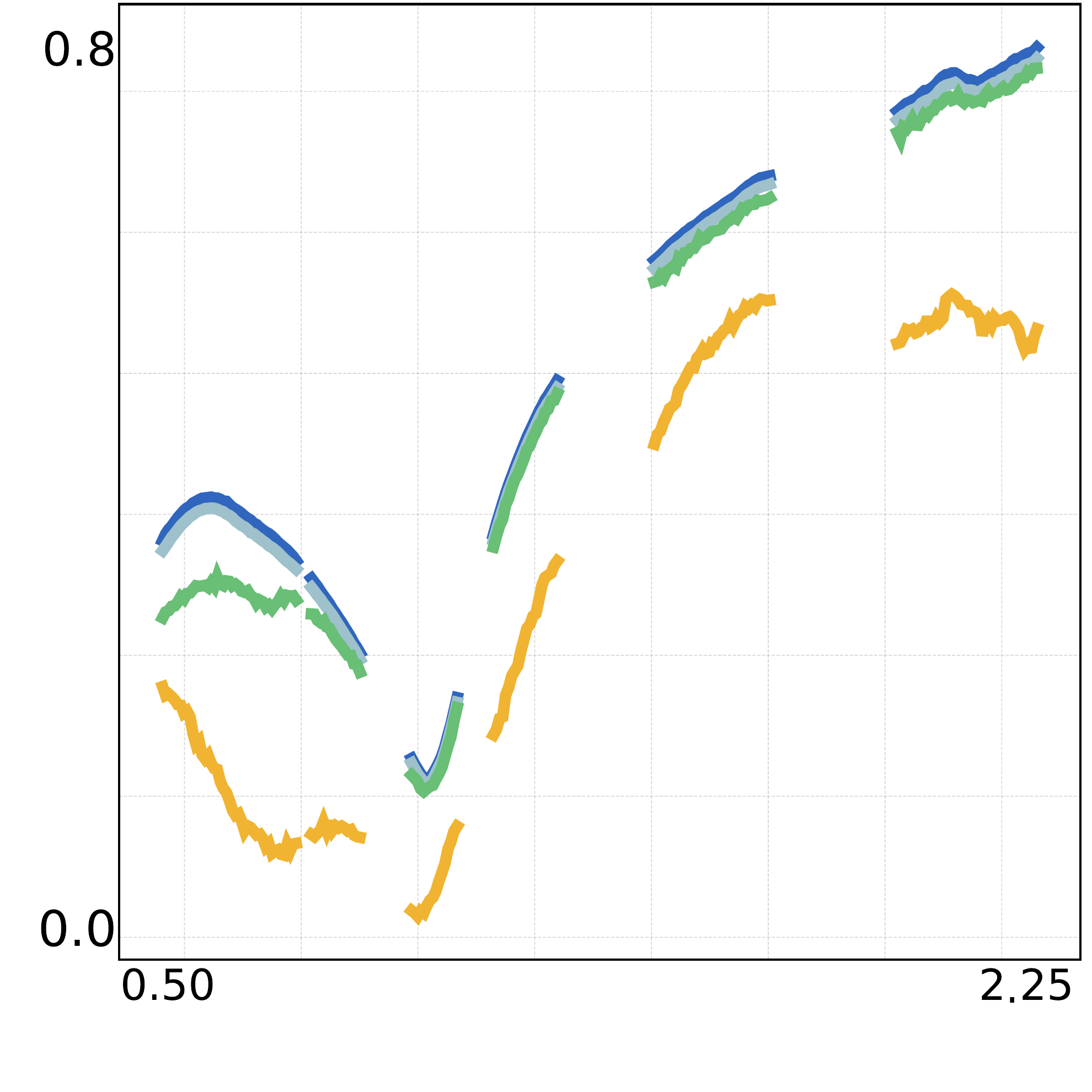}}
	\subfloat{\includegraphics[height=0.18\textwidth]{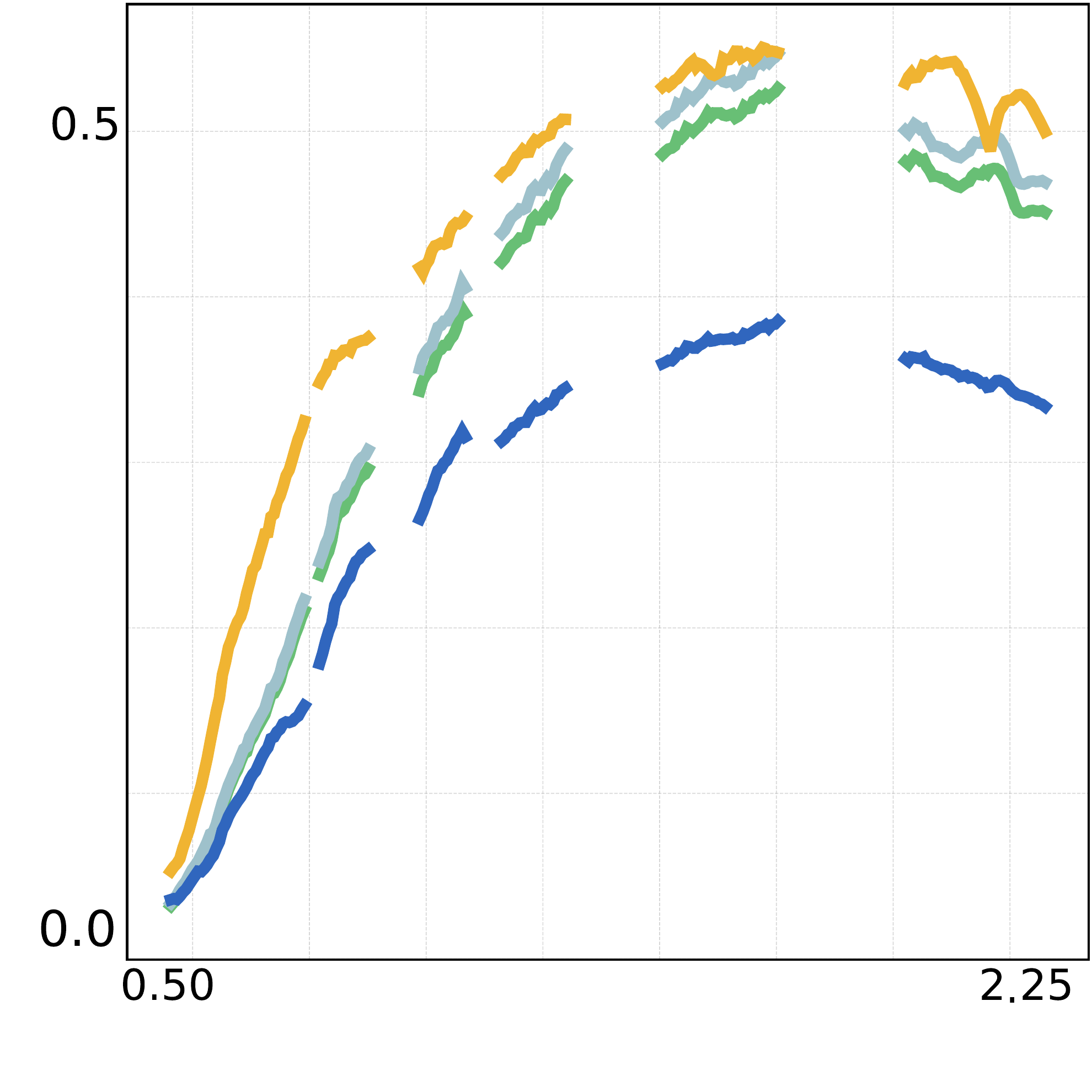}}
	\subfloat{\includegraphics[height=0.18\textwidth]{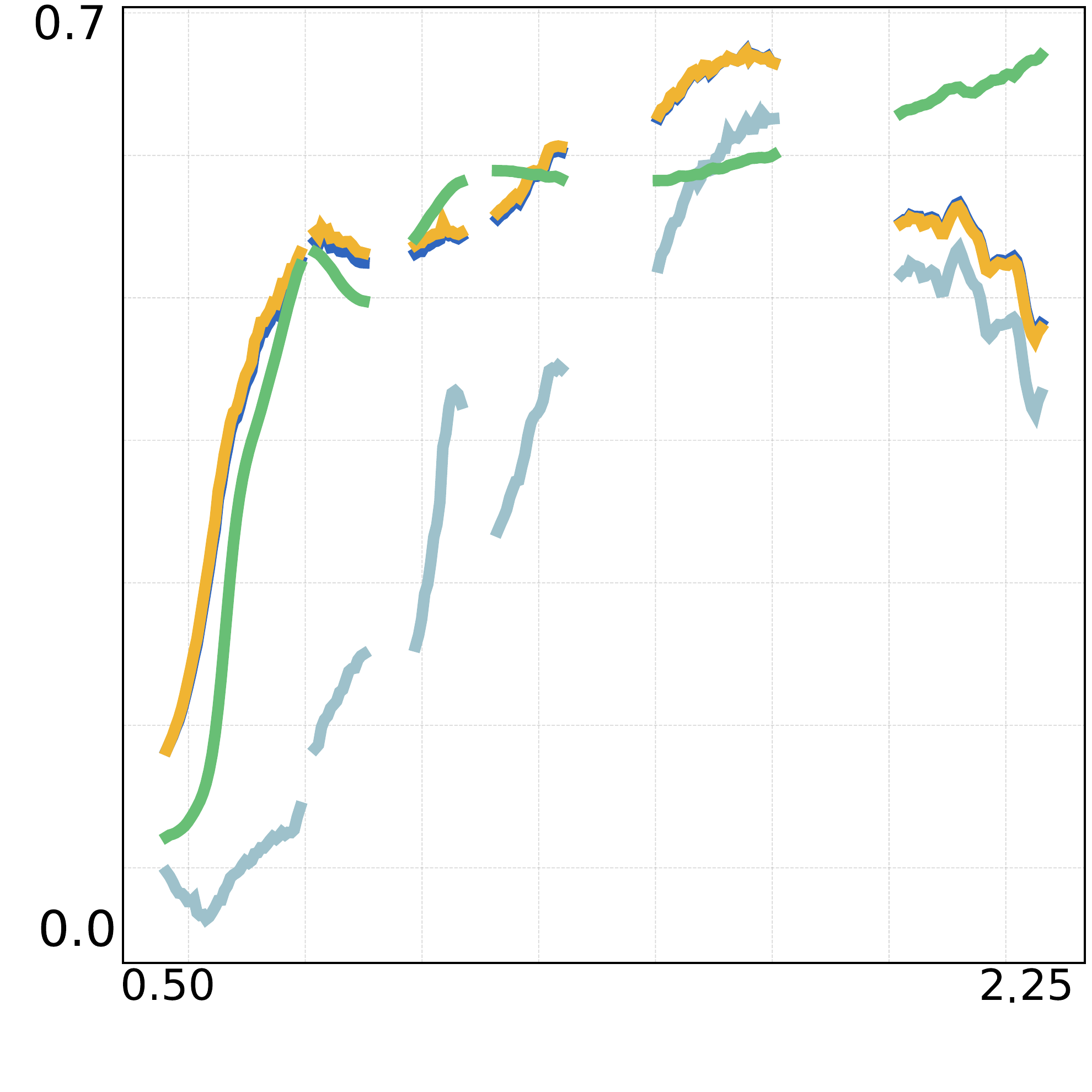}}
	\subfloat{\includegraphics[height=0.18\textwidth]{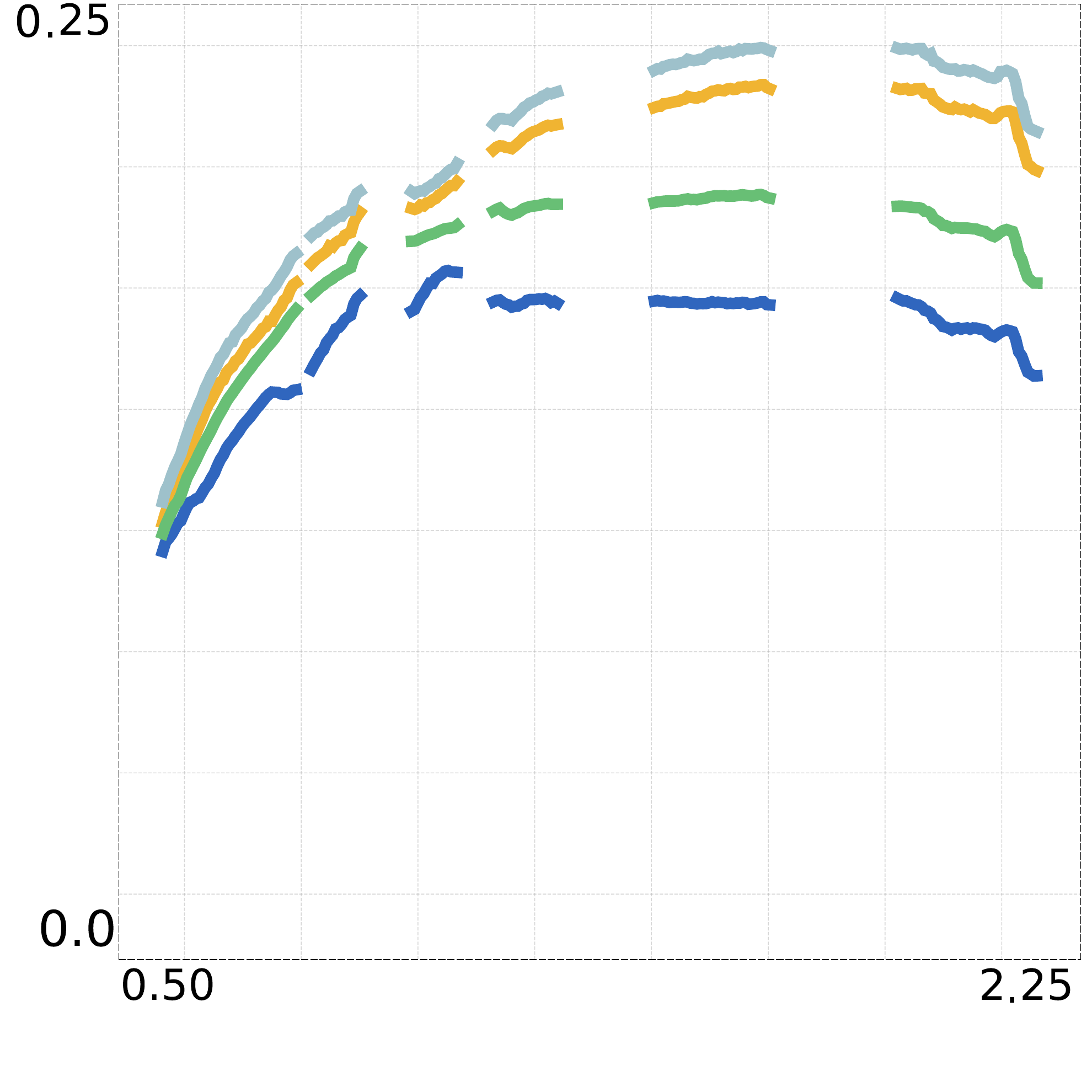}}
	
	\subfloat[(a) Vegetation]{\includegraphics[height=0.18\textwidth]{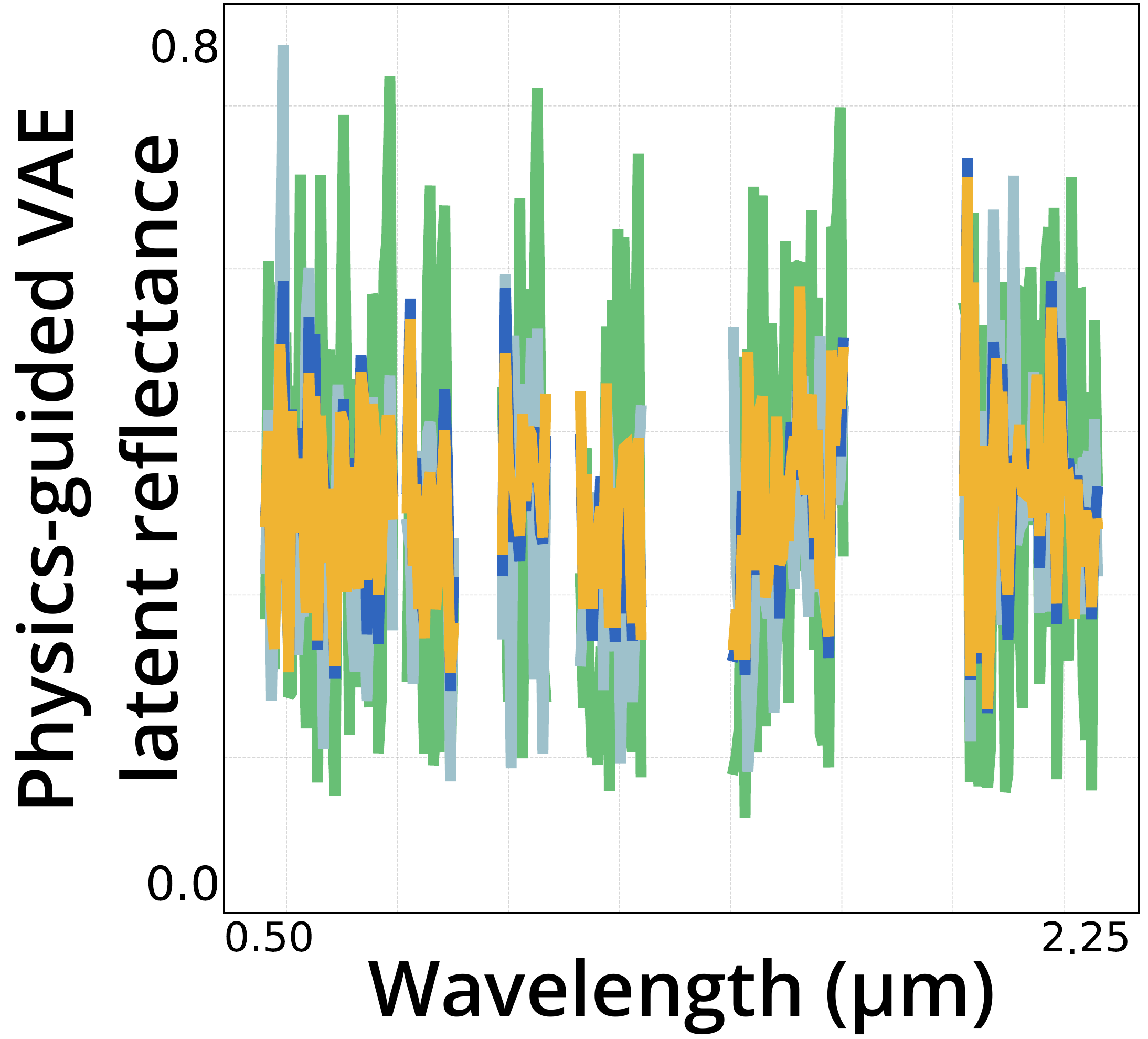}}
	\subfloat[(b) Sheet Metal]{\includegraphics[height=0.18\textwidth]{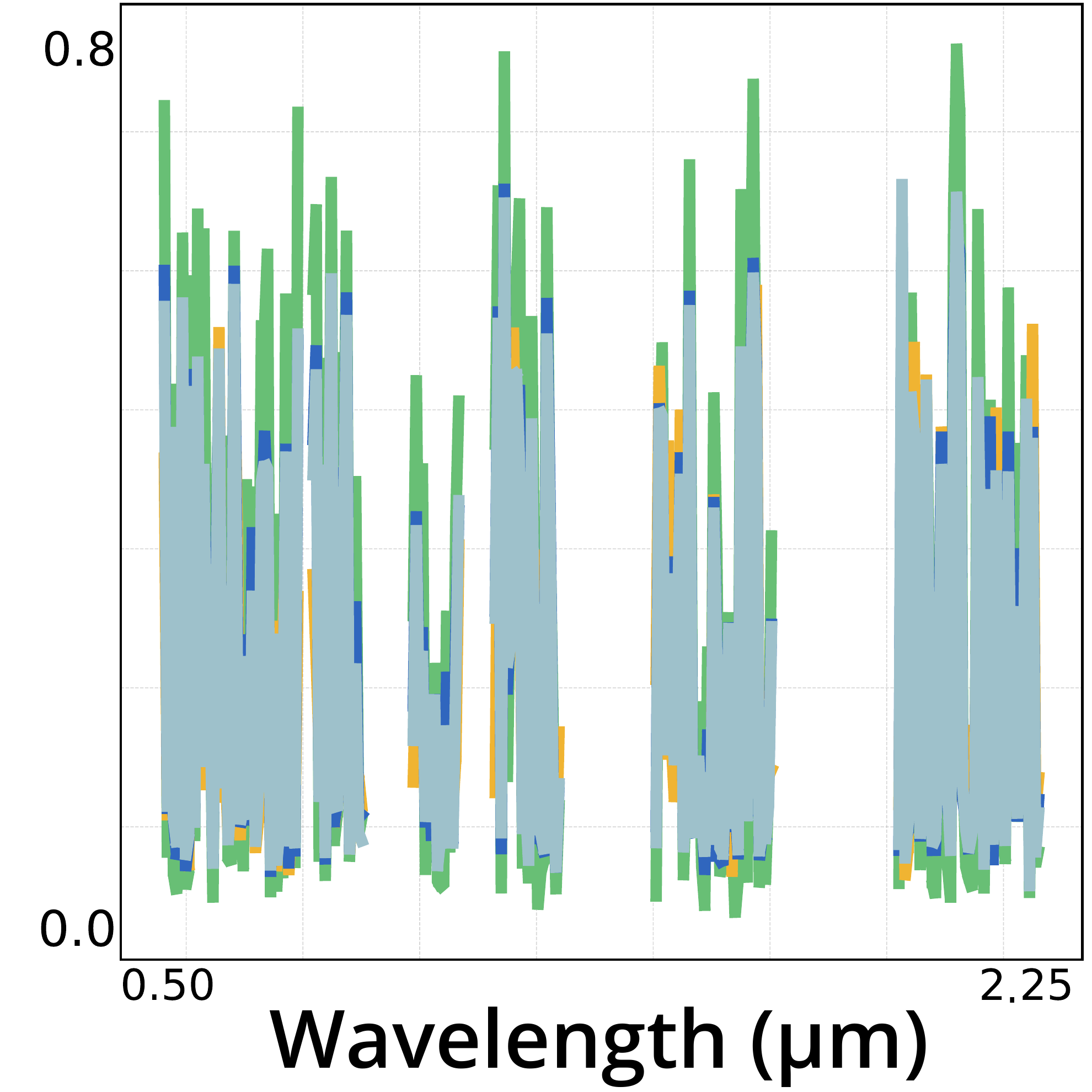}}
	\subfloat[(c) Sandy Loam]{\includegraphics[height=0.18\textwidth]{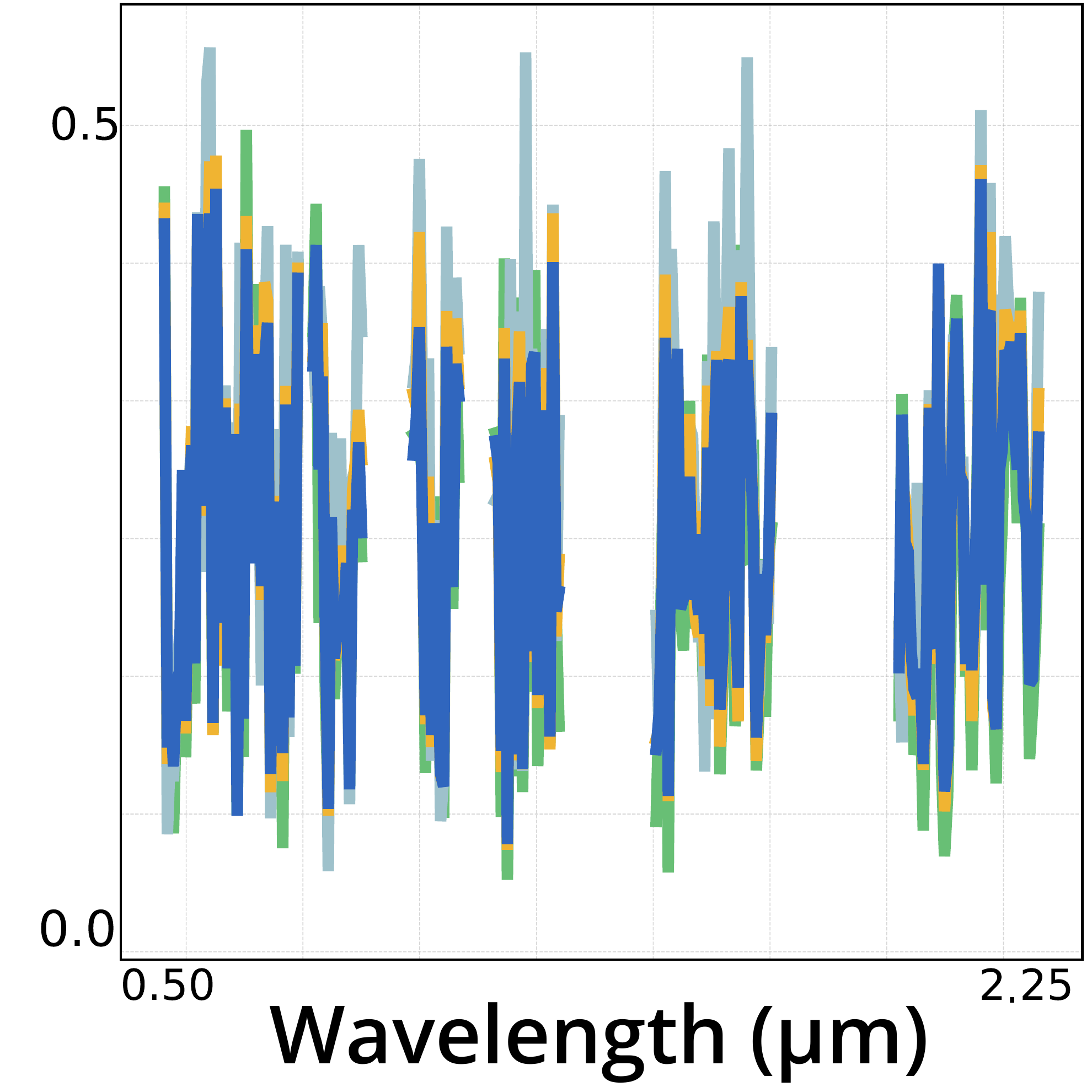}}
	\subfloat[(d) Tile]{\includegraphics[height=0.18\textwidth]{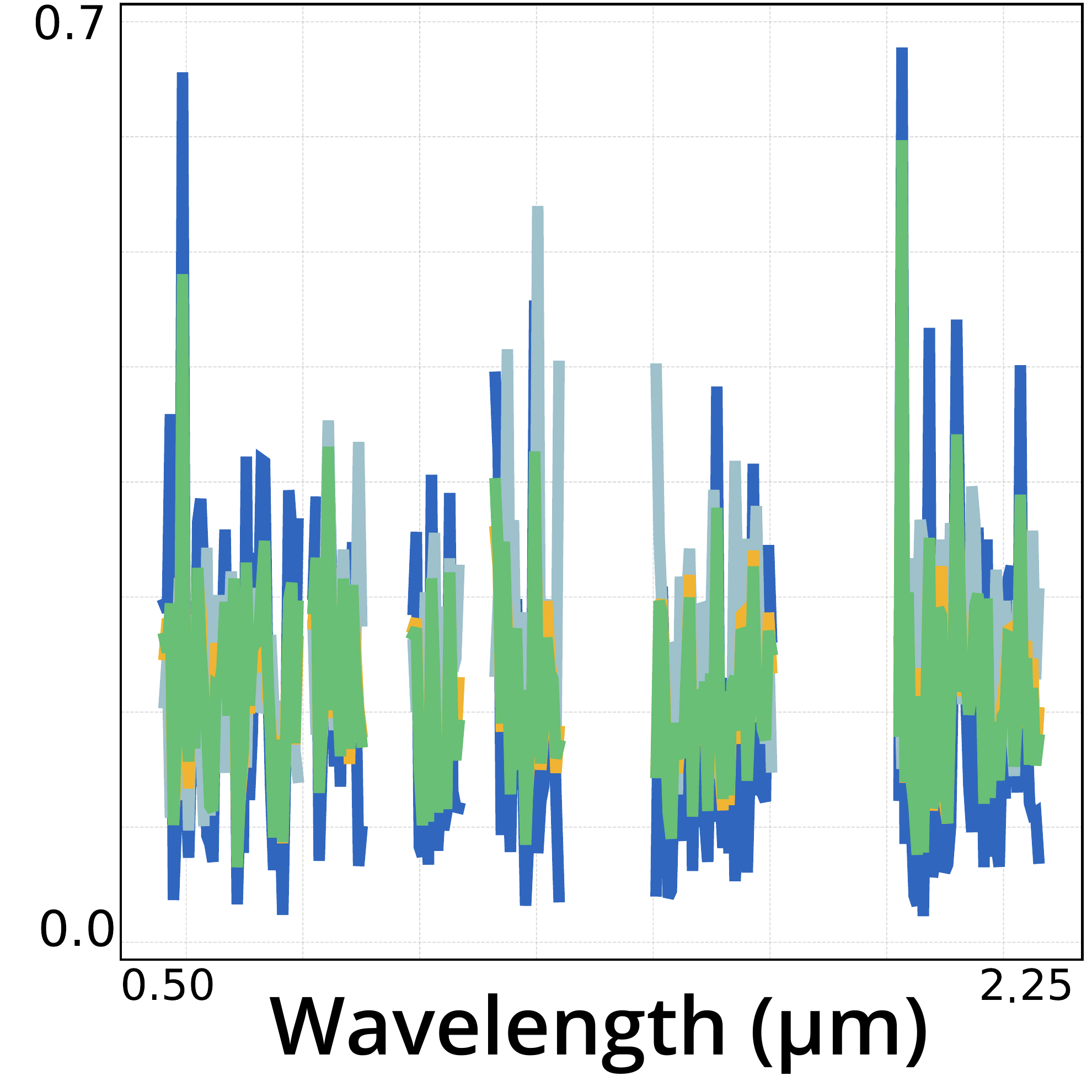}}
	\subfloat[(e) Asphalt]{\includegraphics[height=0.18\textwidth]{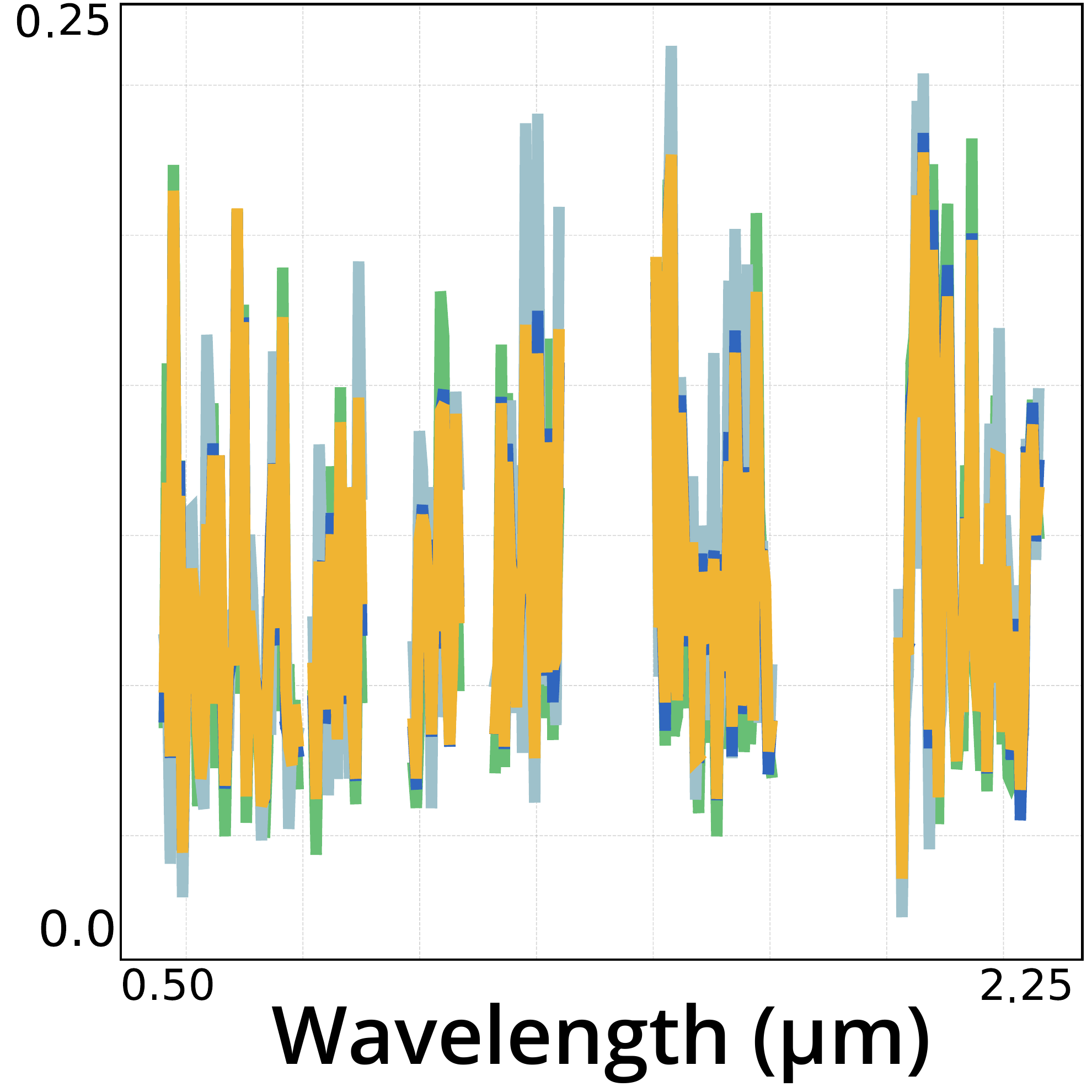}}
	
	\caption{Top row: ground truth reflectance spectra. Middle row: p$^3$VAE latent reflectance spectra. Bottom row: physics-guided VAE latent reflectance spectra. For each land cover class, $\mbox{dim}(\vv) = 4$ latent spectra are computed in order to represent the intrinsic intra-class variability, as defined in section \ref{sec:prior_hyp_seg}. The only difference between p$^3$VAE and its variant, the physics-guided VAE, is that p$^3$VAE integrates prior physical knowledge $f_E$. While p$^3$VAE latent reflectance spectra are physically plausible and close to the ground truth, the physics-guided VAE produced unrealistic spectra.  \label{fig:reflectance_estimates}}
\end{figure*}

\begin{table}
\begin{center}
%\begin{minipage}{0.48\textwidth}
\caption{Mean MIG over 10 runs for every generative factors of variation. The higher the better.}\label{tab:disentanglement}%
%\begin{tabular*}{\textwidth}{@{}ccccccc@{}}
\begin{tabular*}{0.5\textwidth}{lcccccc}
\toprule
& \multicolumn{5}{@{}c@{}}{\textbf{Factors}} & \\ \cmidrule{2-6}%
\textbf{Model}  & $y$  & $\delta_{dir}$ & $\Omega$ & $\alpha$ & $\eta$ \\ \cmidrule{1-7}

VAE & 0.62 & 0.18 & \textbf{0.096} & 0.20 & 0.20 \\

Physics-guided & 1.0 & 0.068 & 0.021 & 0.28 & 0.28\\

ssInfoGAN & 0.92 & 0.16 & 0.040 & 0.29 & 0.29 \\ 

p$^3$VAE & \textbf{1.2} & \textbf{0.85} & 0.012 & \textbf{0.31} & \textbf{0.31} \\ \toprule
\end{tabular*}
%\end{minipage}
\end{center}
\end{table}

\begin{figure*}[h]
	\center
	
	\begin{subfigure}{0.23\linewidth}
		\center
		\includegraphics[height=\textwidth]{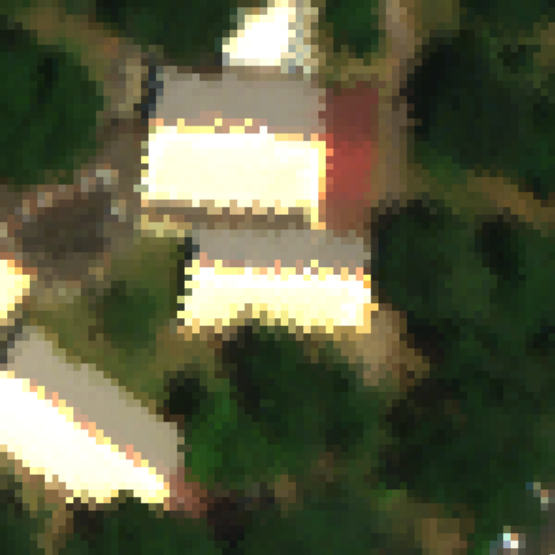}
		\caption{}
	\end{subfigure}
	\begin{subfigure}{0.23\linewidth}
		\center
		\includegraphics[height=\textwidth]{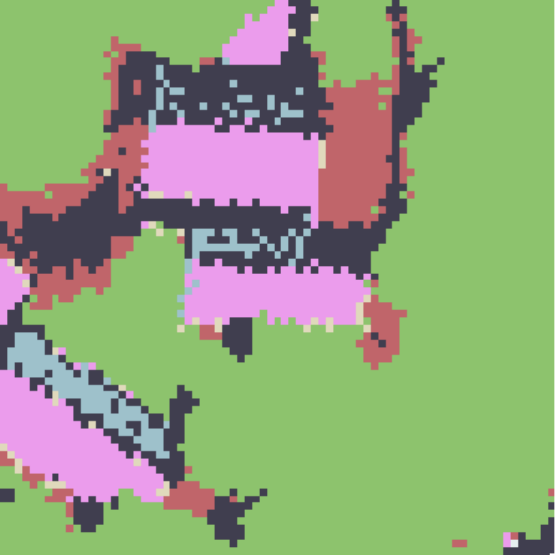}
		\caption{}
	\end{subfigure}
	\begin{subfigure}{0.23\linewidth}
		\center
		\includegraphics[height=\textwidth]{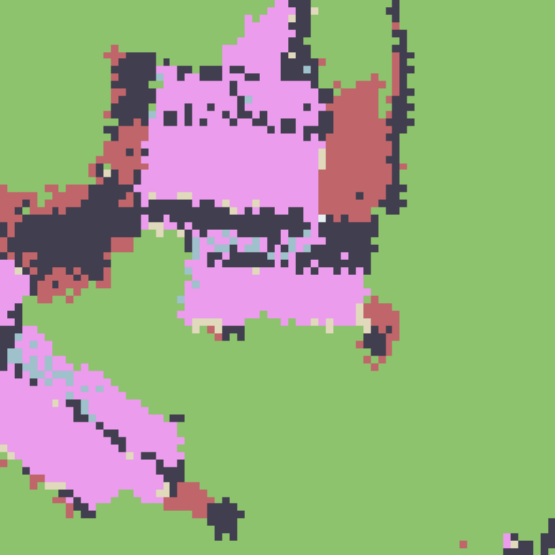}
		\caption{}
	\end{subfigure}
	\begin{subfigure}{0.23\linewidth}
		\center
		\includegraphics[height=\textwidth]{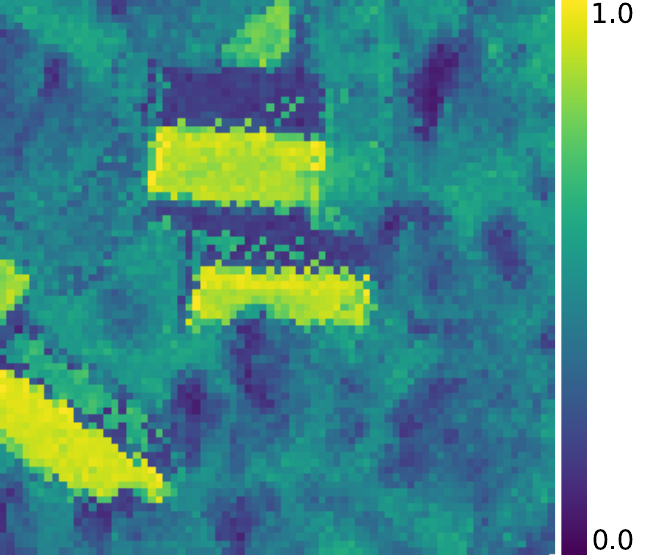}
		\caption{}
	\end{subfigure}
	
	\begin{subfigure}{\linewidth}
		\center
		\includegraphics[width=\textwidth]{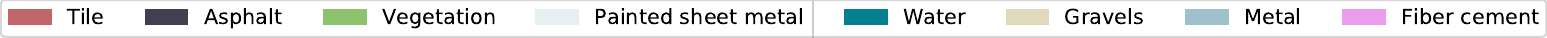}
	\end{subfigure}
	\caption{(a) RGB composition of a real test hyperspectral scene that comprises roofs in fiber cement. (b) Prediction of a standard VAE: fiber cement on slopes with low irradiance is confused with asphalt and painted sheet metal. (c) Prediction of p$^3$VAE: fiber ciment is mostly well classified, despite that the training data set did not include fiber cement observed under low irradiance conditions. (d) $z_E$ prediction by p$^3$VAE, that encodes the direct irradiance: low irradiance values are predicted on the sides of roof which are actually poorly illuminated, improving p$^3$VAE robustness to irradiance shifts. \label{fig:land_cover}}
\end{figure*}

\noindent
\textbf{Test accuracy} Mean F1 score per class over 10 runs is shown on Tab. \ref{tab1}. First, all semi-supervised models outperformed the fully supervised CNN in terms of average F1 score by a margin of at least 4\%. It appears that semi-supervised generative models successfully leveraged the additional unlabeled data to learn better representations. Note that the labeled data set comprises land cover classes that are only observed in specific environments, \textit{e.g.} low irradiance conditions. The unlabeled data set, on the other hand, contains a wider variety of irradiance conditions. Second, using $\qphi(y \vert \x)$ to make the predictions, semi-supervised models had small differences in term of F1 score, except for ssInfoGAN. Third, making predictions using $\ppdf(y \vert \x)$, large gains were obtained with the gaussian VAE and p$^3$VAE over the conventional CNN. In particular, p$^3$VAE made significant improvements over the semi-supervised VAE (+3\%), the physics-guided VAE (+9\%) and the CNN (+13\%). A statistical hypothesis test shows that the average F1 score of the semi-supervised VAE and p$^3$VAE are significantly different: we reject the hypothesis of equal average with a $0.3\%$ p-value. Even with exhaustive annotations (we added the labels of the "unlabeled" pixels in the training data set and we refer to this setting as \textit{full annotations} in Tab. \ref{tab1}), p$^3$VAE outperformed the CNN by a 7\% margin. Besides, we can notice that the accuracy for \textit{Sheet metal}, which is the only class with homogoneous irradiance on the training image, is barely the same for every models when $\qphi(y \vert \x)$ is used. Finally, better predictions were made when computing $\ppdf(y \vert \x)$ except for the physics-guided VAE while ssInfoGAN performed poorly with a 3\% lower F1 score than the CNN.

\noindent
\textbf{Latent reflectance spectra} Fig. \ref{fig:reflectance_estimates} shows the latent reflectance spectra computed by p$^3$VAE. There are $\mbox{dim}(\vv) = 4$ estimated spectra per class. p$^3$VAE latent spectra are realistic and are close, in shape, than the ground truth spectra. For instance, the absorption peak of clay at 2.2 $\mu$m is reconstructed by p$^3$VAE for the classes Tile and Sandy loam. However, the intensity of the spectra is not accurate, which is a consequence of the bias in the $\delta_{dir} cos \: \Theta$ prediction. On the contrary, the physics-guided VAE did not produce physically meaningful latent spectra, showing the benefits of the physical prior $f_E$.

\begin{table*}[t]
\begin{center}
\begin{minipage}{\textwidth}
\caption{Mean F1 score per class over 10 runs on the real data}\label{tab:real_data_f1_score}%
%\begin{tabular*}{\textwidth}{@{}ccccccc@{}}
\scalebox{0.82}{
\begin{tabular*}{1.2\textwidth}{@{\extracolsep{\fill}}lcccccccccc@{\extracolsep{\fill}}}
\toprule

& & \multicolumn{8}{@{}c@{}}{\textbf{Classes}} \\ \cmidrule{3-10}%
%\multicolumn{2}{@{}l@{}}{\textbf{Inference model}} & \multirow{2}{*}{Tile}  & \multirow{2}{*}{Asphalt} & \multirow{2}{*}{Vegetation} & \multirow{2}{*}{Painted} & \multirow{2}{*}{Water} & \multirow{2}{*}{Gravels} & \multirow{2}{*}{Metal} & \multirow{2}{*}{Fiber cement} & \multirow{2}{*}{Average} \\ 
%& & & & & sheet metal & & & & \\ 
\multicolumn{2}{@{}l@{}}{\textbf{Inference model}} & Til. & Asp. & Veg. & Sh.M & Wat. & Gra. & Met. & Fib. & Average \\ 
\midrule
CNN & $q_\phi(y \vert x)$ & 0.92 & 0.41 & 0.91 & 0.92 & 0.83 & 0.89 & 0.87 & 0.56 & 0.79 \\ 
& $q_\phi(y \vert x)$ (full annotations)\footnotemark[1]{} & \textit{0.96} & \textit{0.53} & \textit{0.98} & \textit{1.00} & \textit{0.99} & 0.98 & \textit{0.99} & \textit{0.87} & \textit{0.91} \\ \cmidrule{1-11}
FG-Unet & $q_\phi(y \vert x)$ & 0.94 & 0.50 & 0.87 & 0.97 & 0.54 & 0.98 & 0.92 & \textbf{0.83} & 0.82 \\ \cmidrule{1-11}

ssInfoGAN & $q_\phi(y \vert x)$ & 0.86 & 0.34 & 0.78 & 0.95 & 0.40 & 0.94 & 0.91 & 0.76 & 0.74 \\ \cmidrule{1-11}

\multirow{2}{*}{Gaussian VAE} & $q_\phi(y \vert x)$ & 0.94 & 0.27 & 0.88 & 0.96 & 0.83 & 0.92 & \textbf{0.96} & 0.66 & 0.80 \\ 
& $\argmax_y p_\theta(y \vert x)$& 0.82 & 0.40 & 0.86 & 0.73 & 0.84 & 0.60 & 0.79 & 0.63 & 0.71 \\ \cmidrule{1-11}

\multirow{2}{*}{Physics-guided VAE} & $q_\phi(y \vert x)$ & \textbf{0.95} & 0.48 & 0.97 & \textbf{1.00} & 0.98 & 0.98 & 0.87 & 0.78 & 0.88 \\ 
& $\argmax_y p_\theta(y \vert x)$ & 0.92 & 0.48 & 0.97 & 0.99 & \textbf{1.00} & \textbf{0.99} & 0.84 & 0.78 & 0.87 \\ \cmidrule{1-11}

\multirow{2}{*}{p$^3$VAE} & $q_\phi(y \vert x)$ & \textbf{0.95} & \textbf{0.51} & \textbf{0.98} & 0.99 & 0.98 & \textbf{0.99} & \textbf{0.96} & 0.81 & \textbf{0.90} \\
& $\argmax_y p_\theta(y \vert x)$ & 0.93 & 0.49 & 0.97 & 0.91 & 0.99 & 0.79 & 0.90 & 0.78 & 0.84 \\ 
\toprule
\end{tabular*}}

\end{minipage}
\end{center}
\scriptsize{
$^1$ The CNN was optimized with every pixels labeled on the image (\textit{i.e.} the labeled and unlabeled sets), in contrast with every other cases where the image is partially labeled.}
\end{table*}

\noindent
\textbf{Disentanglement} Tab. \ref{tab:disentanglement} shows the mean MIG metric for each factor over 10 runs. $y$ is the class of the pixel; $\delta_{dir}$ and $\Omega$ denotes the local direct and diffuse irradiance conditions, respectively; $\alpha$ is an intra-class mixing coefficient and $\eta$ represents the subclasses configuration (see Appendix \ref{sec:appendix_hyp_dart_data} for detailed definitions). p$^3$VAE outperformed every competing methods, except for the $\Omega$ factor. p$^3$VAE MIG score for the direct irradiance factor is very high compared to others, highlighting the natural disentanglement induced by the physics model $f_E$ that implicitly grounds $z_E$ to $\delta_{dir}$. The low score for the factor $\Omega$ reflects the rough approximation of the diffuse irradiance made by p$^3$VAE. Additional qualitative comparisons in Appendix \ref{sec:qual_dis} suggest that the standard VAE, the physics-guided VAE and the ssInfoGAN suffered from mode collapse.

\subsubsection{Results on real data}

\textbf{Test accuracy} Mean F1 score per class over 10 runs is  showed  on  Tab. \ref{tab:real_data_f1_score}. The CNN with exhaustive annotations outperformed every models, including p$^3$VAE by a margin of 1\%. Excluding the CNN with exhaustive annotations, p$^3$VAE reached a better F1 score than other models (11\%, 8\%, 16\%, 10\% and 2\% higher than the CNN, FG-UNET, ssInfoGAN, the semi-supervised VAE and the physics-guided VAE, respectively, using $\qphi(y \vert \x)$). In contrast with the experiments on the simulated data set, VAE-like models obtained better performances using the predictive distribution $\qphi(y \vert \x)$ rather than the approximation of $\argmax_y \ppdf(y \vert \x)$.

\noindent
\textbf{Land cover maps} Fig. \ref{fig:land_cover} shows land cover maps produced by the semi-supervised VAE and p$^3$VAE. We do not have a labeled ground truth of the image, which moreover includes pixels belonging to classes that are not represented in the training data set. However, we focus on specific areas for which we know the land cover with confidence. Qualitative evaluation shows that p$^3$VAE predicts faithful variations of irradiance over the image, improving its robustness to environmental-based spectral shifts for classification. 

\begin{figure}[t]
	\center
%	\subfloat[]{\includegraphics[height=0.2\textwidth]{ch4_absorption.pdf}}
%	\subfloat[]{	\includegraphics[height=0.195\textwidth]{radiance.pdf}}
	\subfloat[Methane monochromatic absorption]{\includegraphics[width=0.4\textwidth]{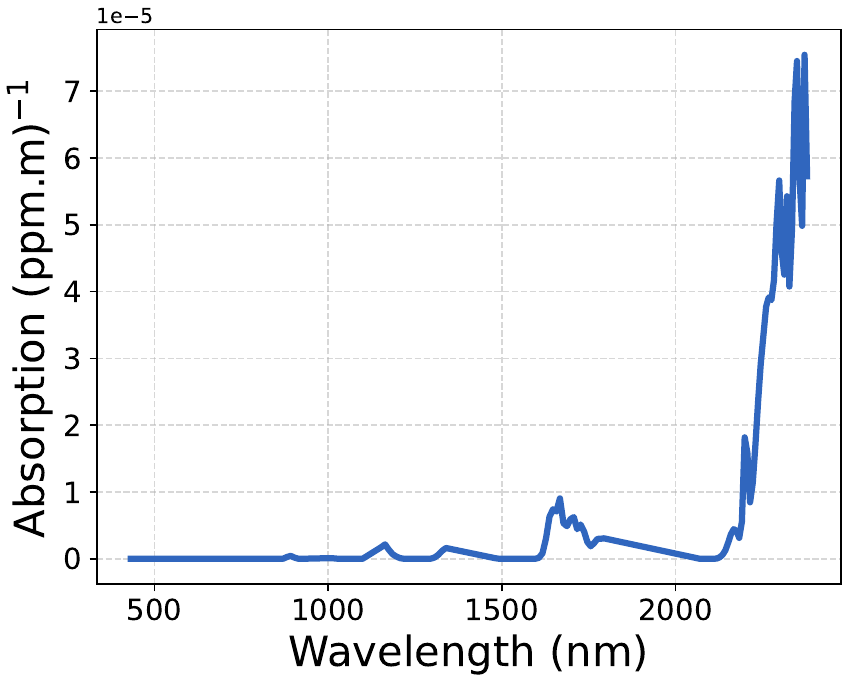}}
	\subfloat[Radiance spectra w/ \& w/o plume]{	\includegraphics[width=0.4\textwidth]{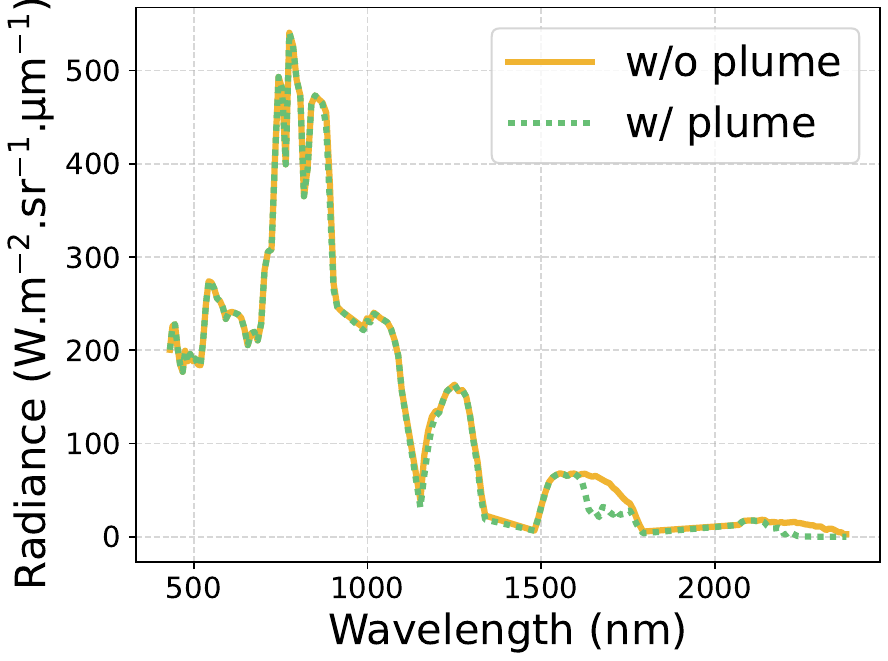}}
	\caption{In the presence of a methane plume (\textit{i.e.} what we call the acquisition conditions), the radiance measured beyond 1600 nm is significantly altered by the plume. \label{fig:ch4_absorption}}
\end{figure}

\subsection{Methane plume inversion}

In this section, we address the inversion of industrial methane plumes from satellite data. Methane is the second strongest anthropogenic greenhouse gas.
Recently, the launch of hyperspectral satellites, such as PRISMA, has sparked interest in global monitoring of industrial methane emissions from space  \cite{nesme2021joint}.
PRISMA indeed covers the visible, near infrared upon to the short-wave infrared domains at a 30 m spatial resolution \cite{labate2009prisma}, covering some of methane absorption bands (cf. Fig \ref{fig:ch4_absorption}). 
Therefore, radiative transfer modeling allows to model how the signal measured by PRISMA would be modified by a methane plume, given methane monochromatic absorption (ppm$^{-1}$.m$^{-1}$) (cf. Fig \ref{fig:ch4_absorption}) and the path-integrated concentration of the plume (ppm.m).
We integrate this physical prior knowledge into p$^3$VAE in order to predict the methane plume concentration. 
In this case, the environmental generative factor of variation is the path-integrated plume concentration, and the intrinsic factor of variation is the land cover. Due to the low spatial resolution of PRISMA, we do not take into account the local irradiance shifts.

\subsubsection{Data set}

The data set comprises PRISMA satellite spectra, acquired over Aude, France, in February, 2020. There are no ground truth data of methane concentration emitted by industries. Therefore, we simulated a methane plume over 85\% of the spectra, with path-integrated concentrations varying from 2E4 ppm.m to 1E5 ppm.m. 
Since ground truth data is not available in practice, we only use the true methane path-integrated concentration for test.
We only assume that we can fairly detect the methane plume, and therefore distinguish pixels that belong to or are outside the plume.
Additional details are provided in Appendix \ref{sec:appendix_methane}.

\subsubsection{Prior physical knowledge $f_E$}

We define the prior physical knowledge $f_E : [0, 1]^B \times \mathbb{R}_+ \longmapsto \mathbb{R}_+^B$ as follows:
\begin{align*}
f_E(\x, z_E) = \x \mbox{ exp} \big(- z_E A_{CH_4} (1 + \frac{1}{\mbox{cos } \Theta})\big)
\end{align*}
where $z_E$ is meant to encode the methane path-integrated concentration, $A_{CH_4}$ is the methane monochromatic absorption (illustrated in Fig. \ref{fig:ch4_absorption}), $\Theta$ is the solar zenith angle, and $\x$ is meant to approximate the radiance observed if there was no methane plume, computed by the neural network $f_I^{\param}$. Additional details about the derivation of $f_E$, as well as on the latent variable modeling, are provided in Appendix \ref{sec:appendix_methane}.

\subsubsection{Competing methods}

\noindent
\textbf{Optimal estimation} \cite{rodgers2000inverse,nesme2021joint} The optimal estimation method \cite{rodgers2000inverse} is a Bayesian framework for inverse problems. We compared to the algorithm introduced in \cite{nesme2021joint}, that applies optimal estimation to methane plume inversion. We could not compare to supervised or semi-supervised machine learning models, since we do not have labeled data.

\subsubsection{Results}

\begin{figure}[t]
	\center
	\subfloat[Optimal estimation \cite{rodgers2000inverse,nesme2021joint}]{\includegraphics[width=0.45\textwidth]{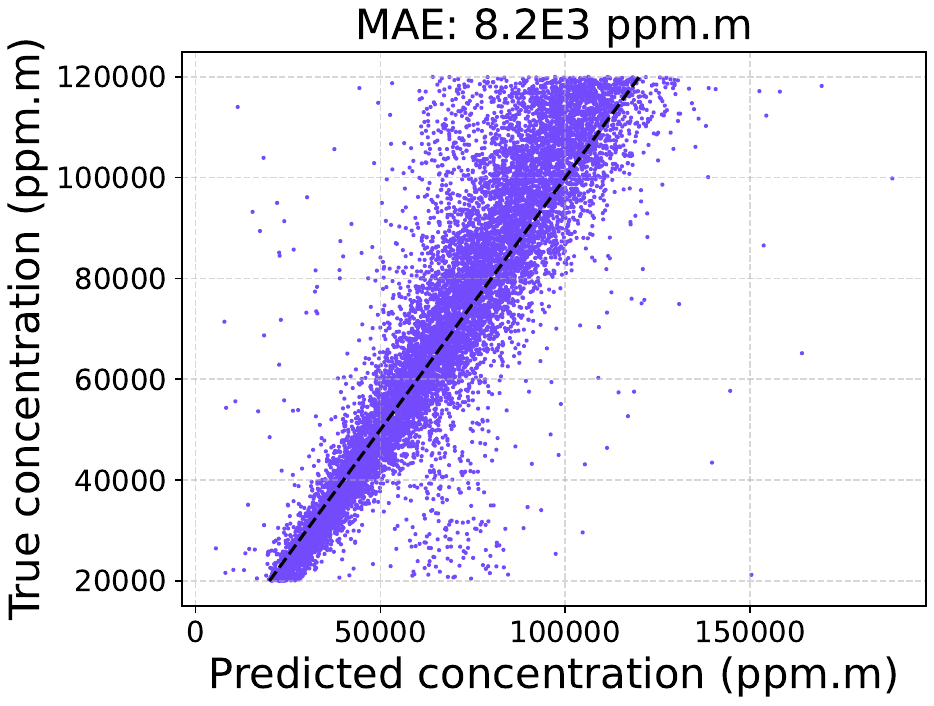}}
	\subfloat[p$^3$VAE]{	\includegraphics[width=0.45\textwidth]{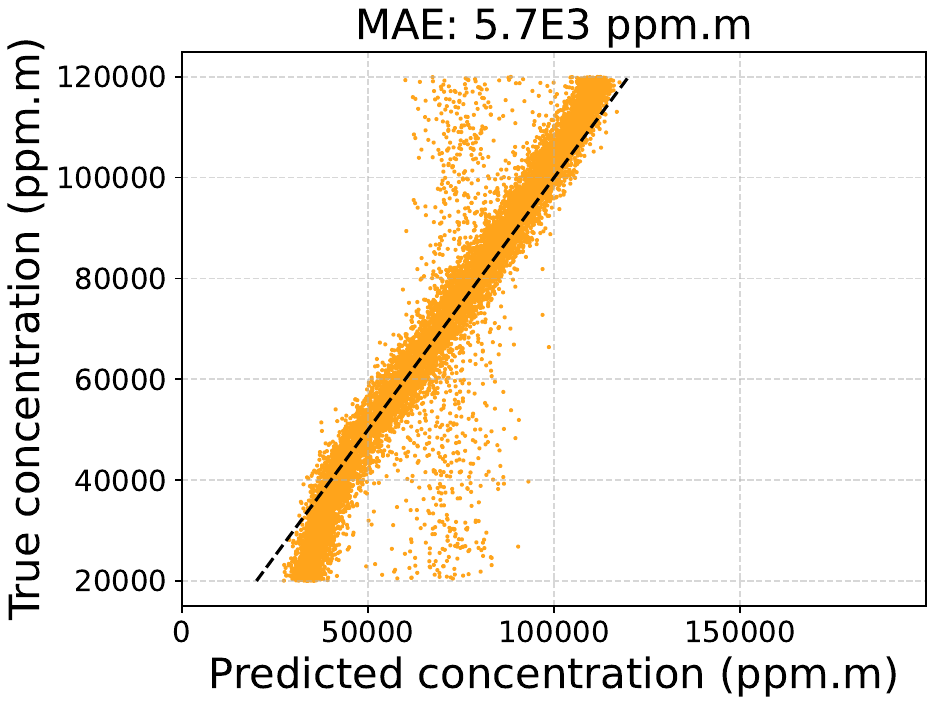}}
	\caption{Predicted against true methane path-integrated concentrations by (a) the optimal estimation method and (b) p$^3$VAE. \label{fig:ch4_pred}}
\end{figure}

Fig. \ref{fig:ch4_pred} shows the predicted against true methane path-integrated concentrations for the optimal estimation (OE) method \cite{rodgers2000inverse,nesme2021joint} and p$^3$VAE. p$^3$VAE reached a lower absolute error (MAE) than the OE algorithm. We believe that the better performance of p$^3$VAE is explained by a better approximation of the background radiance (\textit{i.e.} the signal that would have been measured without the methane plume) than the algorithm used in \cite{nesme2021joint}. Interestingly, we observe that both methods under-estimate the path-integrated concentration for low concentrations, and conversely, over-estimate the path-integrated concentration for high concentrations. Besides, both methods are biased towards the average a priori path-integrated concentration, equal to 60,000 ppm.m.

\subsection{Ablation study}

In this section, we study the impact of the stop-gradient (SG) operator (described in section \ref{sec3}) on the optimization of p$^3$VAE. Tab. \ref{tab:ablation} shows the average F1 score of p$^3$VAE optimized with or without SG on the simulated and real hyperspectral data sets for pixel-wise classification. On simulated data, statistical hypothesis tests show that SG has not a significant influence on the model performance, for both inference techniques. However, a look at the latent reflectance spectra by p$^3$VAE without SG shows that the machine learning part learns unrealistic spectra. For instance, we plot the latent reflectance spectra of the class asphalt by p$^3$VAE without SG on Fig. \ref{fig:no_sg_spectra}. Among the four estimated spectra, one is actually similar to a real asphalt spectrum, one is similar to a vegetation spectrum and others do not correspond to any material in the scene. In contrast, we recall that Fig. \ref{fig:max_estimate} shows that p$^3$VAE with SG learns physically plausible spectra. On real data, statistical hypothesis tests show that SG does have a significant influence on the model performance, for both inference techniques (with a 1.8\% p-value for the inference with $\qphi(y \vert \x)$). 

\begin{minipage}{\textwidth}
\begin{minipage}{0.45\textwidth}
\captionof{table}{p$^3$VAE average F1 score w/ and w/o the stop-gradient operator.}\label{tab:ablation}%
\begin{tabular}{lcc}
\toprule
\textbf{Inference technique} & \textbf{w/o SG}  & \textbf{w/ SG} \\ \cmidrule{1-3}
\multicolumn{3}{c}{Simulated data set} \\ \cmidrule{1-3}
$\qphi(y \vert \x)$ &  0.86 & 0.86 \\
$\ppdf(y \vert \x)$ & 0.92 & 0.93 \\  \cmidrule{1-3}
\multicolumn{3}{c}{Real data set} \\ \cmidrule{1-3}
$\qphi(y \vert \x)$ & 0.88 & 0.90\\
$\ppdf(y \vert \x)$ & 0.73 & 0.84\\ 
\toprule
\end{tabular}
\end{minipage}
\begin{minipage}{0.45\textwidth}
\centering
\includegraphics[width=0.7\textwidth]{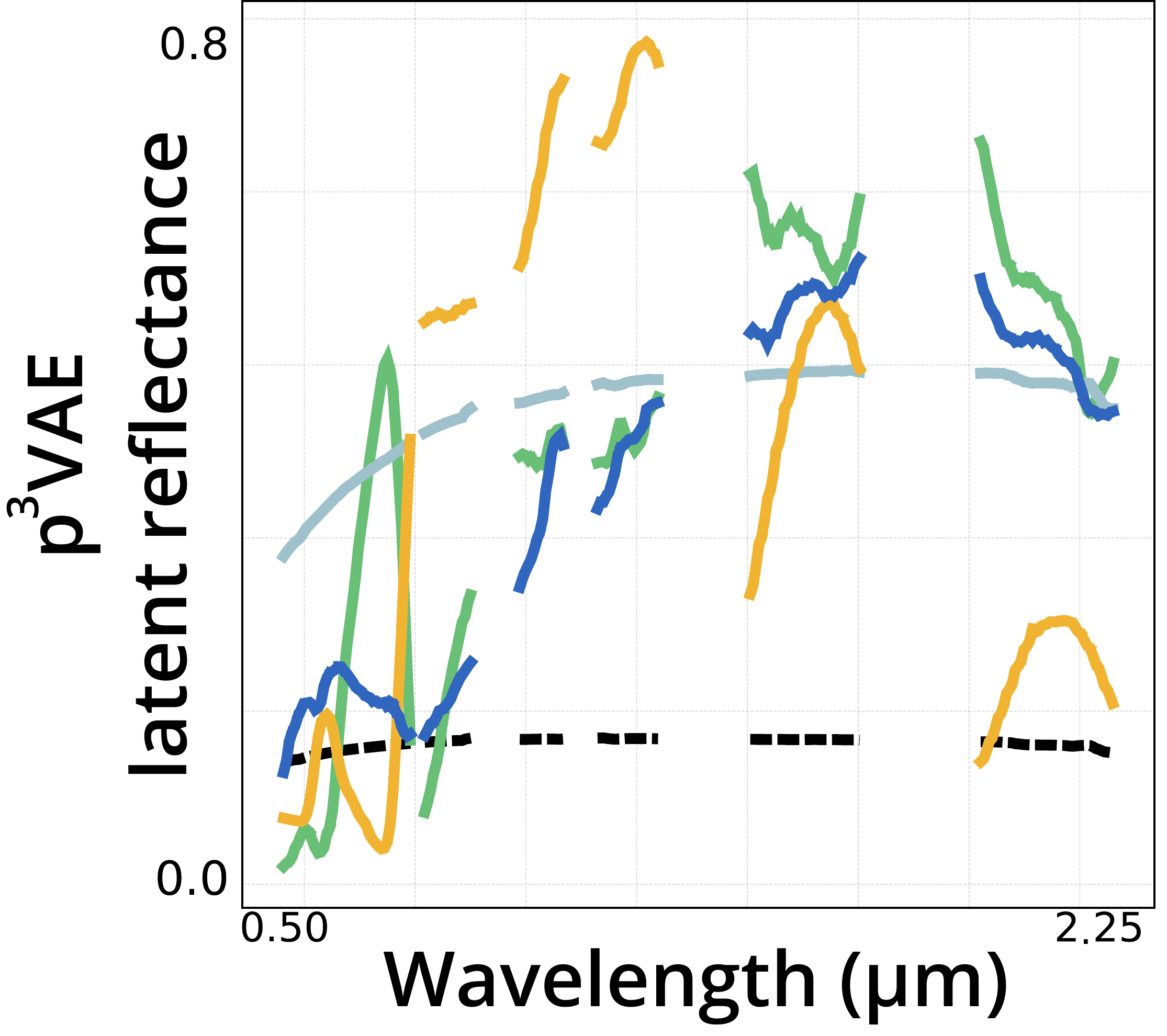}
    \captionof{figure}{Latent reflectance spectra inferred by p$^3$VAE w/o the stop-gradient operator, for the class asphalt. The ground truth spectrum is shown in the dashed dark line. Blue and green spectra inferred by p$^3$VAE are physically improbable and do not correspond to any spectrum of the ground truth. The yellow spectrum corresponds to a  spectrum of vegetation, and only the gray spectrum actually looks like an asphalt spectrum.}
    \label{fig:no_sg_spectra}
\end{minipage}
\end{minipage}

\section{Discussion and conclusion}\label{sec7}

In this work, we introduced p$^3$VAE, a framework that incorporates physical prior knowledge into variational autoencoders. p$^3$VAE is based on three key contributions that led to higher extrapolation performances, disentanglement and interpretability than competing machine learning models. First, prior physical knowledge that models how environmental generative factors modify the data, integrated as non-trainable layers in the deepest layers of the decoder. Second, a semi-supervised training algorithm which strikes a balance between the machine learning part and the physics part of the decoder, and grounds subsets of the latent space to physical variables. Finally, an inference algorithm that explicitly uses the physical part in order to make predictions that are more robust to environmental shifts. 

\noindent
In future work, we would like to address some limitations of p$^3$VAE, including instabilities during optimization.
As a matter of fact, we had more trouble to reach convergence with p$^3$VAE than other models in our experiments. 
In particular, the training loss for some configurations of weights at initialization systematically diverged, irrespective to the learning rate.
A deeper theoretical and empirical understanding of how the physics part contributes to learning may suggest architectural improvements in order to mitigate such pathologies.
We would also like to investigate how transfer learning could improve the robustness of p$^3$VAE to distribution shifts over intrinsic latent variables at test time. Finally, an appealing perspective is the extension of p$^3$VAE to finetune large pretrained machine learning models. We hope that the framework of p$^3$VAE will open up the way towards more robust and interpretable machine learning.

\section*{Acknowledgments} 

We thank Pr. Jean Philippe Gastellu-Etchegorry for his precious help handling the DART \cite{gastellu2012dart} software.

%Bibliography
\bibliographystyle{unsrt}  
\bibliography{references}

\begin{thebibliography}{10}

\bibitem{raissi2019physics}
Maziar Raissi, Paris Perdikaris, and George~E Karniadakis.
\newblock Physics-informed neural networks: A deep learning framework for
  solving forward and inverse problems involving nonlinear partial differential
  equations.
\newblock {\em Journal of Computational physics}, 378:686--707, 2019.

\bibitem{takeishi2021physics}
Naoya Takeishi and Alexandros Kalousis.
\newblock Physics-integrated variational autoencoders for robust and
  interpretable generative modeling.
\newblock {\em Advances in Neural Information Processing Systems},
  34:14809--14821, 2021.

\bibitem{yin2021augmenting}
Yuan Yin, LE~Vincent, DONA J{\'e}r{\'e}mie, Emmanuel de~Bezenac, Ibrahim Ayed,
  Nicolas Thome, and Patrick Gallinari.
\newblock Augmenting physical models with deep networks for complex dynamics
  forecasting.
\newblock {\em International Conference on Learning Representations}, 2021.

\bibitem{zerah2024physics}
Yo{\"e}l Z{\'e}rah, Silvia Valero, and Jordi Inglada.
\newblock Physics-constrained deep learning for biophysical parameter retrieval
  from sentinel-2 images: Inversion of the prosail model.
\newblock {\em Remote Sensing of Environment}, 312:114309, 2024.

\bibitem{mitchell1980need}
Tom~M Mitchell.
\newblock {\em The need for biases in learning generalizations}.
\newblock Department of Computer Science, Laboratory for Computer Science
  Research~…, 1980.

\bibitem{zhao2018bias}
Shengjia Zhao, Hongyu Ren, Arianna Yuan, Jiaming Song, Noah Goodman, and
  Stefano Ermon.
\newblock Bias and generalization in deep generative models: An empirical
  study.
\newblock {\em Advances in Neural Information Processing Systems}, 31, 2018.

\bibitem{jacobsen2021disentangling}
Christian Jacobsen and Karthik Duraisamy.
\newblock Disentangling generative factors of physical fields using variational
  autoencoders.
\newblock {\em arXiv preprint arXiv:2109.07399}, 2021.

\bibitem{wei2020thermodynamic}
C~Wei, J~Zhang, and Chenglin Wu.
\newblock Thermodynamic consistent neural networks for learning material
  interfacial mechanics.
\newblock In {\em NeurIP workshop}, 2020.

\bibitem{trask2022unsupervised}
Nathaniel Trask, Carianne Martinez, Kookjin Lee, and Brad Boyce.
\newblock Unsupervised physics-informed disentanglement of multimodal data for
  high-throughput scientific discovery.
\newblock {\em arXiv preprint arXiv:2202.03242}, 2022.

\bibitem{kingma2014auto}
Diederik~P. Kingma and Max Welling.
\newblock Auto-encoding variational bayes.
\newblock In Yoshua Bengio and Yann LeCun, editors, {\em 2nd International
  Conference on Learning Representations, {ICLR} 2014, Banff, AB, Canada, April
  14-16, 2014, Conference Track Proceedings}, 2014.

\bibitem{higgins2016early}
Irina Higgins, Loic Matthey, Xavier Glorot, Arka Pal, Benigno Uria, Charles
  Blundell, Shakir Mohamed, and Alexander Lerchner.
\newblock Early visual concept learning with unsupervised deep learning.
\newblock {\em stat}, 1050:20, 2016.

\bibitem{higgins2016beta}
Irina Higgins, Loic Matthey, Arka Pal, Christopher Burgess, Xavier Glorot,
  Matthew Botvinick, Shakir Mohamed, and Alexander Lerchner.
\newblock beta-vae: Learning basic visual concepts with a constrained
  variational framework.
\newblock 2016.

\bibitem{figurnov2018implicit}
Mikhail Figurnov, Shakir Mohamed, and Andriy Mnih.
\newblock Implicit reparameterization gradients.
\newblock {\em Advances in neural information processing systems}, 31, 2018.

\bibitem{joo2020dirichlet}
Weonyoung Joo, Wonsung Lee, Sungrae Park, and Il-Chul Moon.
\newblock Dirichlet variational autoencoder.
\newblock {\em Pattern Recognition}, 107:107514, 2020.

\bibitem{xiao2018dirichlet}
Yijun Xiao, Tiancheng Zhao, and William~Yang Wang.
\newblock Dirichlet variational autoencoder for text modeling.
\newblock {\em arXiv preprint arXiv:1811.00135}, 2018.

\bibitem{dupont2018learning}
Emilien Dupont.
\newblock Learning disentangled joint continuous and discrete representations.
\newblock {\em Advances in Neural Information Processing Systems}, 31, 2018.

\bibitem{aragon2020self}
Miguel~A Aragon-Calvo and JC~Carvajal.
\newblock Self-supervised learning with physics-aware neural networks--i.
  galaxy model fitting.
\newblock {\em Monthly Notices of the Royal Astronomical Society},
  498(3):3713--3719, 2020.

\bibitem{kingma2014semi}
Durk~P Kingma, Shakir Mohamed, Danilo Jimenez~Rezende, and Max Welling.
\newblock Semi-supervised learning with deep generative models.
\newblock {\em Advances in neural information processing systems}, 27, 2014.

\bibitem{chen2020physics}
Chacha Chen, Guanjie Zheng, Hua Wei, and Zhenhui Li.
\newblock Physics-informed generative adversarial networks for sequence
  generation with limited data.
\newblock In {\em NeurIPS Workshop on Interpretable Inductive Biases and
  Physically Structured Learning}, 2020.

\bibitem{wang2021understanding}
Sifan Wang, Yujun Teng, and Paris Perdikaris.
\newblock Understanding and mitigating gradient flow pathologies in
  physics-informed neural networks.
\newblock {\em SIAM Journal on Scientific Computing}, 43(5):A3055--A3081, 2021.

\bibitem{yildiz2019ode2vae}
Cagatay Yildiz, Markus Heinonen, and Harri Lahdesmaki.
\newblock Ode2vae: Deep generative second order odes with bayesian neural
  networks.
\newblock {\em Advances in Neural Information Processing Systems}, 32, 2019.

\bibitem{linial2021generative}
Ori Linial, Neta Ravid, Danny Eytan, and Uri Shalit.
\newblock Generative ode modeling with known unknowns.
\newblock In {\em Proceedings of the Conference on Health, Inference, and
  Learning}, pages 79--94, 2021.

\bibitem{kulkarni2015deep}
Tejas~D Kulkarni, William~F Whitney, Pushmeet Kohli, and Josh Tenenbaum.
\newblock Deep convolutional inverse graphics network.
\newblock {\em Advances in neural information processing systems}, 28, 2015.

\bibitem{ding2020guided}
Zheng Ding, Yifan Xu, Weijian Xu, Gaurav Parmar, Yang Yang, Max Welling, and
  Zhuowen Tu.
\newblock Guided variational autoencoder for disentanglement learning.
\newblock In {\em Proceedings of the IEEE/CVF Conference on Computer Vision and
  Pattern Recognition}, pages 7920--7929, 2020.

\bibitem{rodriguez2021disentanglement}
Elliott~Gordon Rodriguez.
\newblock On disentanglement and mutual information in semi-supervised
  variational auto-encoders.
\newblock In {\em Proceedings of the IEEE/CVF Conference on Computer Vision and
  Pattern Recognition}, pages 1257--1262, 2021.

\bibitem{chen2018isolating}
Ricky~TQ Chen, Xuechen Li, Roger~B Grosse, and David~K Duvenaud.
\newblock Isolating sources of disentanglement in variational autoencoders.
\newblock {\em Advances in neural information processing systems}, 31, 2018.

\bibitem{lecun-mnisthandwrittendigit-2010}
Yann LeCun and Corinna Cortes.
\newblock {MNIST} handwritten digit database.
\newblock 2010.

\bibitem{locatello2019challenging}
Francesco Locatello, Stefan Bauer, Mario Lucic, Gunnar Raetsch, Sylvain Gelly,
  Bernhard Sch{\"o}lkopf, and Olivier Bachem.
\newblock Challenging common assumptions in the unsupervised learning of
  disentangled representations.
\newblock In {\em international conference on machine learning}, pages
  4114--4124. PMLR, 2019.

\bibitem{van2020survey}
Jesper~E Van~Engelen and Holger~H Hoos.
\newblock A survey on semi-supervised learning.
\newblock {\em Machine Learning}, 109(2):373--440, 2020.

\bibitem{triguero2015self}
Isaac Triguero, Salvador Garc{\'\i}a, and Francisco Herrera.
\newblock Self-labeled techniques for semi-supervised learning: taxonomy,
  software and empirical study.
\newblock {\em Knowledge and Information systems}, 42(2):245--284, 2015.

\bibitem{salah2011contractive}
Rifai Salah, P~Vincent, X~Muller, et~al.
\newblock Contractive auto-encoders: Explicit invariance during feature
  extraction.
\newblock In {\em Proc. of the 28th International Conference on Machine
  Learning}, pages 833--840, 2011.

\bibitem{erhan2010does}
Dumitru Erhan, Aaron Courville, Yoshua Bengio, and Pascal Vincent.
\newblock Why does unsupervised pre-training help deep learning?
\newblock In {\em Proceedings of the thirteenth international conference on
  artificial intelligence and statistics}, pages 201--208. JMLR Workshop and
  Conference Proceedings, 2010.

\bibitem{ranzato2008semi}
Marc'Aurelio Ranzato and Martin Szummer.
\newblock Semi-supervised learning of compact document representations with
  deep networks.
\newblock In {\em Proceedings of the 25th international conference on Machine
  learning}, pages 792--799, 2008.

\bibitem{weston2012deep}
Jason Weston, Fr{\'e}d{\'e}ric Ratle, Hossein Mobahi, and Ronan Collobert.
\newblock Deep learning via semi-supervised embedding.
\newblock In {\em Neural networks: Tricks of the trade}, pages 639--655.
  Springer, 2012.

\bibitem{rifai2011higher}
Salah Rifai, Gr{\'e}goire Mesnil, Pascal Vincent, Xavier Muller, Yoshua Bengio,
  Yann Dauphin, and Xavier Glorot.
\newblock Higher order contractive auto-encoder.
\newblock In {\em Joint European conference on machine learning and knowledge
  discovery in databases}, pages 645--660. Springer, 2011.

\bibitem{miyato2018virtual}
Takeru Miyato, Shin-ichi Maeda, Masanori Koyama, and Shin Ishii.
\newblock Virtual adversarial training: a regularization method for supervised
  and semi-supervised learning.
\newblock {\em IEEE transactions on pattern analysis and machine intelligence},
  41(8):1979--1993, 2018.

\bibitem{castillo2021semi}
Javiera Castillo-Navarro, Bertrand Le~Saux, Alexandre Boulch, Nicolas Audebert,
  and S{\'e}bastien Lef{\`e}vre.
\newblock Semi-supervised semantic segmentation in earth observation: The
  minifrance suite, dataset analysis and multi-task network study.
\newblock {\em Machine Learning}, pages 1--36, 2021.

\bibitem{goodfellow2014generative}
Ian Goodfellow, Jean Pouget-Abadie, Mehdi Mirza, Bing Xu, David Warde-Farley,
  Sherjil Ozair, Aaron Courville, and Yoshua Bengio.
\newblock Generative adversarial nets.
\newblock {\em Advances in neural information processing systems}, 27, 2014.

\bibitem{rezende2015variational}
Danilo Rezende and Shakir Mohamed.
\newblock Variational inference with normalizing flows.
\newblock In {\em International conference on machine learning}, pages
  1530--1538. PMLR, 2015.

\bibitem{spurr2017guiding}
Adrian Spurr, Emre Aksan, and Otmar Hilliges.
\newblock Guiding infogan with semi-supervision.
\newblock In {\em Joint European Conference on Machine Learning and Knowledge
  Discovery in Databases}, pages 119--134. Springer, 2017.

\bibitem{chen2016infogan}
Xi~Chen, Yan Duan, Rein Houthooft, John Schulman, Ilya Sutskever, and Pieter
  Abbeel.
\newblock Infogan: Interpretable representation learning by information
  maximizing generative adversarial nets.
\newblock {\em Advances in neural information processing systems}, 29, 2016.

\bibitem{izmailov2020semi}
Pavel Izmailov, Polina Kirichenko, Marc Finzi, and Andrew~Gordon Wilson.
\newblock Semi-supervised learning with normalizing flows.
\newblock In {\em International Conference on Machine Learning}, pages
  4615--4630. PMLR, 2020.

\bibitem{tothhamiltonian}
Peter Toth, Danilo~J Rezende, Andrew Jaegle, S{\'e}bastien Racani{\`e}re,
  Aleksandar Botev, and Irina Higgins.
\newblock Hamiltonian generative networks.
\newblock In {\em International Conference on Learning Representations}, 2020.

\bibitem{rezende2014stochastic}
Danilo~Jimenez Rezende, Shakir Mohamed, and Daan Wierstra.
\newblock Stochastic backpropagation and approximate inference in deep
  generative models.
\newblock In {\em International conference on machine learning}, pages
  1278--1286. PMLR, 2014.

\bibitem{NEURIPS2018_1ee3dfcd}
Ricky T.~Q. Chen, Xuechen Li, Roger~B Grosse, and David~K Duvenaud.
\newblock Isolating sources of disentanglement in variational autoencoders.
\newblock In S.~Bengio, H.~Wallach, H.~Larochelle, K.~Grauman, N.~Cesa-Bianchi,
  and R.~Garnett, editors, {\em Advances in Neural Information Processing
  Systems}, volume~31. Curran Associates, Inc., 2018.

\bibitem{kingma2014adam}
Diederik~P Kingma and Jimmy Ba.
\newblock Adam: A method for stochastic optimization.
\newblock {\em arXiv preprint arXiv:1412.6980}, 2014.

\bibitem{gastellu2012dart}
Jean-Philippe Gastellu-Etchegorry, Eloi Grau, and Nicolas Lauret.
\newblock Dart: A 3d model for remote sensing images and radiative budget of
  earth surfaces.
\newblock {\em Modeling and simulation in engineering}, (2), 2012.

\bibitem{ROUPIOZ2023109109}
L.~Roupioz, X.~Briottet, K.~Adeline, A.~{Al Bitar}, D.~Barbon-Dubosc,
  R.~Barda-Chatain, P.~Barillot, S.~Bridier, E.~Carroll, C.~Cassante,
  A.~Cerbelaud, P.~Déliot, P.~Doublet, P.E. Dupouy, S.~Gadal, S.~Guernouti,
  A.~{De Guilhem De Lataillade}, A.~Lemonsu, R.~Llorens, R.~Luhahe, A.~Michel,
  A.~Moussous, M.~Musy, F.~Nerry, L.~Poutier, A.~Rodler, N.~Riviere,
  T.~Riviere, J.L. Roujean, A.~Roy, A.~Schilling, D.~Skokovic, and J.~Sobrino.
\newblock Multi-source datasets acquired over toulouse (france) in 2021 for
  urban microclimate studies during the camcatt/ai4geo field campaign.
\newblock {\em Data in Brief}, 48:109109, 2023.

\bibitem{miesch2005direct}
C.~{Miesch}, L.~{Poutier}, V.~{Achard}, X.~{Briottet}, X.~{Lenot}, and
  Y.~{Boucher}.
\newblock Direct and inverse radiative transfer solutions for visible and
  near-infrared hyperspectral imagery.
\newblock {\em IEEE Transactions on Geoscience and Remote Sensing},
  43(7):1552--1562, 2005.

\bibitem{audebert2019deep}
Nicolas Audebert, Bertrand Le~Saux, and S{\'e}bastien Lef{\`e}vre.
\newblock Deep learning for classification of hyperspectral data: A comparative
  review.
\newblock {\em IEEE geoscience and remote sensing magazine}, 7(2):159--173,
  2019.

\bibitem{stoian2019land}
Andrei Stoian, Vincent Poulain, Jordi Inglada, Victor Poughon, and Dawa
  Derksen.
\newblock Land cover maps production with high resolution satellite image time
  series and convolutional neural networks: Adaptations and limits for
  operational systems.
\newblock {\em Remote Sensing}, 11(17):1986, 2019.

\bibitem{ronneberger2015u}
Olaf Ronneberger, Philipp Fischer, and Thomas Brox.
\newblock U-net: Convolutional networks for biomedical image segmentation.
\newblock pages 234--241, 2015.

\bibitem{arjovsky2017wasserstein}
Martin Arjovsky, Soumith Chintala, and L{\'e}on Bottou.
\newblock Wasserstein generative adversarial networks.
\newblock pages 214--223, 2017.

\bibitem{gulrajani2017improved}
Ishaan Gulrajani, Faruk Ahmed, Martin Arjovsky, Vincent Dumoulin, and Aaron~C
  Courville.
\newblock Improved training of wasserstein gans.
\newblock {\em Advances in neural information processing systems}, 30, 2017.

\bibitem{nesme2021joint}
Nicolas Nesme, Rodolphe Marion, Olivier Lezeaux, St{\'e}phanie Doz, Claude
  Camy-Peyret, and Pierre-Yves Foucher.
\newblock Joint use of in-scene background radiance estimation and optimal
  estimation methods for quantifying methane emissions using prisma
  hyperspectral satellite data: Application to the korpezhe industrial site.
\newblock {\em Remote Sensing}, 13(24):4992, 2021.

\bibitem{labate2009prisma}
Demetrio Labate, Massimo Ceccherini, Andrea Cisbani, Vittorio De~Cosmo, Claudio
  Galeazzi, Lorenzo Giunti, Mauro Melozzi, Stefano Pieraccini, and Moreno
  Stagi.
\newblock The prisma payload optomechanical design, a high performance
  instrument for a new hyperspectral mission.
\newblock {\em Acta Astronautica}, 65(9-10):1429--1436, 2009.

\bibitem{rodgers2000inverse}
Clive~D Rodgers.
\newblock {\em Inverse methods for atmospheric sounding: theory and practice},
  volume~2.
\newblock World scientific, 2000.

\end{thebibliography}

\clearpage
\begin{appendices}
\begin{minipage}{\textwidth}
\Large{\textbf{Appendix}} 
\end{minipage}

\section{Damped pendulum \label{sec:appendix_pendulum}}

\subsection{Derivation of the physical prior knowledge}

In this section, we explain how we modeled $f_E$ as an exponential decrease. Let us assume that following ODE models the pendulum dynamics:
\begin{align}
	\frac{d^2\vartheta}{dt^2}(t) + \xi \frac{d\vartheta}{dt}(t) + f_I(\vartheta(t) ; \z_I) = 0 \label{eq:ode_app}
\end{align}
where $f_I$ is unknown. We further assume that the pendulum is at equilibrium at $(\vartheta, \dot{\vartheta}) = (0, 0)$, and that $\int_0^\vartheta f_I(u; \z_I) du > 0$ if $\vartheta \neq 0$. Under those assumptions, we are going to show that the system is stable by defining a proper Lyapunov function:
\begin{align}
	V : & [-\frac{\pi}{2}, \frac{\pi}{2}] \times \mathbb{R} \longrightarrow \mathbb{R} \\
	& (\vartheta, \dot{\vartheta}) \longmapsto \frac{1}{2} \dot{\vartheta}^2 + \int_0^\vartheta f_I(u; \z_I) du % + \epsilon \vartheta \dot{\vartheta}
\end{align}
%where $\epsilon$ is a small constant such that $V(\vartheta, \dot{\vartheta}) > 0 \mbox{ for all } (\vartheta, \dot{\vartheta}) \neq (0, 0)$. 
We verify that:
\begin{align*}
	\begin{cases}
	V(0, 0) & = 0 \\
	V(\vartheta, \dot{\vartheta}) & > 0 \mbox{ for all } (\vartheta, \dot{\vartheta}) \neq (0, 0) \\
	\dot{V}(\vartheta, \dot{\vartheta}) & = - \xi \dot{\vartheta}^2 \leq 0 \mbox{ for all } (\vartheta, \dot{\vartheta})
	\end{cases}
\end{align*}
since $\xi > 0$. Therefore, from the Lyapunov's direct method and Lasalle's
invariance principle, the solution of eq. \ref{eq:ode_app} is stable and asymptotically converges towards $(0, 0)$, at a rate characterized by $\xi$. Finally, we approximate the convergence of the pendulum towards the equilibrium by an exponential decrease.

\subsection{Experimental results}

\begin{figure*}[ht]
    \centering
    
    \subfloat{\includegraphics[height=0.245\textwidth]{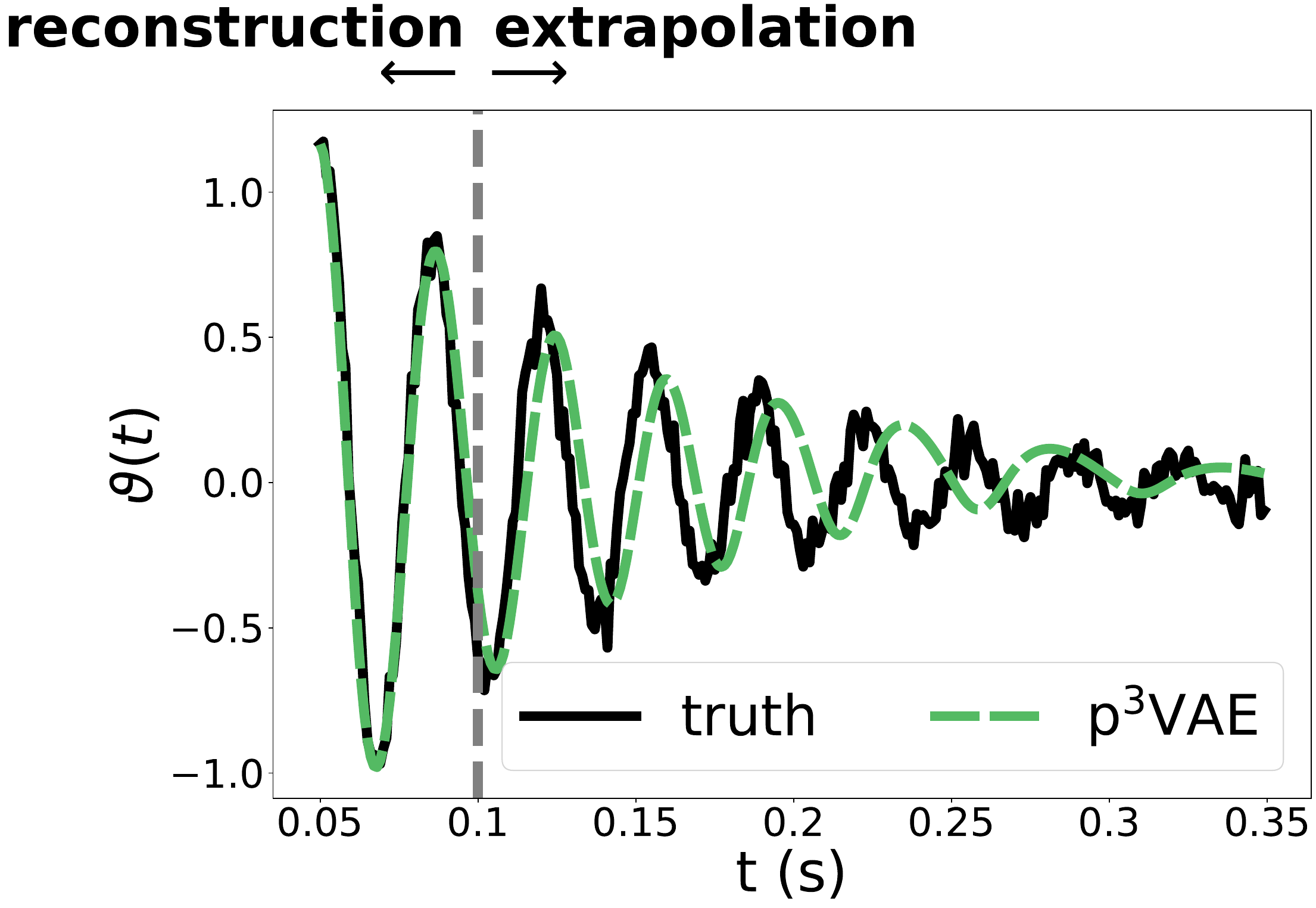}}
    \subfloat{\includegraphics[height=0.215\textwidth]{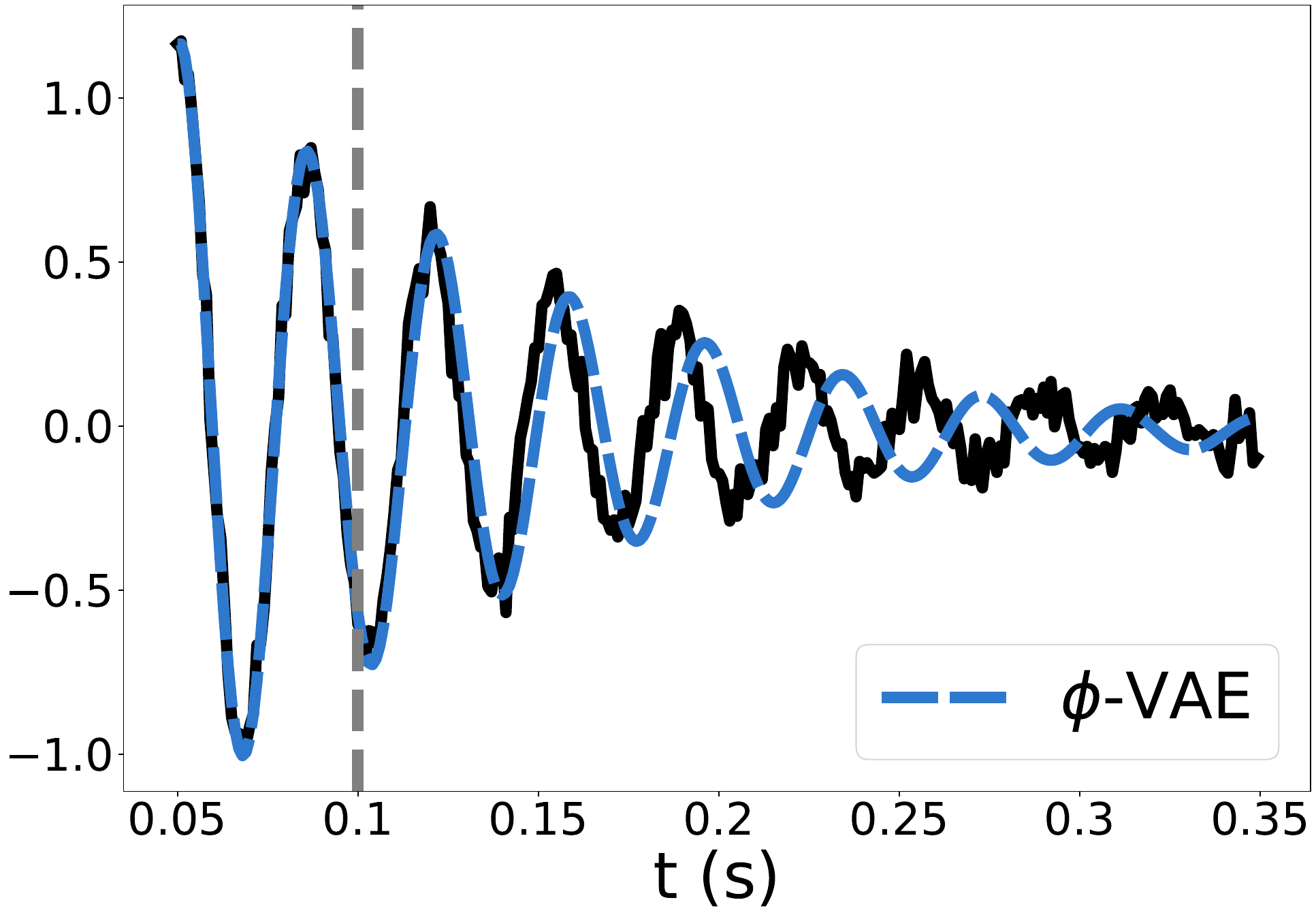}}
    \subfloat{\includegraphics[height=0.215\textwidth]{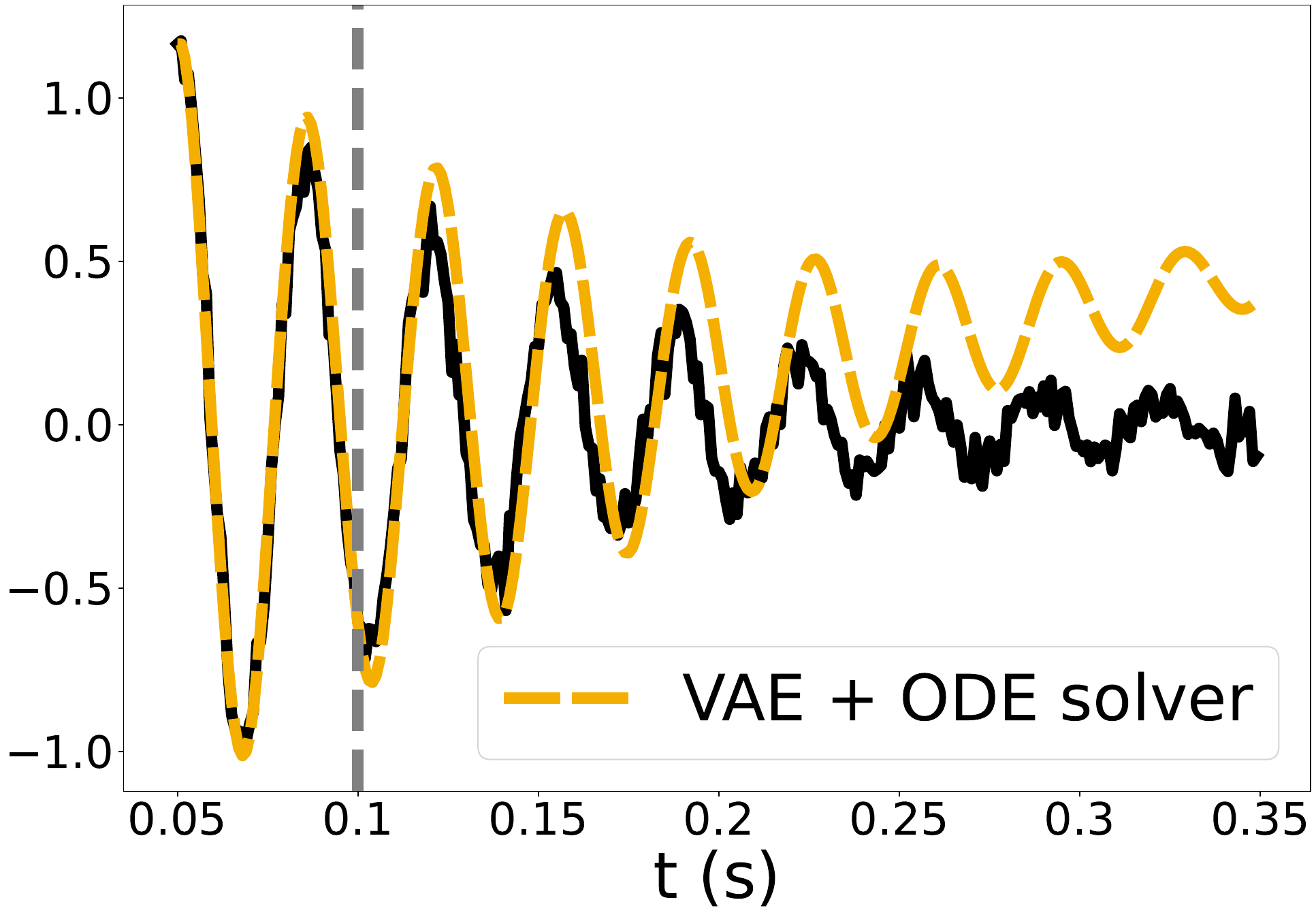}}
    
    \subfloat{\includegraphics[height=0.215\textwidth]{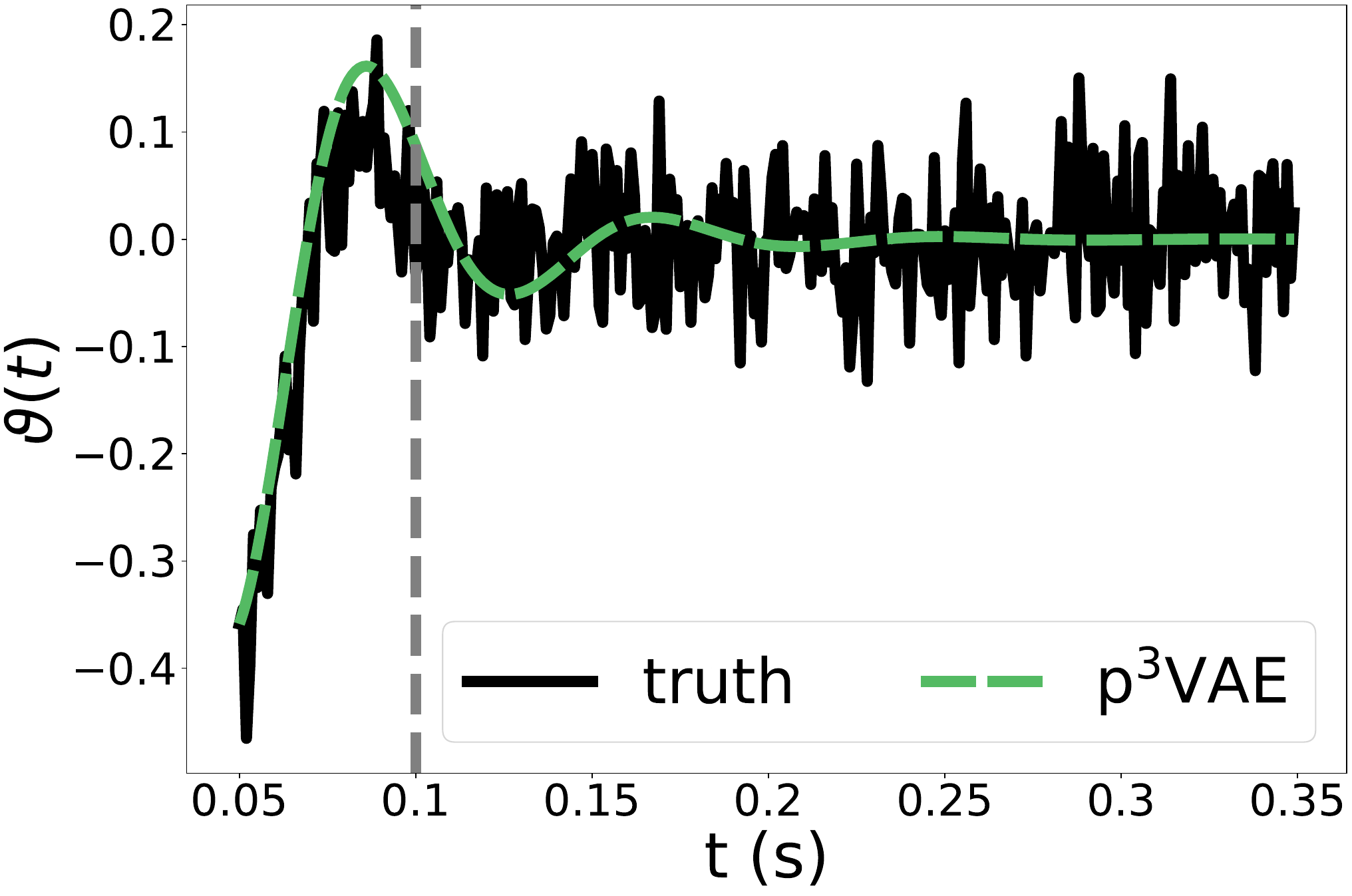}}
    \subfloat{\includegraphics[height=0.215\textwidth]{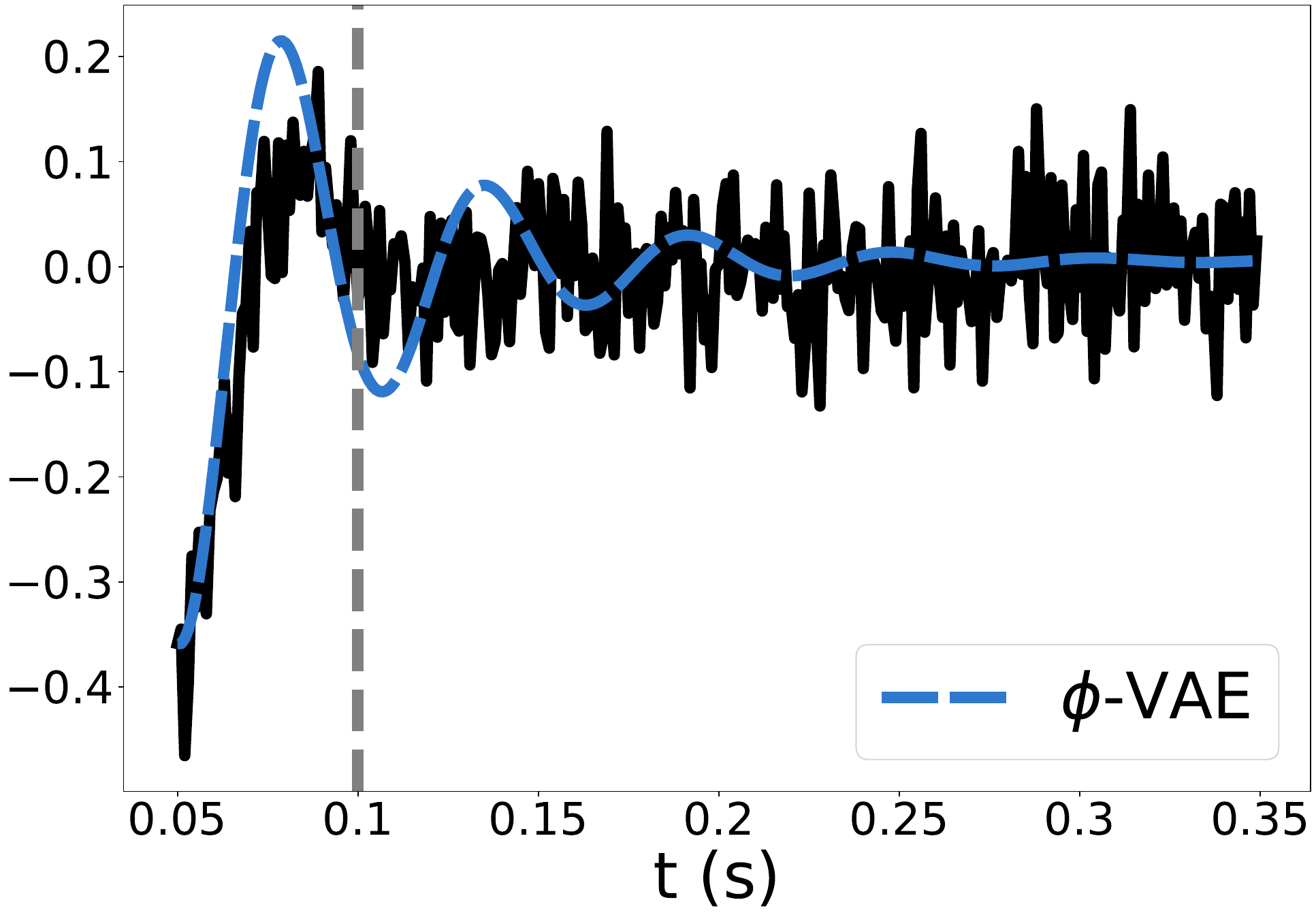}}
    \subfloat{\includegraphics[height=0.215\textwidth]{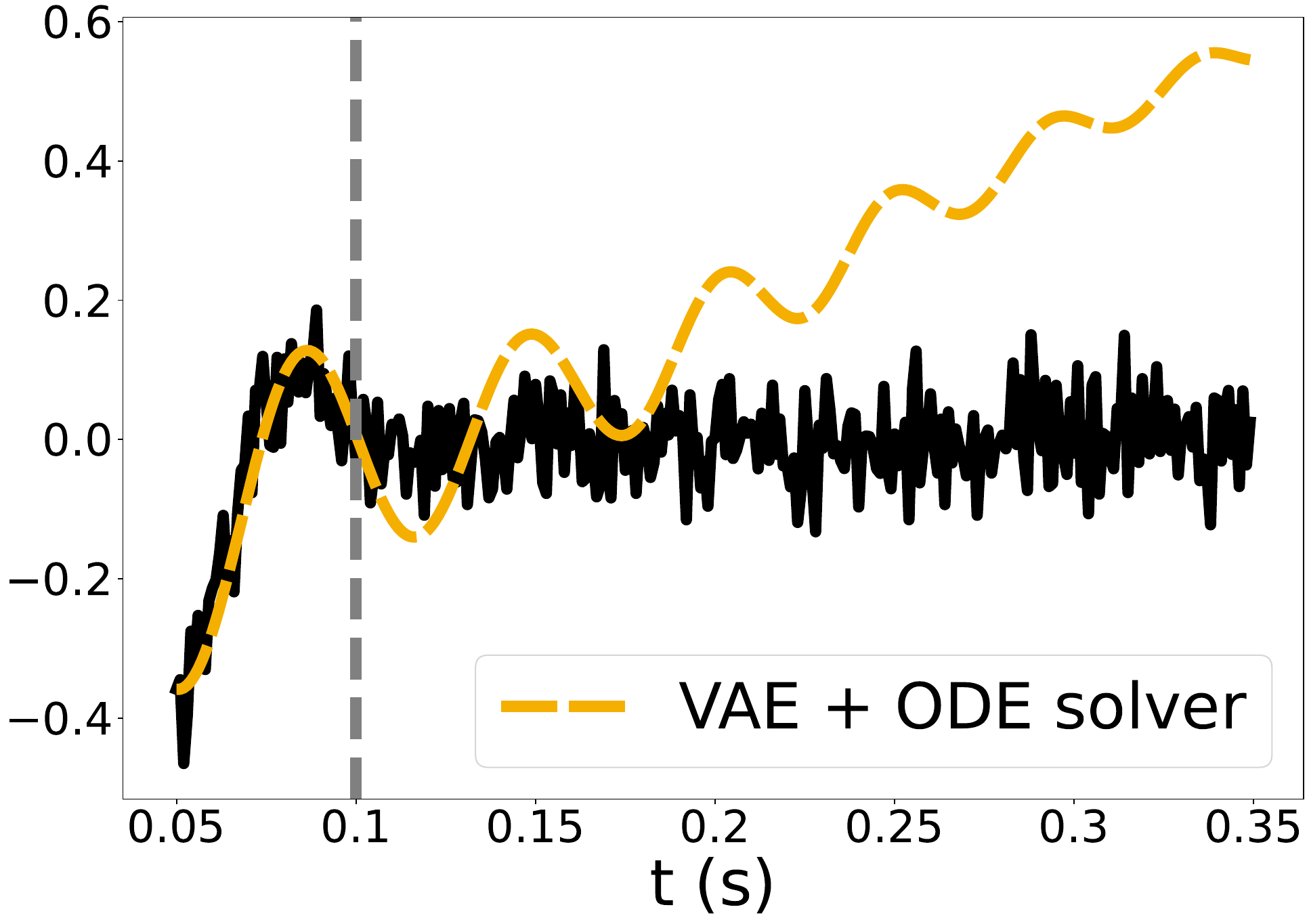}}
    
    \subfloat{\includegraphics[height=0.215\textwidth]{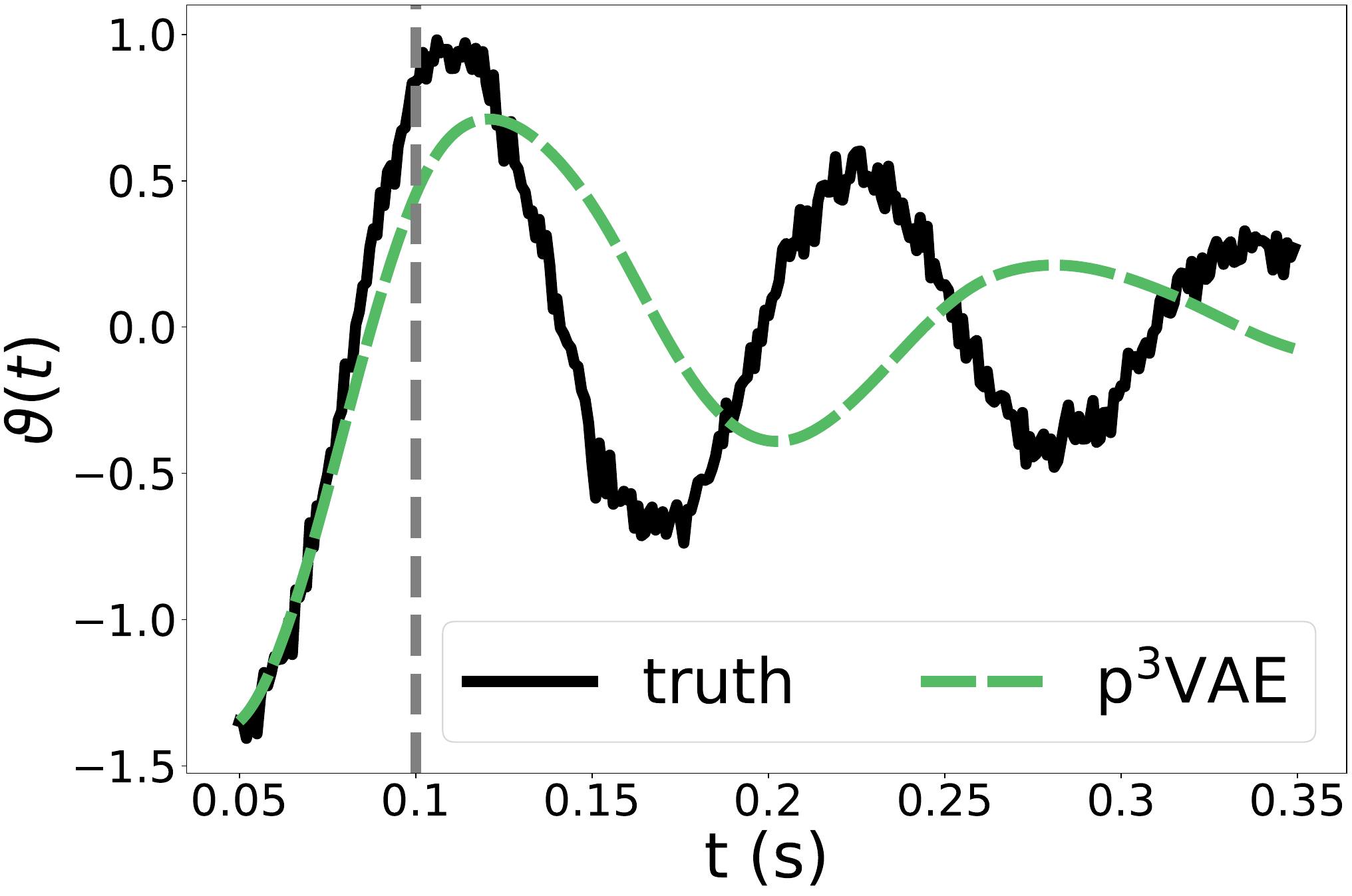}}
    \subfloat{\includegraphics[height=0.215\textwidth]{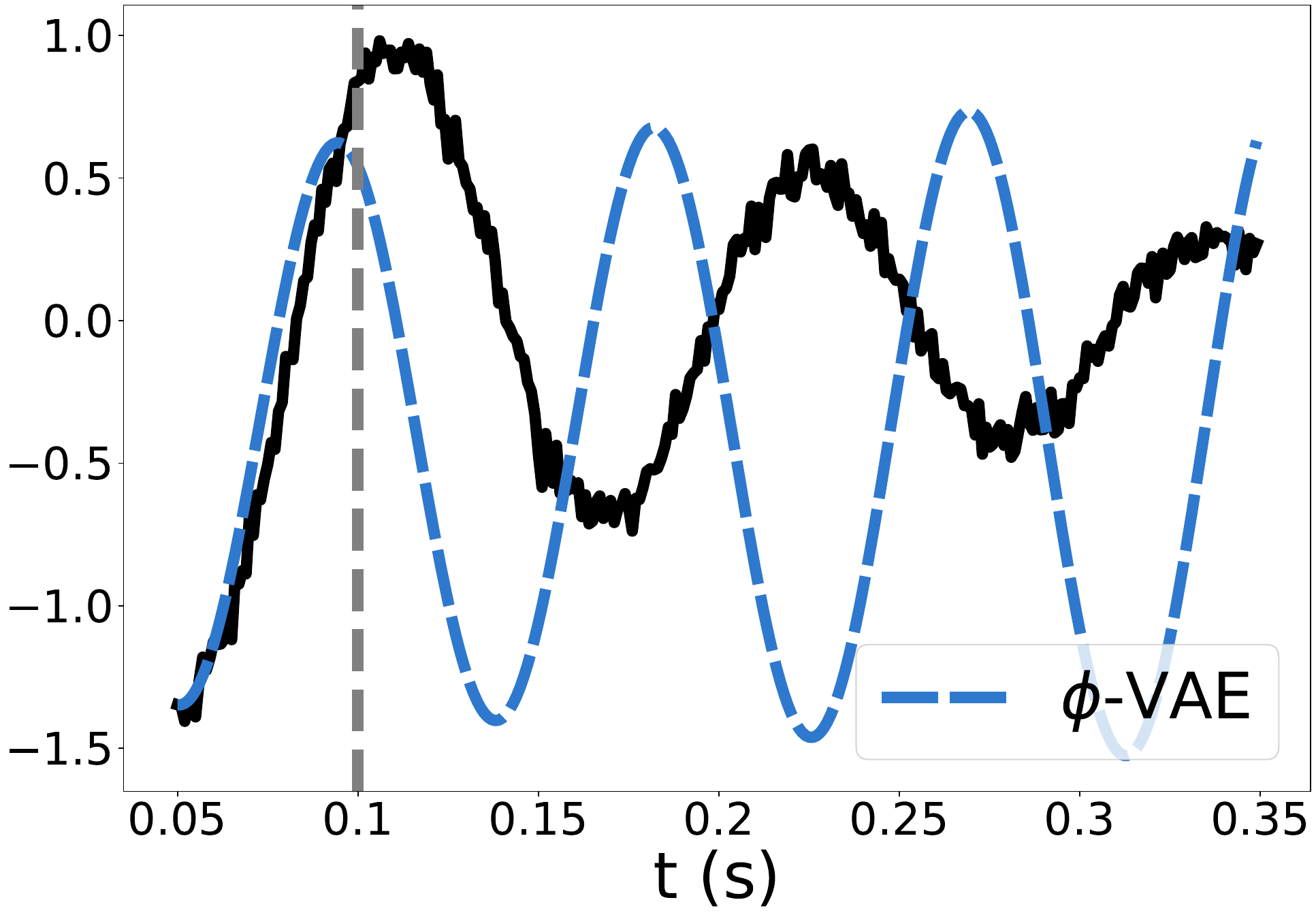}}
    \subfloat{\includegraphics[height=0.215\textwidth]{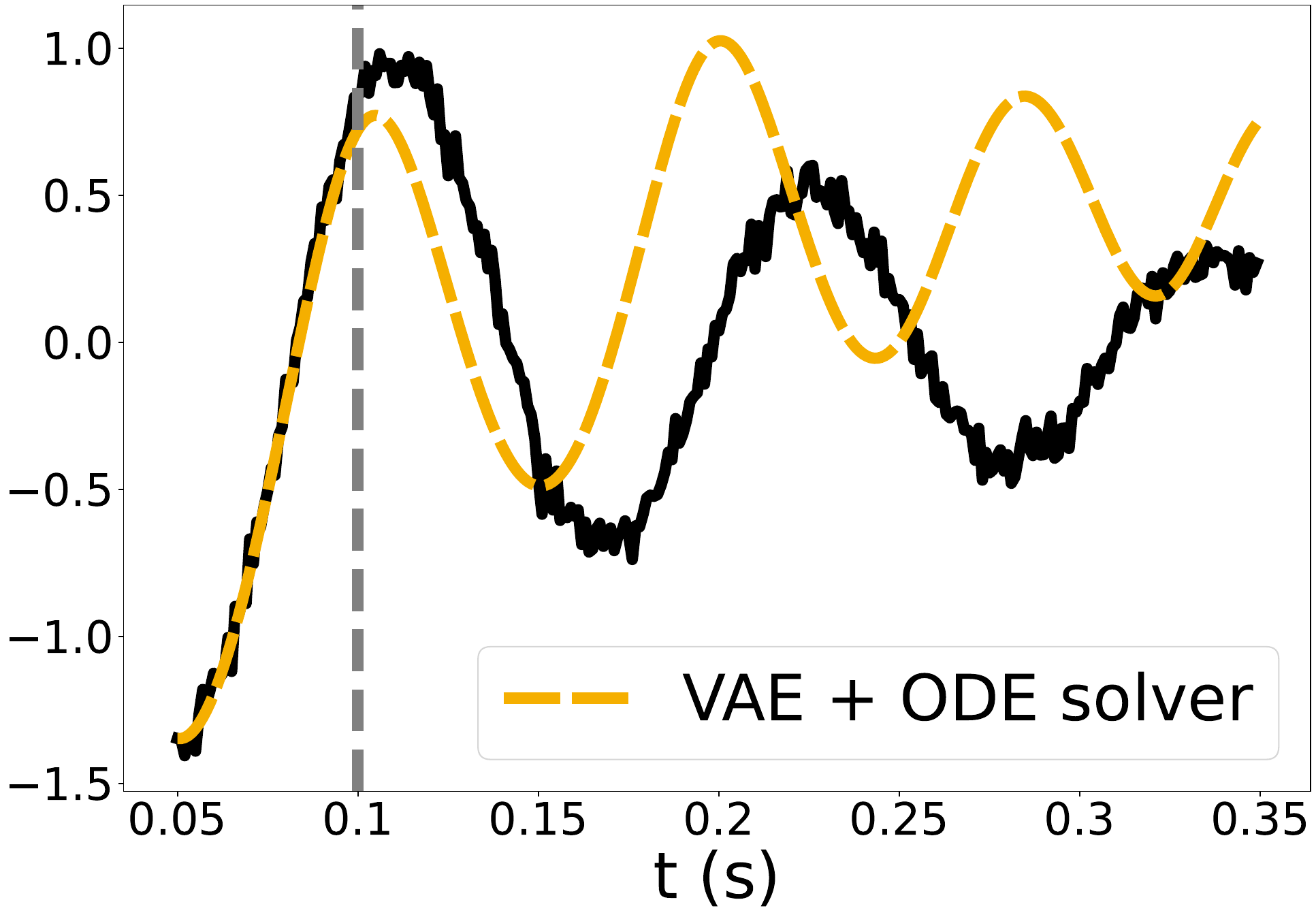}}
        
    \subfloat{\includegraphics[height=0.215\textwidth]{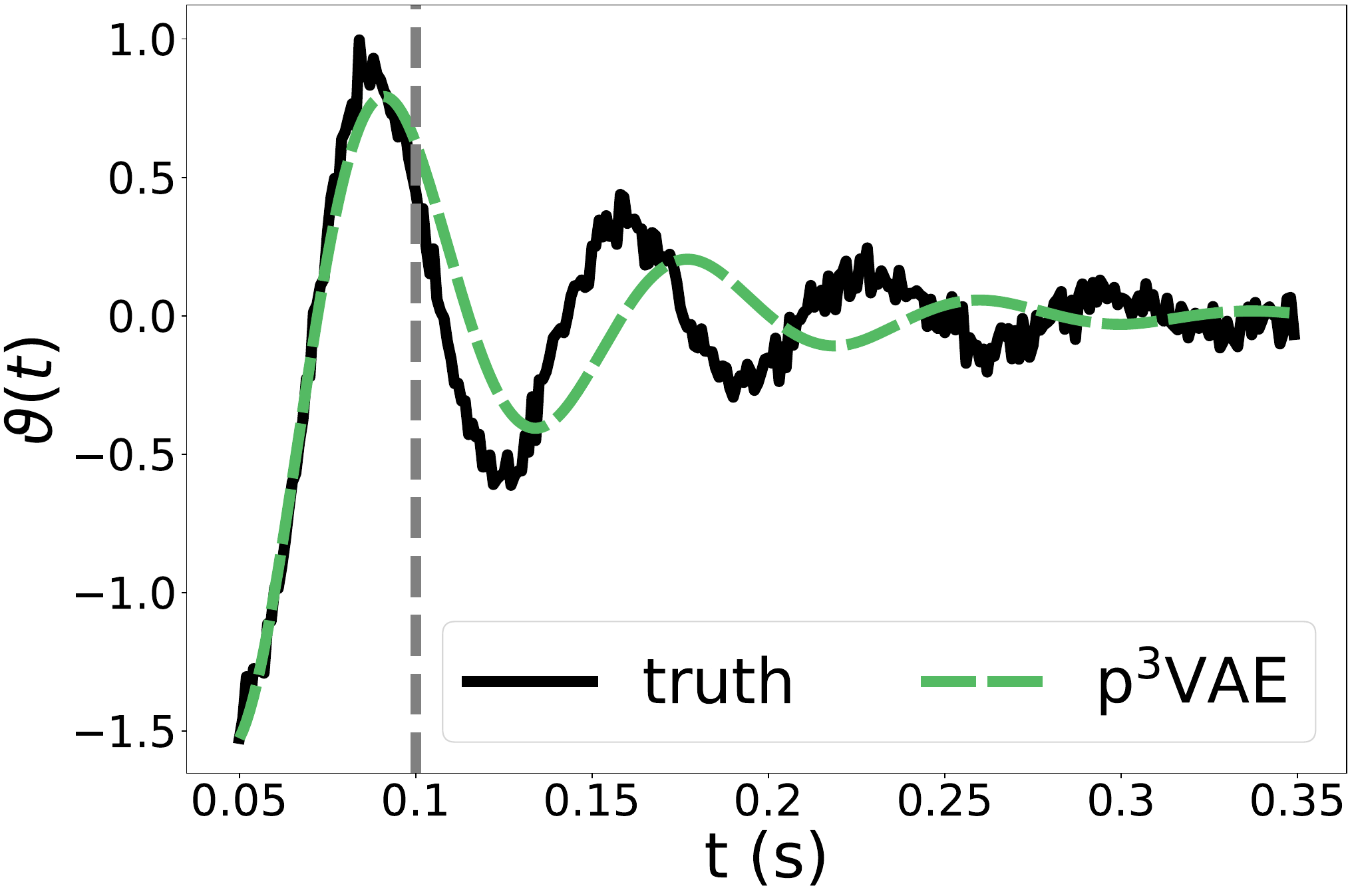}}
    \subfloat{\includegraphics[height=0.215\textwidth]{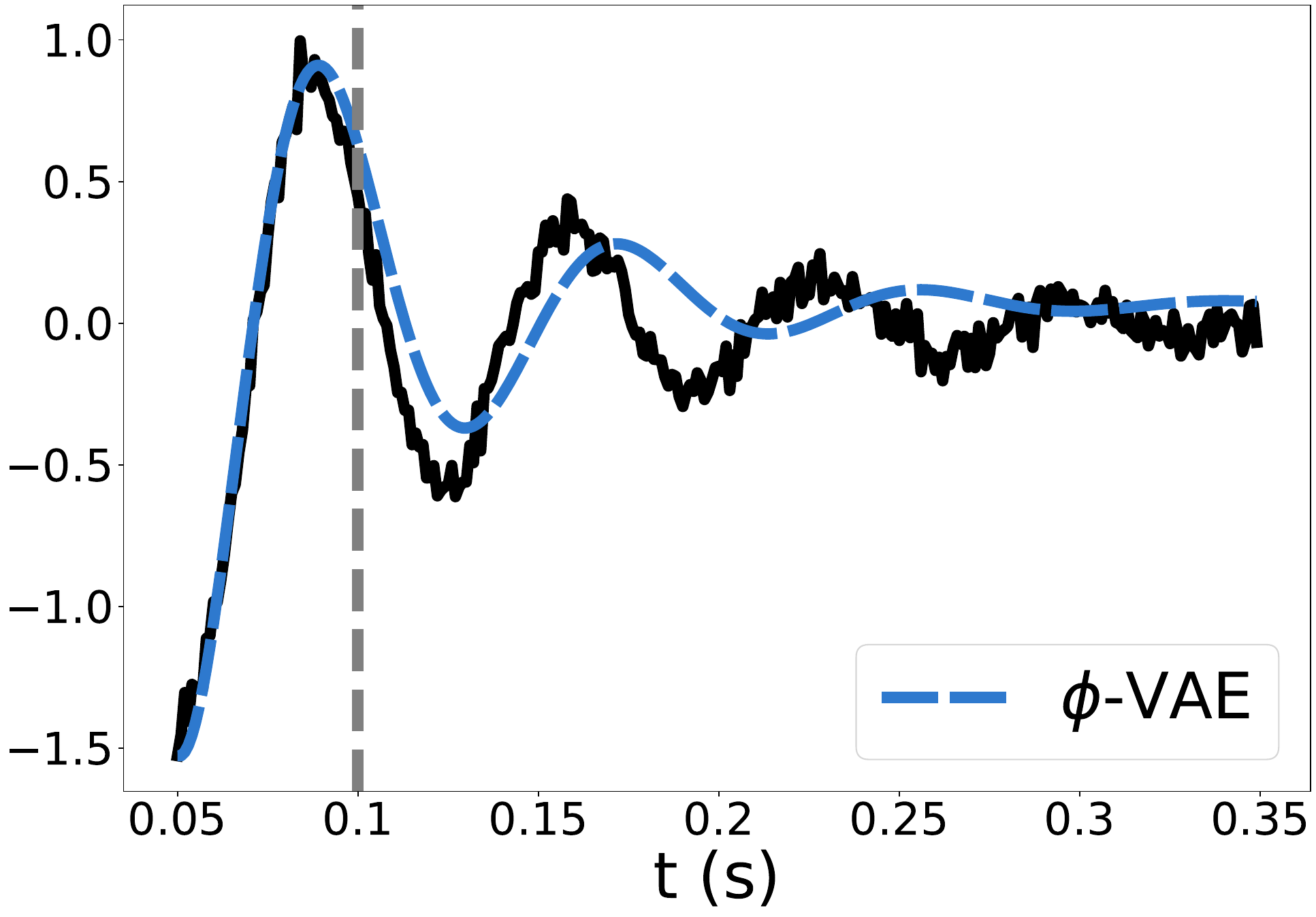}}
    \subfloat{\includegraphics[height=0.215\textwidth]{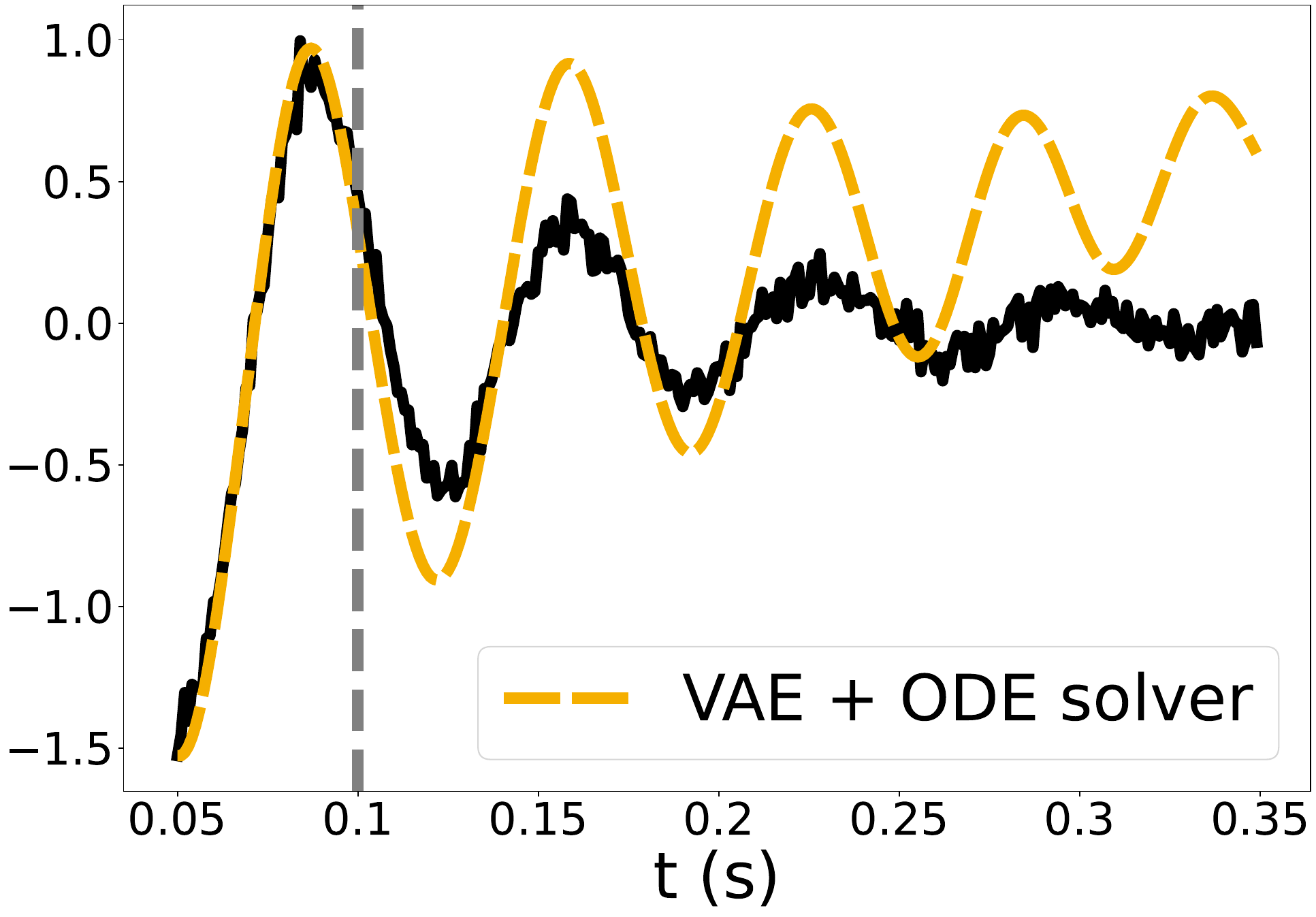}}
    
    \subfloat{\includegraphics[height=0.215\textwidth]{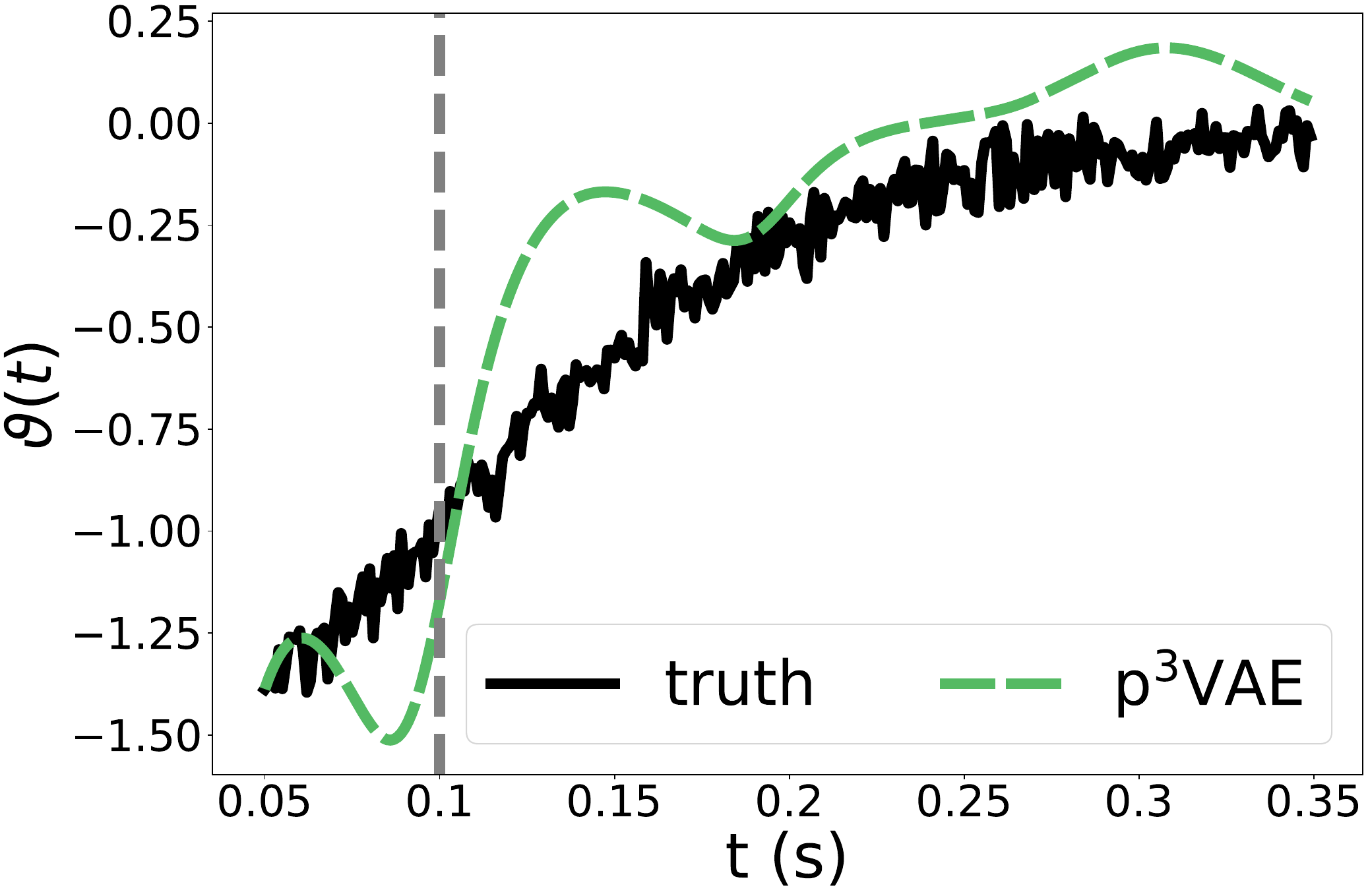}}
    \subfloat{\includegraphics[height=0.215\textwidth]{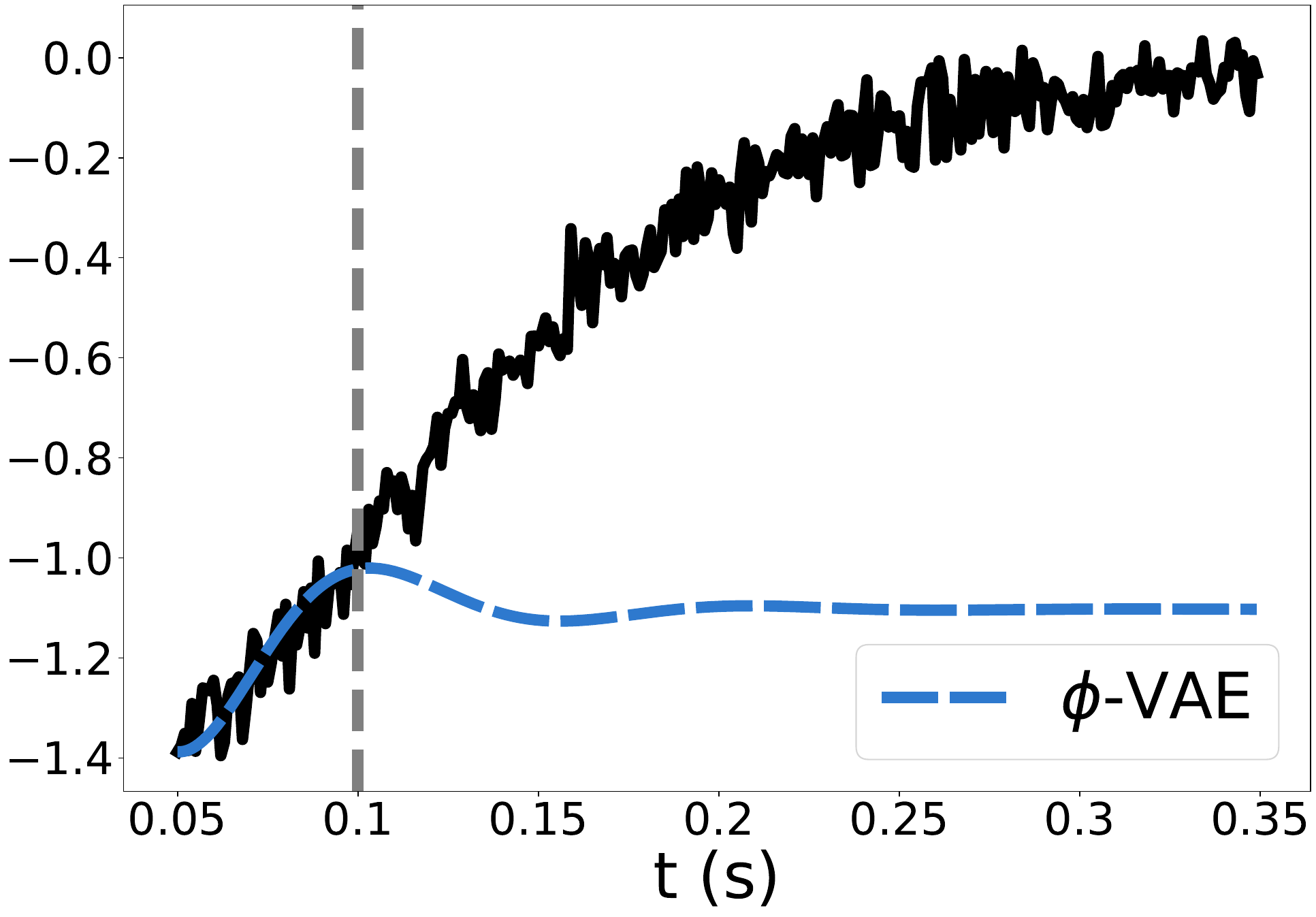}}
    \subfloat{\includegraphics[height=0.215\textwidth]{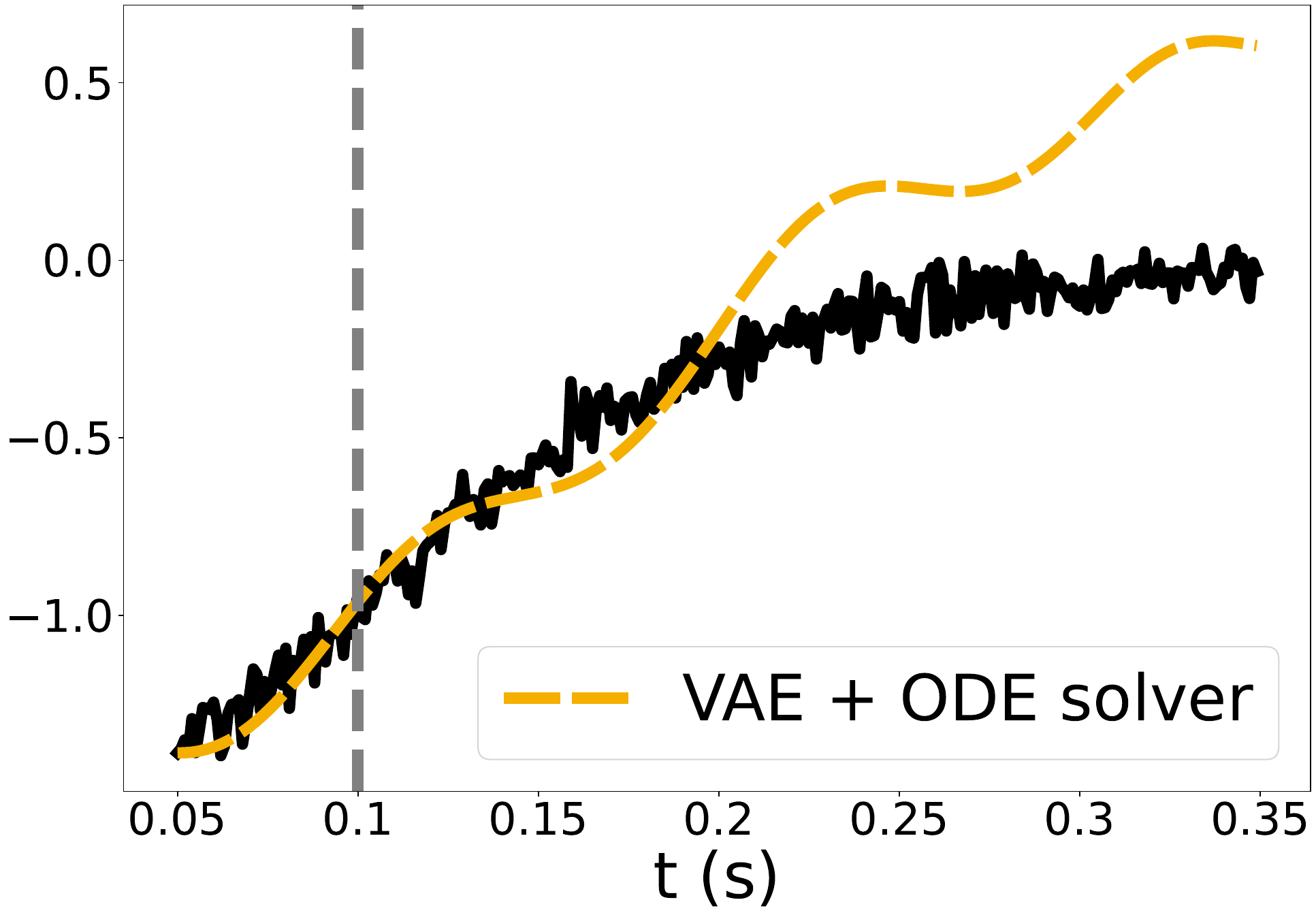}}
    \caption{Predicted dynamics of p$^3$VAE (ours), $\phi$-VAE \cite{takeishi2021physics} and VAE + ODE solver \cite{yildiz2019ode2vae, tothhamiltonian} for the damped pendulum vs. ground truth trajectory $\frac{d^2\vartheta}{dt^2}(t) + \xi \frac{d\vartheta}{dt}(t) + \omega^2 \mbox{ sin } \vartheta(t) = 0$ with noise, for several $\xi$, $\omega$, and initial conditions. \label{fig:appendix_pendulum}}
\end{figure*}

Fig. \ref{fig:appendix_pendulum} shows additional qualitative comparisons.

\section{Hyperspectral image classification}

\subsection{Synthetic data \label{sec:appendix_hyp_dart_data}}

\begin{figure*}[t]%
    \centering
    \subfloat[]{\includegraphics[width=0.3\textwidth]{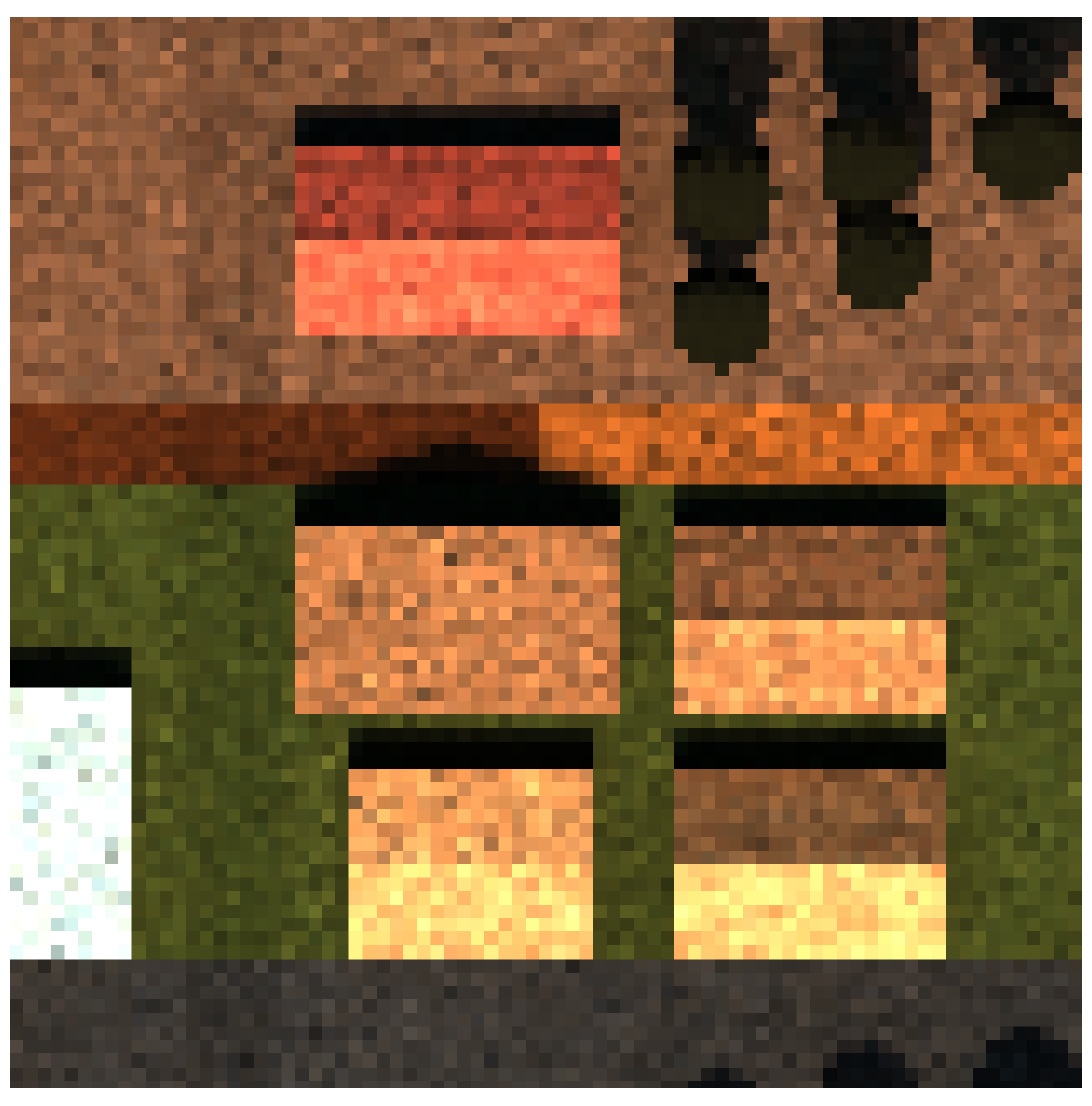}}
    \hspace{0.05cm}
    \subfloat[]{\includegraphics[width=0.3\textwidth]{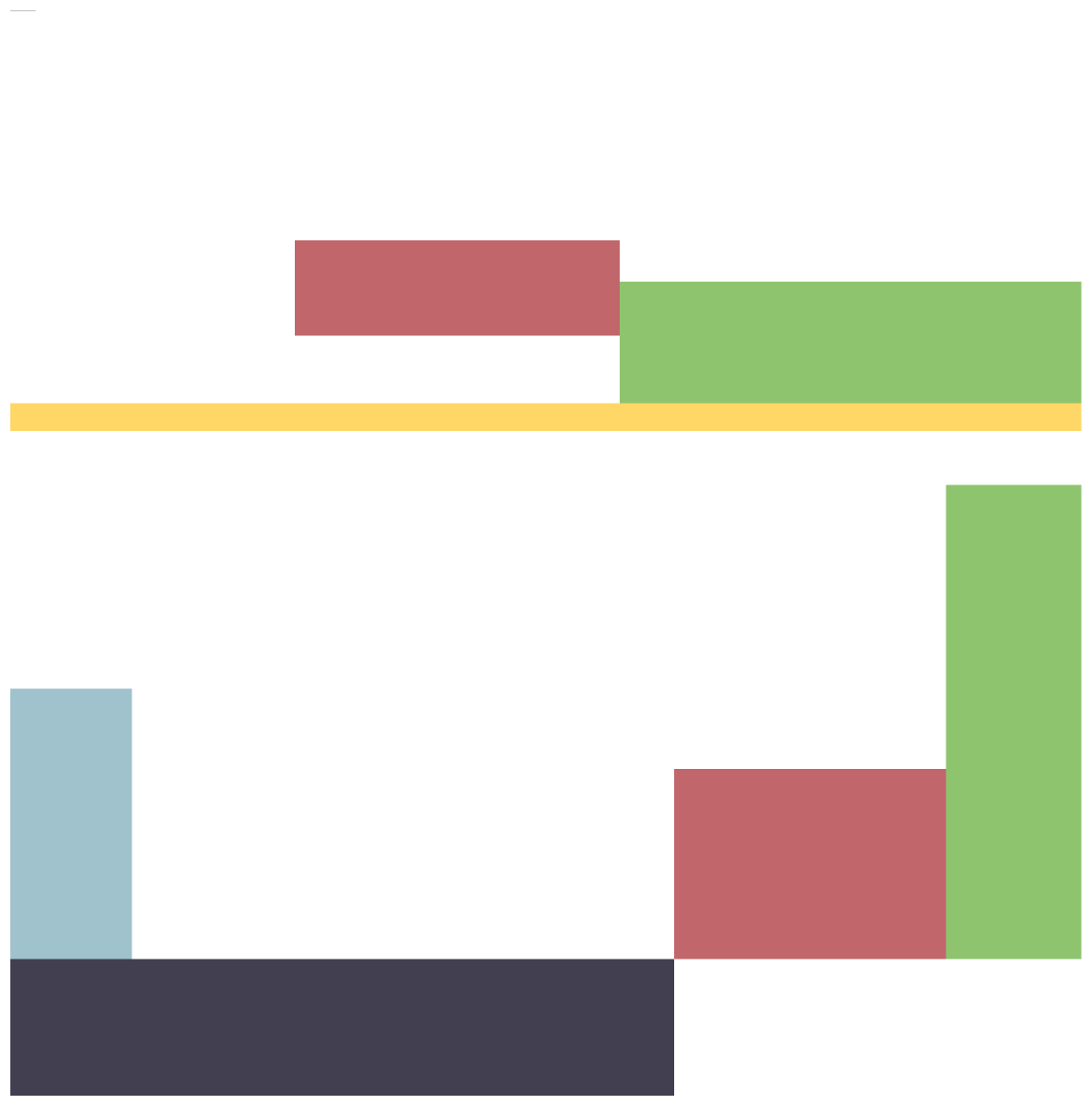}}
    \hspace{0.05cm}
    \subfloat[]{\includegraphics[width=0.3\textwidth]{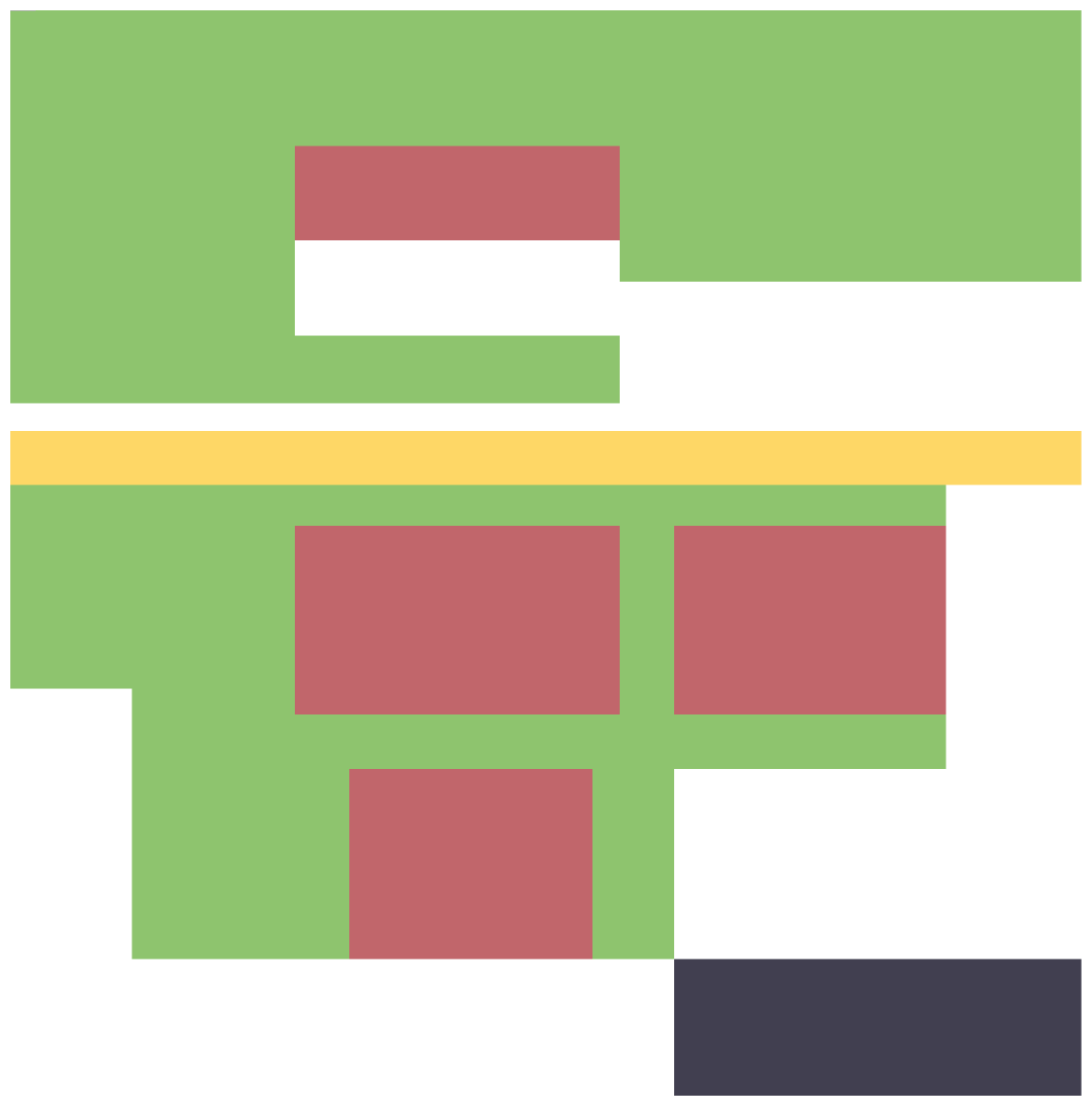}}

    \caption{(a) False color composition of the simulated hyperspectral image and its (b) training and (c) validation ground truths.  \label{fig:traindataset}}
\end{figure*}

We simulated an airborne hyperspectral image with the radiative transfer software DART \cite{gastellu2012dart}. 300 spectral bands with a $6.5$ nm resolution and a $0.5$ m ground sampling distance were simulated, without simulating the Earth-atmosphere coupling. A false-color image and its ground truth are shown in Fig. \ref{fig:traindataset}. The scene was simulated with five materials (vegetation, sheet metal, sandy loam, tile and asphalt) whose mean reflectance spectra are shown in fig. \ref{fig:reflectance_estimates} (spectra were sampled with a gaussian noise). Some classes gather different materials to have a realistic intrinsic inter-class variability. For instance, the vegetation class has three spectra which correspond to healthy grass, stressed grass and eucalyptus reflectances. We refer to those different materials within one class as subclasses. The azimuth angle is equal to 180 degrees, the solar zenithal angle is equal to 30 degrees and the roofs have different slopes. Moreover, to reproduce the scarcity of annotations in remote sensing, we labeled only a few pixels in the training data set and used the unlabeled pixels in our validation set. In particular, the training data set:
\begin{itemize}
\setlength{\itemindent}{.2in}
	\item[\textemdash] does not include vegetation, asphalt and sandy loam pixels in the shadows,
    \item[\textemdash] partially includes tile pixels on different slopes,
    \item[\textemdash] contains all pixels of sheet metal.
\end{itemize}
Simulating a scene was very convenient as we precisely knew the topography and the reflectance of each material. To test our model, we computed roughly $10,000$ test spectra per class with different slopes, direct irradiance and sky viewing angle. We also made different combinations of reflectance spectra within each class. Precisely, the factors that were used to generate a spectra $\x$ were the class $y$, the portion of direct irradiance $\delta_{dir}$, the solar incidence angle $\Theta$, the sky viewing angle factor $\Omega$, the anisotropy correction coefficient $p_\Theta$, the intra-class mixing coefficient $\alpha \in [0,1]$ and the subclass configuration $\eta$: 
\begin{align}
\begin{split}
   \x(\lambda) = & \frac{\delta_{dir} \mbox{cos} \: \Theta I_{dir}^{code}(\lambda) + \Omega \: p_\Theta I_{dif}^{code}(\lambda)}{\mbox{cos} \: \Theta^{code} \cdot I_{dir}^{code}(\lambda) + I_{dif}^{code}(\lambda)} \tilde{\boldsymbol{\rho}}(\lambda) \\ 
\end{split}
\end{align}
with:
\begin{equation}
    \tilde{\boldsymbol{\rho}}(\lambda) =  \alpha \boldsymbol{\rho}[y][\eta_1] + (1-\alpha) \boldsymbol{\rho}[y][\eta_2]
\end{equation}
where $\rho[y][\eta_i]$ denotes the reflectance spectrum of the $\eta_i$th subclass of class $y$. \\

\noindent
For the test data set, the simulation of spectra rather than images was very convenient to cover the whole range of variation factors.

\subsection{Real data \label{sec:appendix_hyp_real_data}}

We used subsets of the recently published airborne hyperspectral data set of the CAMCATT-AI4GEO experiment\footnote{Data is publicly available here: \url{https://camcatt.sedoo.fr/}} in Toulouse, France \cite{ROUPIOZ2023109109}.
Data was acquired with a AisaFENIX 1K camera, which covers the spectral range from 0.4 $\mu$m to 2.5 $\mu$m with a 1 m ground sampling distance. 
Data was converted in radiance at aircraft level through radiometric and geometric corrections. 
Then, the radiance image was converted to surface reflectance with the atmospheric correction algorithm COCHISE \cite{miesch2005direct} (that makes a flat surface assumption). 
We labeled a ground truth through a field campaign and photo interpretation. We selected eight land cover classes (Tile, Asphalt, Vegetation, Painted sheet metal, Water, Gravels, Metal and Fiber cement) which were spatially split in a labeled training set, an unlabeled training set and a test set (spectra are shown in Fig. \ref{fig:real_spectra}). 
The labeled training set, unlabeled training set and test set contain 3671 pixels, 7762 pixels and 10333 pixels, respectively. 

\subsection{Physical prior knowledge \label{sec:appendix_radiative_modeling}}

We assume, as most atmospheric codes do, that land surfaces are lambertian, \textit{i.e.} that they reflect radiation isotropically. The reflectance of a pixel at wavelength $\lambda$ is commonly defined in the literature as follows:
\begin{align}
    \x(\lambda) & = \frac{\pi R_{dir}(\lambda)}{I_{tot}(\lambda) \tau_{dir}(\lambda)}
\end{align}
with 
\begin{align}    
     \left\{ \begin{array}{ll}
	R_{dir}(\lambda) & = R_{tot}(\lambda) - R_{env}(\lambda) - R_{atm}(\lambda) \\ 
    & \\ 
	I_{tot}(\lambda) & = I_{dir}(\lambda) + I_{dif}(\lambda) + I_{coup}(\lambda) + I_{refl}(\lambda)
	\end{array} \right.
\end{align}
where:
\begin{itemize}
    \item[\textemdash] $R_{tot}$ is the radiance measured by the sensor,
	\item[\textemdash] $R_{atm}$ is the portion of $R_{tot}$ that is scattered by the atmosphere without any interactions with the ground,
	\item[\textemdash] $R_{env}$ is the portion of $R_{tot}$ that comes from the neighbourhood of the pixel,
	\item[\textemdash] $R_{dir}$ is the portion of $R_{tot}$ that comes from the pixel, 
	\item[\textemdash] $I_{dir}$ is the irradiance directly coming from the sun, 
	\item[\textemdash] $I_{dif}$ is the irradiance scattered by the atmosphere, 
	\item[\textemdash] $I_{coup}$ is the irradiance coming from the coupling between the ground and the atmosphere,
	\item[\textemdash] $I_{refl}$ is the irradiance coming from neighbouring 3D structures.  
\end{itemize}

\noindent
In section \ref{sec:prior_hyp_seg}, we neglect the contributions of $I_{coup}$ and $I_{refl}$ in order to derive eq. \ref{eq:cochise_correction}.
%You may include other additional sections here. 

\subsection{Latent variables modeling \label{sec:appendix_latent_modeling_hyp}}

An intuitive interpretation of $\z_I$ is the abundance of different subclasses within a land cover class. 
Therefore, we model $\z_I$ prior and posterior with Dirichlet distributions. We model $z_E$ prior and posterior with Beta distributions for their flexibility and domain bounded between 0 and 1. Specifically, we define the priors as follows:
\begin{align}
    p(z_E) & := Beta(z_E \vert \alpha^o, \beta^o); &
    p(\z_I) & := Dir(\z_I \vert \gamma^o ) 
\end{align}
We empirically set $\alpha^o$ to $1$ and $\beta^o = \frac{1- \mbox{cos} \: \Theta^{code}}{\mbox{cos} \: \Theta^{code} + \epsilon} \alpha^o$ where $\epsilon$ is a small constant to avoid division by zero. 
This prior distribution favors high values of $z_P$ while $\mathbb{E}_{p(z_E)}z_E = \mbox{cos} \: \Theta^{code}$, which is what we expect on a flat ground. 
We set $\gamma^o =[1 \ldots 1]_{1 \times \vert \z_I \vert}$, which is equivalent to having a uniform prior. 
We assume that $y$ has a uniform prior and that $y$, $\z_I$ and $z_E$ are a priori independent. 

\subsection{p$^3$VAE likelihood}\label{hybrid_model_likelihood}

In contrast to the standard Gaussian modeling, we model the likelihood of p$^3$VAE for the hyperspectral image classification task as follows: 
\begin{align}
    \begin{split}
    \ppdf(\x \vert y, z_E, \z_I) := \frac{1}{Z} & \: \mathcal{N}(\x \vert \boldsymbol{\mu}, \sigma^2 \boldsymbol{I}) \cdot \mbox{exp}\big(- \lambda \: \mbox{arccos}(\frac{\x^T \boldsymbol{\mu}}{\|\x\| \|\boldsymbol{\mu}\|})\big)
    \end{split}
\end{align}
where $\boldsymbol{\mu} = f_E(f_I^{\param}(y, \z_I), z_E)$, $\sigma$ and $\lambda$ are hyperparameters, and $Z$ is a finite constant such that the density integrates to one. The negative log-likelihood derives as follows: 
\begin{align}
	-\mbox{log} \: \ppdf (\x \vert y, z_E, \z_I) = & \frac{1}{\sigma^2} \mbox{MSE}(\x, \boldsymbol{\mu}) + \lambda \: \mbox{arccos}(\frac{\x^T \boldsymbol{\mu}}{\|\x\| \|\boldsymbol{\mu}\|}) + C
\end{align}
where C is a constant. 
Therefore, maximizing the likelihood of the data is equivalent to minimizing a linear combination of the mean squared error and the spectral angle between the observations $\x^{(i)}$ and $\boldsymbol{\mu}^{(i)}$. The other consequence of this additional term in the density is that, without knowing the value of the constant $Z$, we cannot properly evaluate the likelihood for a given data point and neither sample from the likelihood. However, it is not an issue in our case because we are only interested in the predictions of the model.  \\

\noindent
Now, let us prove that this decoder actually defines a proper likelihood. Let us define $f(\x)=\mathcal{N}(\x \vert f_E(f_I^{\param}(y, \z_I), z_E), \sigma^2 \boldsymbol{I})$ and $\mathcal{S}(\x) = \mbox{exp}\big(- \lambda \: \mbox{arccos}(\frac{\x^T \boldsymbol{\mu}}{\|\x\| \|\boldsymbol{\mu}\|})\big)$, such that: 
\begin{align}
    \ppdf(\x \vert y, z_E, \z_I) := \frac{1}{Z} \: f(\x) \mathcal{S}(\x) 
\end{align}
where $Z$ is a finite constant such that the density integrates to one. Let us prove that such a constant $Z := \int f(\x) \cdot \mathcal{S}(\x) dx$ exists. 
First, we can easily show that $f : [0,1]^B \longrightarrow \mathbb{R}$ is square-integrable, as well as $\mathcal{S} : [0,1]^B \longrightarrow \mathbb{R}$:
\begin{align}
    \int \vert \mathcal{S}(\x) \vert ^2 d\x & = \int_{[0,1]^B} \mbox{exp}\big(-2 \lambda \: \mbox{arccos}(\frac{\x^T \boldsymbol{\mu}}{\|\x\| \|\boldsymbol{\mu}\|})\big) d\x \leq \int_{[0,1]^B} d\x = 1
\end{align}
since $\mbox{exp}\big(-2 \lambda \: \mbox{arccos}(\frac{\x^T \boldsymbol{\mu}}{\|\x\| \|\boldsymbol{\mu}\|})\big) \leq 1 $ for all $\boldsymbol{\mu}$, $\x \in [0,1]^B$. Moreover, $f$ and $\mathcal{S}$ being continuous over $[0,1]^B$, the Cauchy-Schwarz inequality implies that:
\begin{align}
    \Bigg\lvert \int f(\x) \cdot \mathcal{S}(\x) d\x \Bigg\rvert & \leq \Bigg( \int f(\x)^2 dx\Bigg)^{\frac{1}{2}} \Bigg( \int \mathcal{S}(\x)^2  d\x\Bigg)^{\frac{1}{2}} \nonumber \\
    & = C \in \mathbb{R}
\end{align}
Thus, $Z \in \mathbb{R}$ and $\ppdf(\x \vert y, z_E, \z_I)$ properly defines a probability density function. 

\subsection{Qualitative disentanglement evaluation \label{sec:qual_dis}}

\begin{figure*}[h]%
    \centering
    \subfloat[Physics-guided VAE]{
        \includegraphics[width=0.48\textwidth]{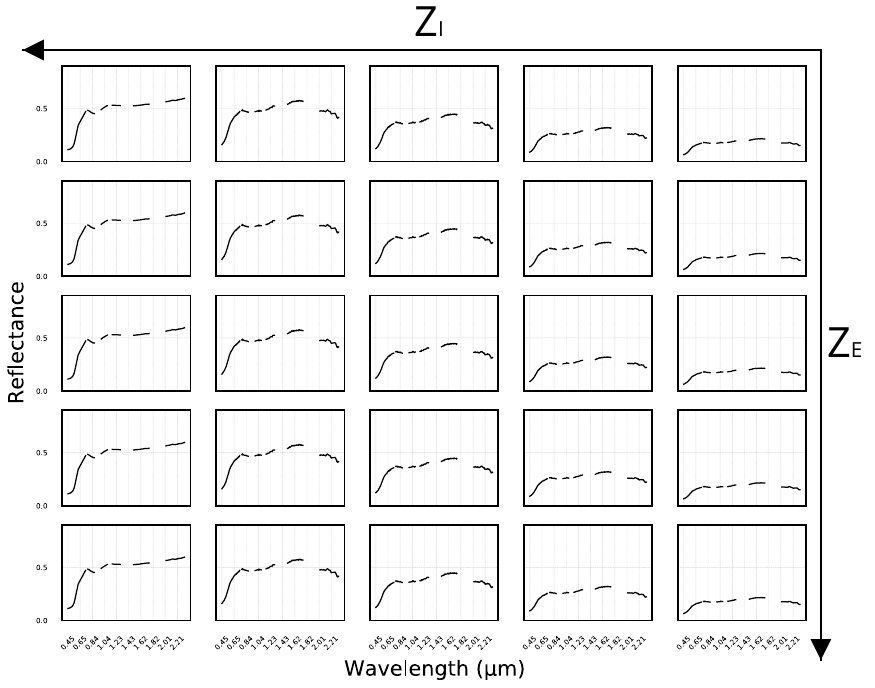}}
    \subfloat[p$^3$VAE]{
        \includegraphics[width=0.48\textwidth]{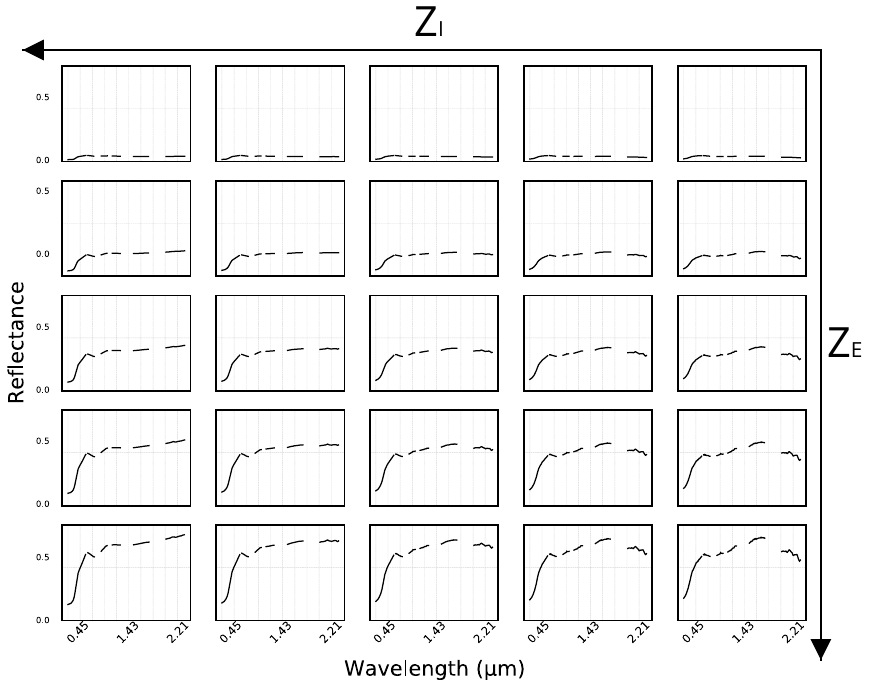}}
        
    \subfloat[Gaussian VAE]{\includegraphics[width=0.5\textwidth]{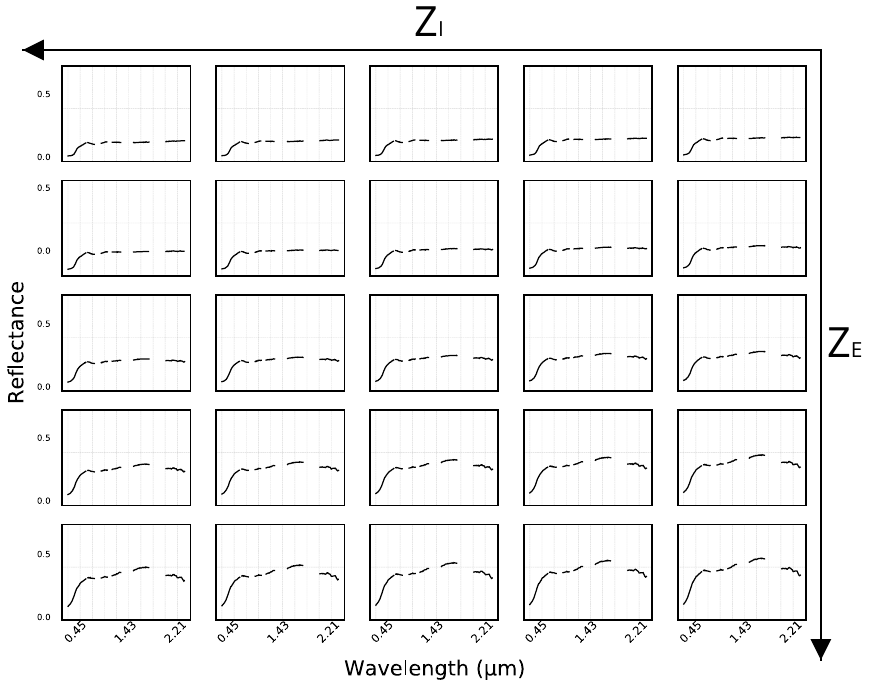}}
    \subfloat[ssInfoGAN]{\includegraphics[width=0.5\textwidth]{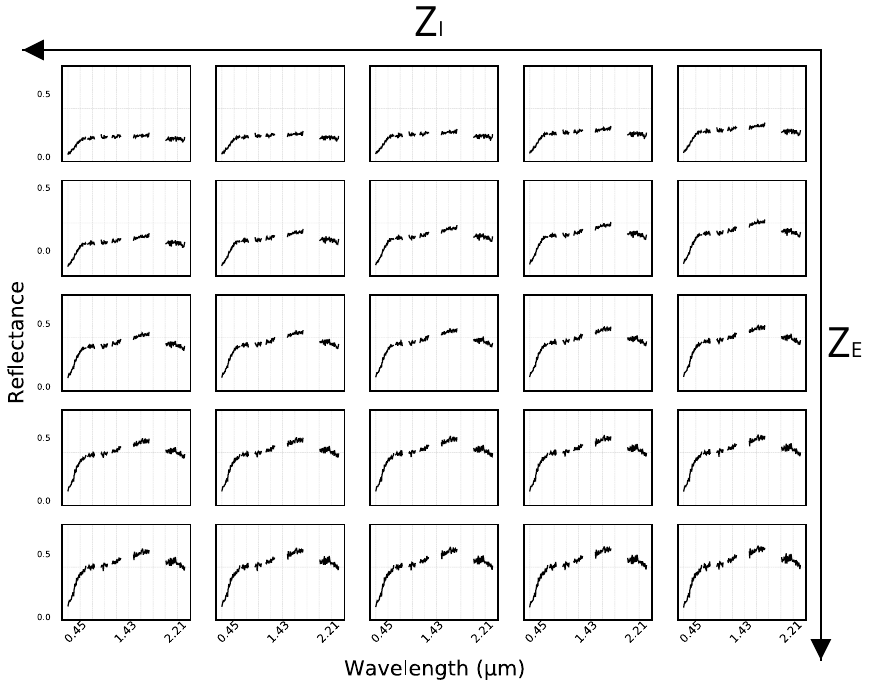}}
    
    \caption{Maximum likelihood estimates of the learned generative models along $z_E$ (on the y-axis) and $\z_I$ (on the x-axis) for the class \textit{Tile}. $z_E$ is interpolated between its minimum (top row) and maximum (bottom row) values inferred on the training data set. $\z_I$ is interpolated between two realizations of $\z_I$ that were inferred on two different types of tile of the training set (left and right columns) under high and low direct irradiance conditions, respectively. ssInfoGAN incompressible noise is randomly fixed.} \label{fig:max_estimate}
\end{figure*}

A qualitative way to evaluate the disentanglement of the latent space is to plot spectra that maximize the generative model likelihood for different values of the latent variables. Fig. \ref{fig:max_estimate} shows the maximum likelihood estimates of the class tile for the physics-guided VAE, p$^3$VAE, the gaussian VAE, and ssInfoGAN. The different estimates were obtained for different realizations of $\z_I$ and $z_E$. $z_E$ was interpolated between its lowest and highest values inferred on the training set. $\z_I$ was interpolated between two values inferred on different types of tile under high and low direct irradiance conditions. We see on Fig. \ref{fig:max_estimate} that latent variables $z_E$ and $\z_I$ are disentangled by p$^3$VAE. Firstly, for a given column, the variations of the spectra are only induced by the variation in irradiance conditions : the shape roughly stays the same but the intensity changes. Secondly, for a given row, the variations of the spectra are only induced by the change in the nature of matter : the intensity is the same but the shape changes. In contrast, the physics-guided VAE does not disentangle the latent variables. A variation along one latent variables induces a change both in the irradiance conditions and on the type of tile. \\

\noindent
A major difference that we should highlight between the quantitative and qualitative disentanglement evaluation is that the quantitative approach evaluates how a variation of the factors induces a variation of the latent code. The qualitative approach on the other hand evaluates how a change in the latent code induces a change in the observation. Therefore, assessing the disentanglement qualitatively requires expert knowledge. \\

\subsection{p$^3$VAE estimate of irradiance conditions}

\begin{figure*}
    \center
    \subfloat[Vegetation]{\includegraphics[width=0.19\textwidth]{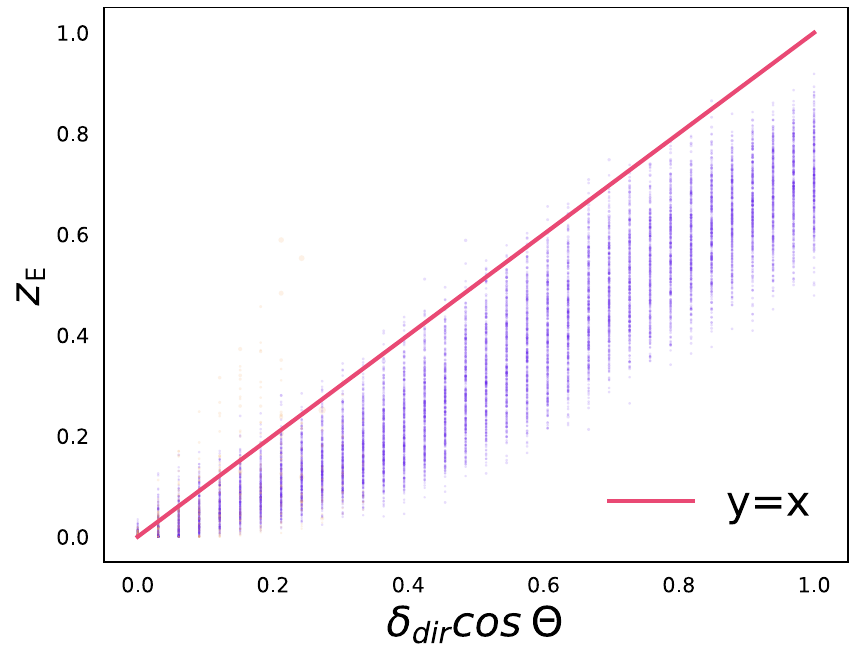}}
    \subfloat[Sheet metal]{\includegraphics[width=0.19\textwidth]{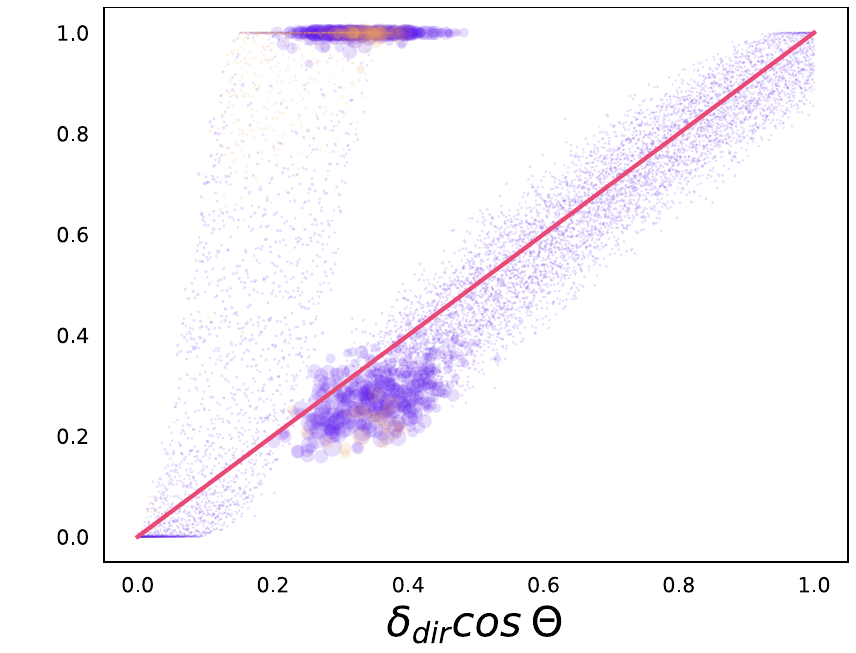}}
    \subfloat[Sandy loam]{\includegraphics[width=0.19\textwidth]{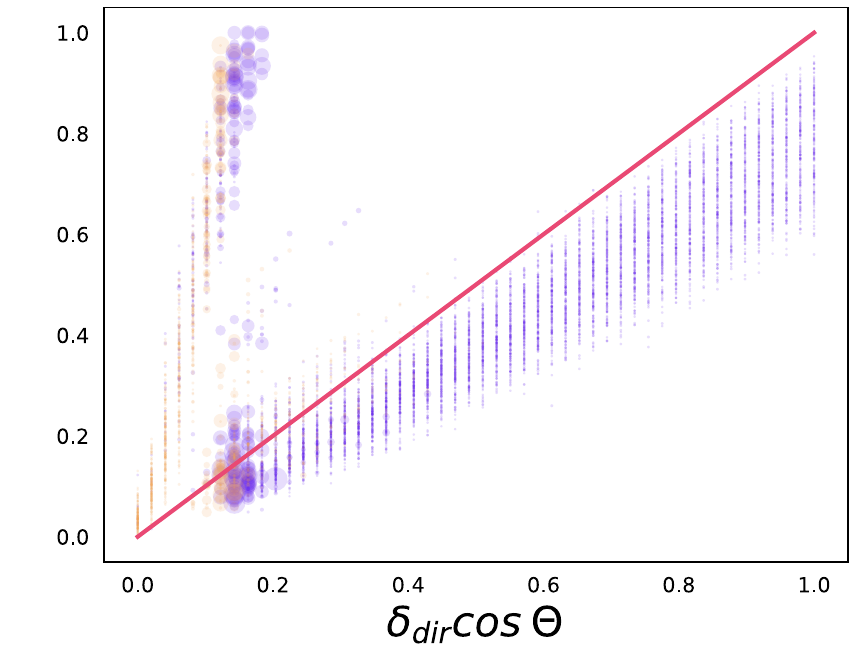}}
    \subfloat[Tile]{\includegraphics[width=0.19\textwidth]{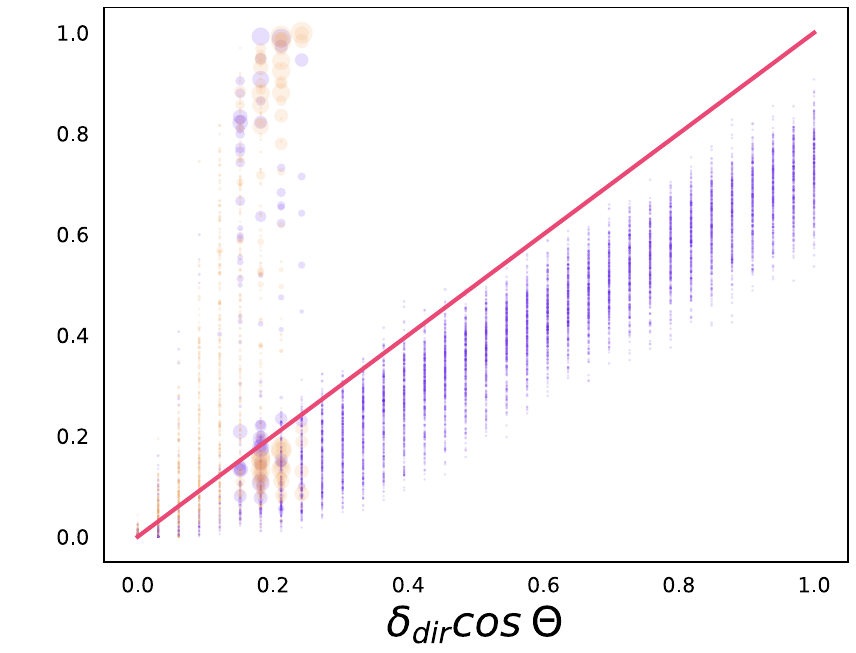}}
    \subfloat[Asphalt]{\includegraphics[width=0.19\textwidth]{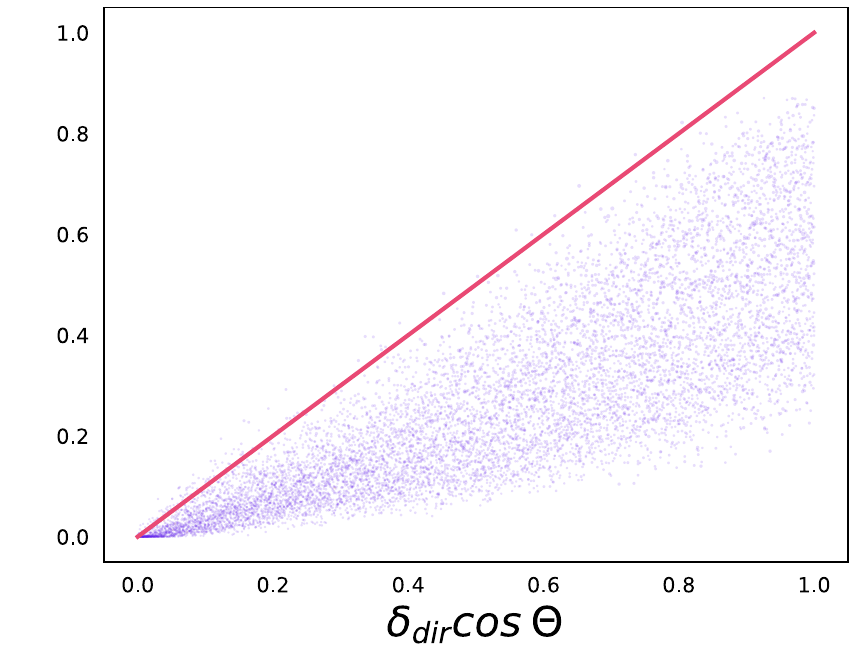}}
    
     \caption{Predicted $z_E$ against true $\delta_{dir} cos \: \Theta$ for each class with p$^3$VAE. Each point represents a pixel, which is correctly classified if shown in purple, or wrongly classified if shown in orange. The size of the points is proportional to the exponential of the empirical standard deviation of $z_E$. \label{fig:fusion_rho_cos}}
\end{figure*}

Fig. \ref{fig:fusion_rho_cos} shows, for each class, the inferred $z_E$ by p$^3$VAE against the true $\delta_{dir} cos \: \Theta$ of the test data set. Correctly classified pixels are shown as purple points while wrongly predicted pixels are shown as orange points. The size of the points is proportional to the exponential of the empirical standard deviation of $z_E$. First, we notice that most confusions are made when $\delta_{dir} cos \: \Theta$ is poorly estimated or when its true value is low. Secondly, we see that most confusions go hand in hand with high $z_P$ uncertainty estimates, meaning that two classes are likely, but under very different irradiance conditions. Finally, we see that there is a bias, in the sense that the average prediction of $\delta_{dir} cos \: \Theta$ is different from its true value. $\delta_{dir} cos \: \Theta$ is mostly under-estimated, which is the counterpart of over-estimated reflectance spectra as shown in Fig. \ref{fig:reflectance_estimates}. \\ 

\subsection{Model architectures}

Models architectures are described in Tab. \ref{tab:cnn_arch}, Tab. \ref{tab:VAE_arch}, Tab. \ref{tab:info_arch}, Tab. \ref{tab:nowork}. 

\begin{figure*}[h]
    \centering
    \subfloat[Tile]{\includegraphics[width=\textwidth]{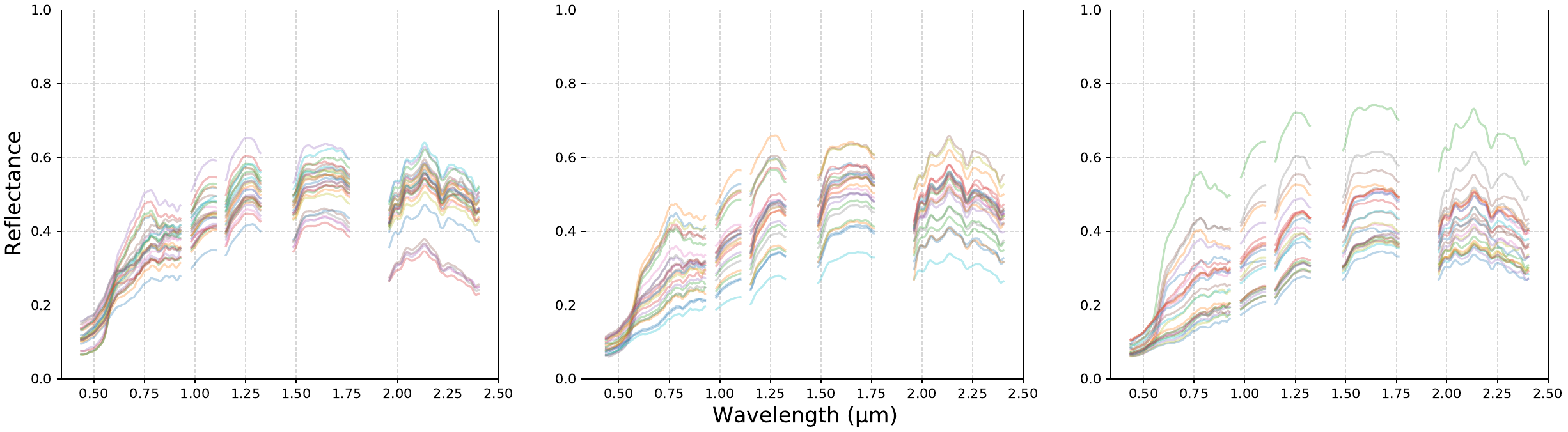}}
    
    \subfloat[Asphalt]{\includegraphics[width=\textwidth]{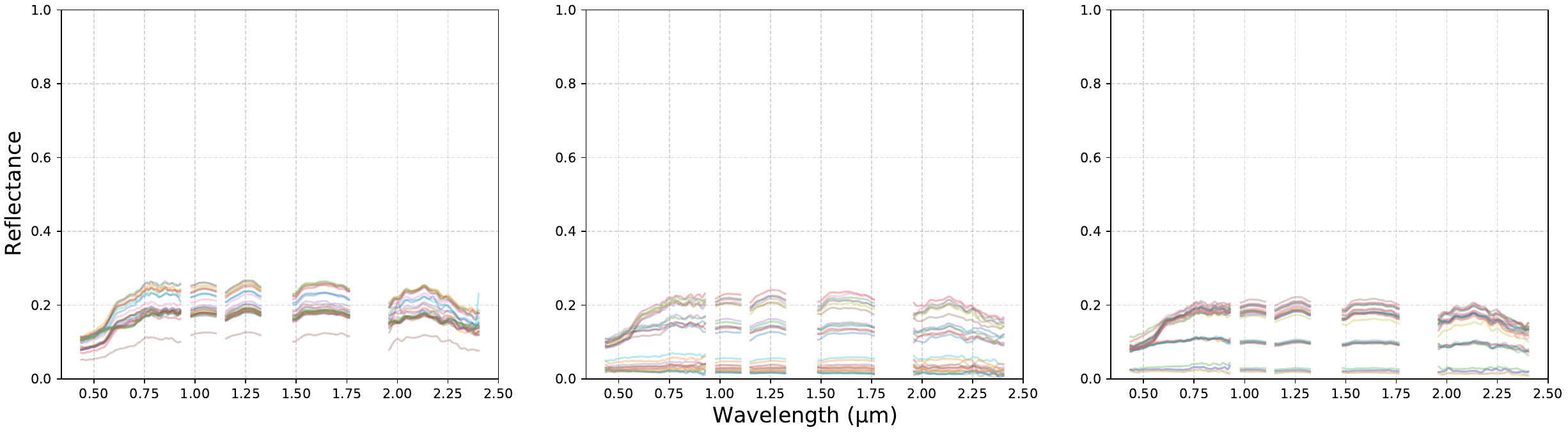}}
    
    \subfloat[Vegetation]{\includegraphics[width=\textwidth]{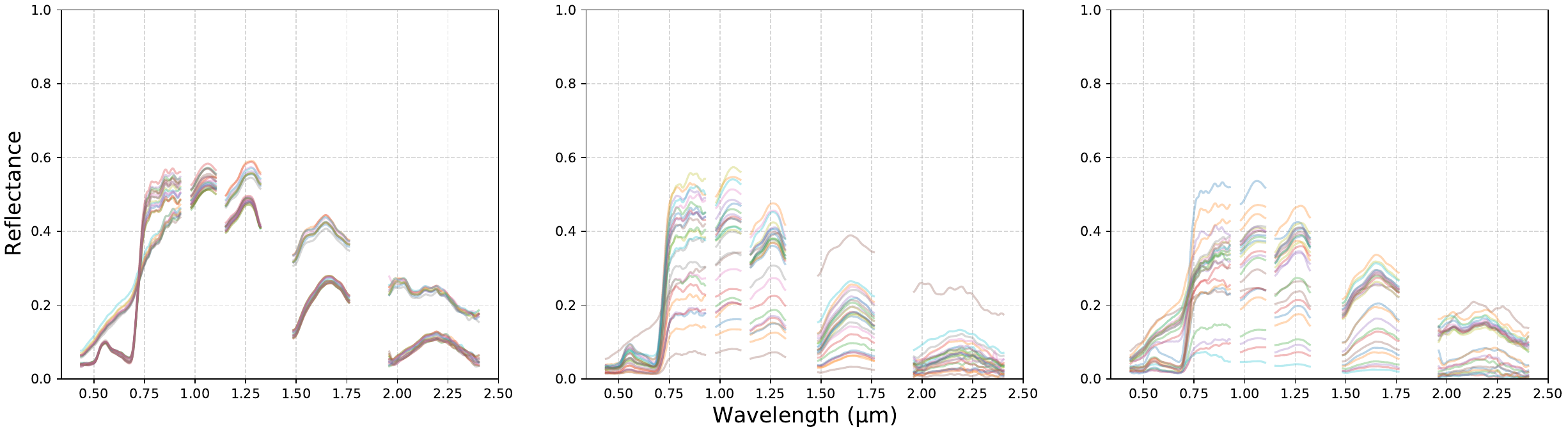}}
    
    \subfloat[Painted sheet metal]{\includegraphics[width=\textwidth]{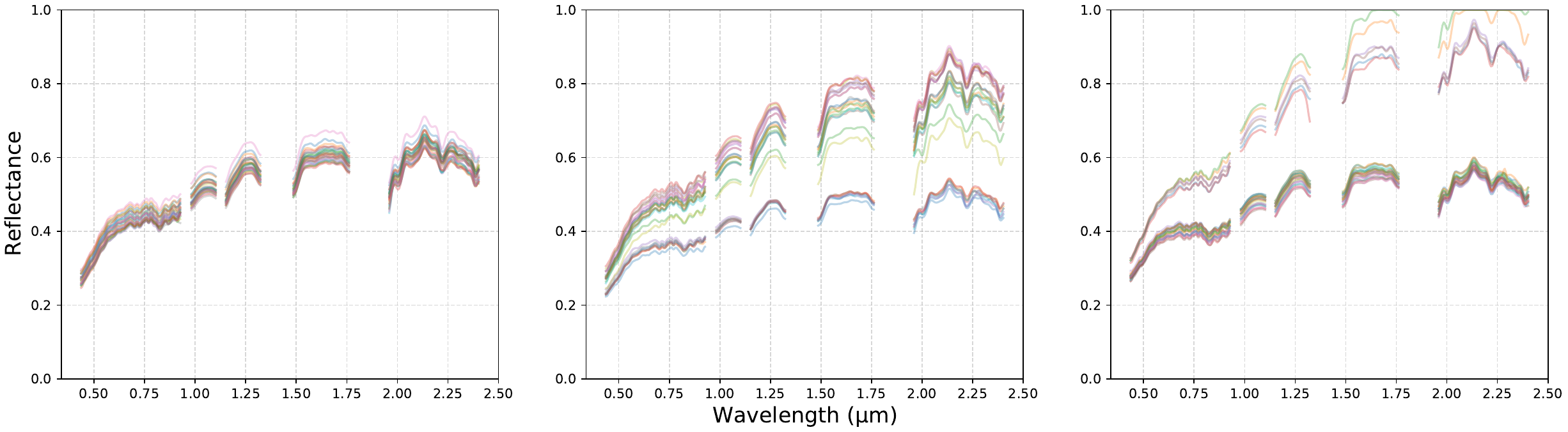}}
    
    \caption{Spectra from the (left) labeled, (middle) unlabeled and (right) test real data sets}
    \label{fig:real_spectra}
\end{figure*}

\begin{figure*}[h]\ContinuedFloat
    \centering
    \subfloat[Water]{\includegraphics[width=\textwidth]{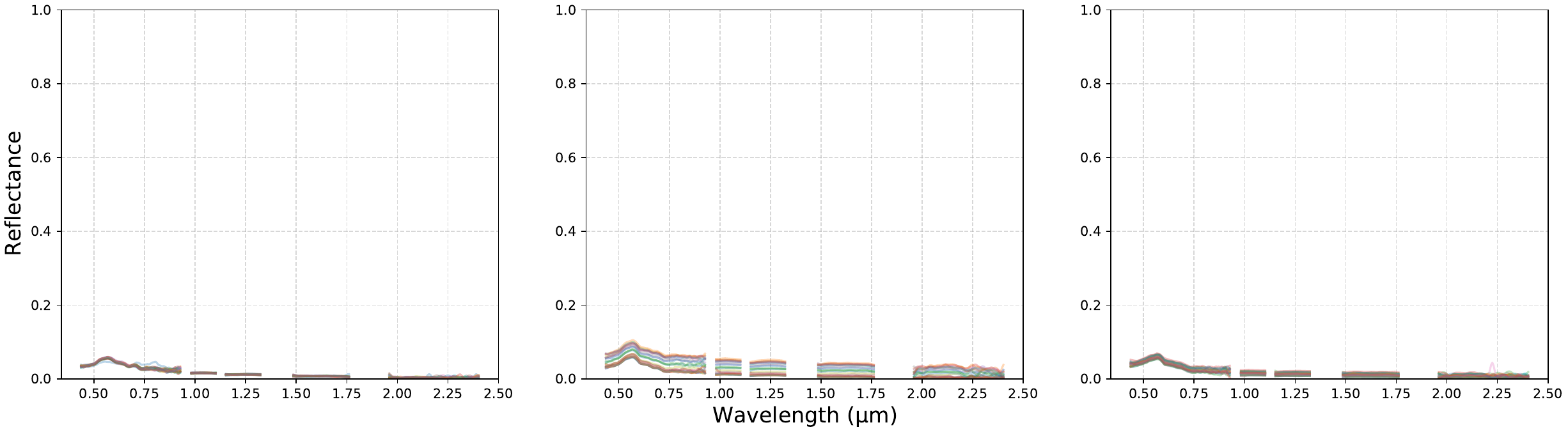}}
    
    \subfloat[Gravels]{\includegraphics[width=\textwidth]{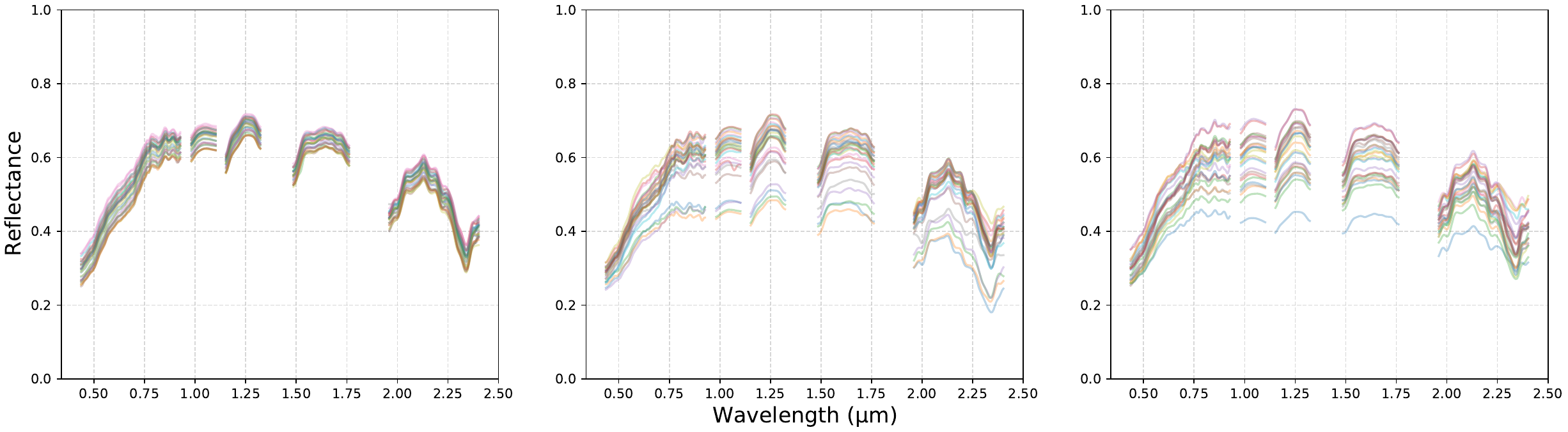}}
    
    \subfloat[Metal]{\includegraphics[width=\textwidth]{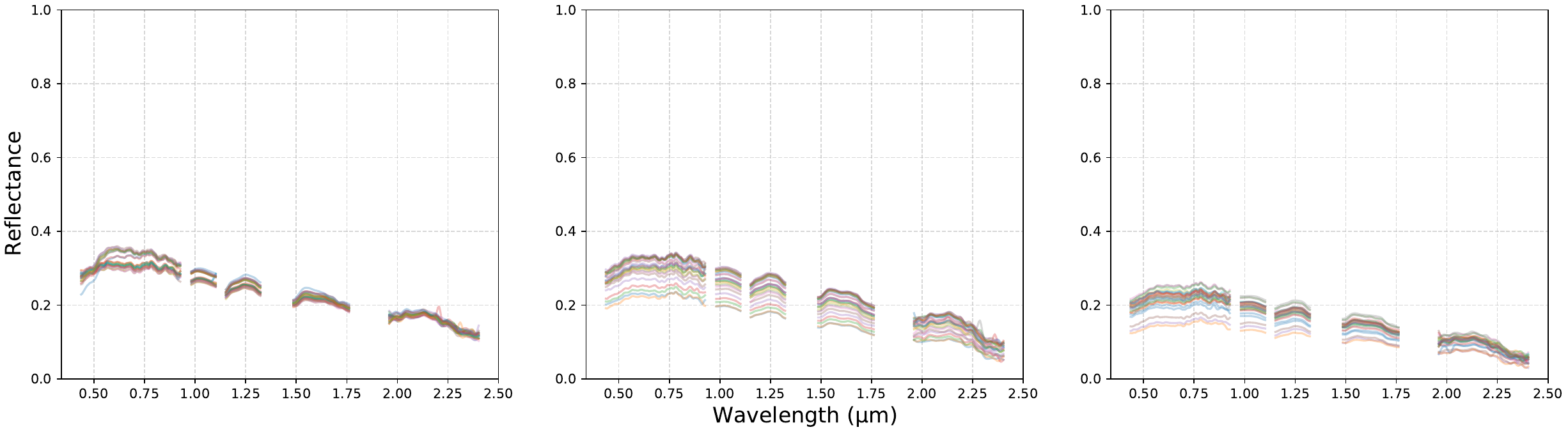}}
    
    \subfloat[Fiber cement]{\includegraphics[width=\textwidth]{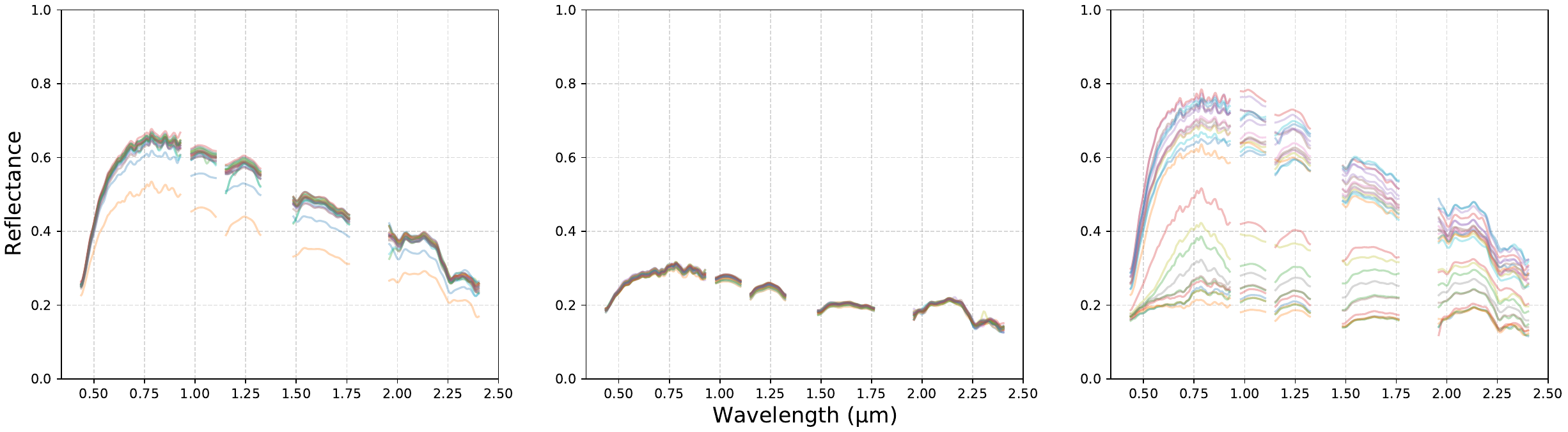}}
    
    \caption{Spectra from the (left) labeled, (middle) unlabeled and (right) test real data sets}
    \label{fig:real_spectra}
\end{figure*}

\subsection{Qualitative accuracy evaluation}

\begin{figure*}[h!]%
    \centering
    \subfloat[]{\includegraphics[width=0.18\textwidth]{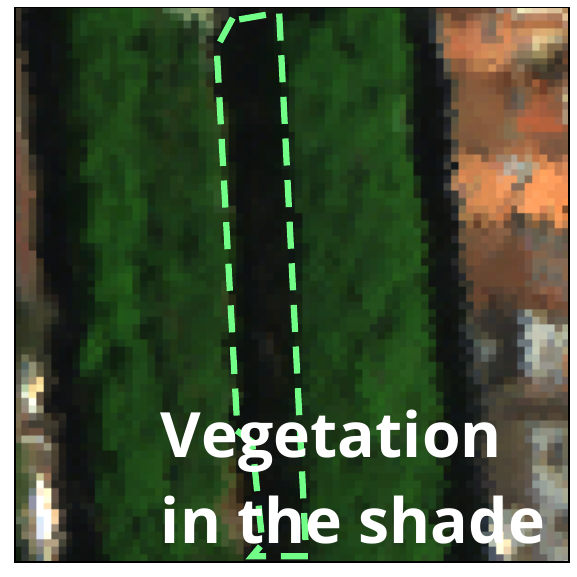}}
    \hspace{0.05cm}
    \subfloat[]{\includegraphics[width=0.18\textwidth]{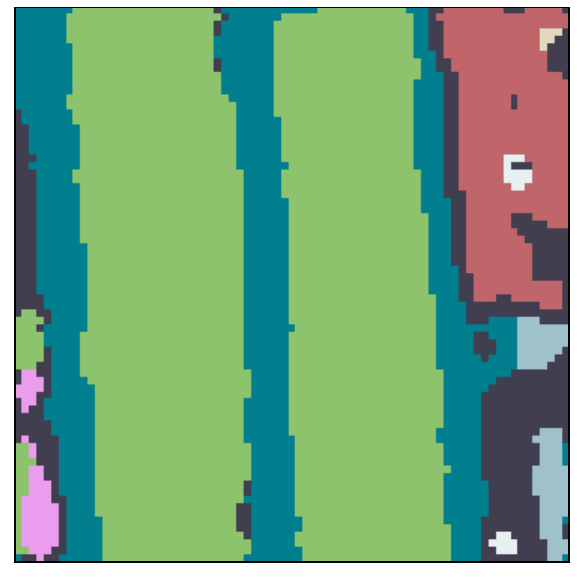}}
    \hspace{0.05cm}
    \subfloat[]{\includegraphics[width=0.18\textwidth]{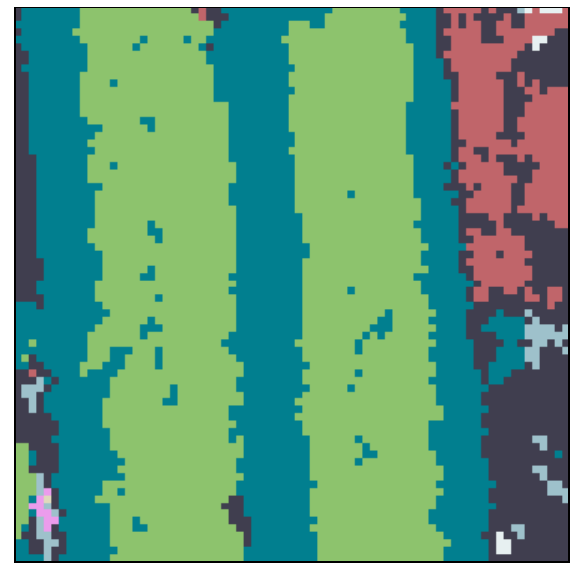}}
    \hspace{0.05cm}
    \subfloat[]{\includegraphics[width=0.18\textwidth]{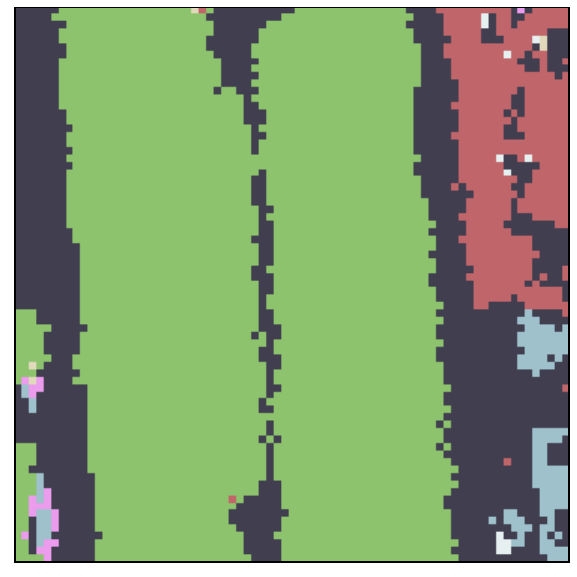}}
    \hspace{0.05cm}
    \subfloat[]{\includegraphics[width=0.18\textwidth]{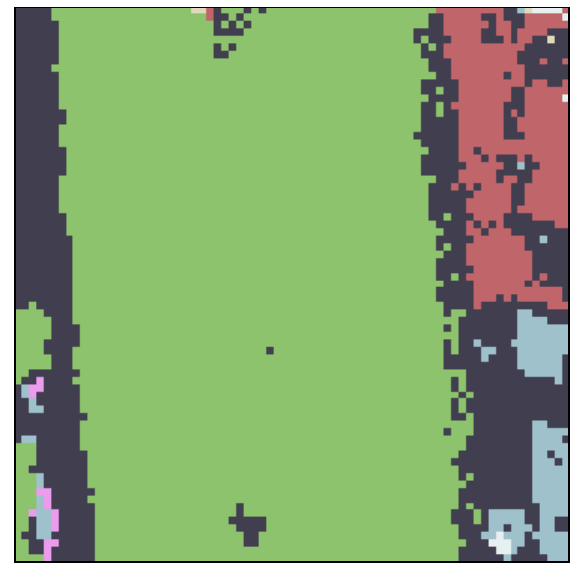}}
    
    \subfloat[]{\includegraphics[width=0.18\textwidth]{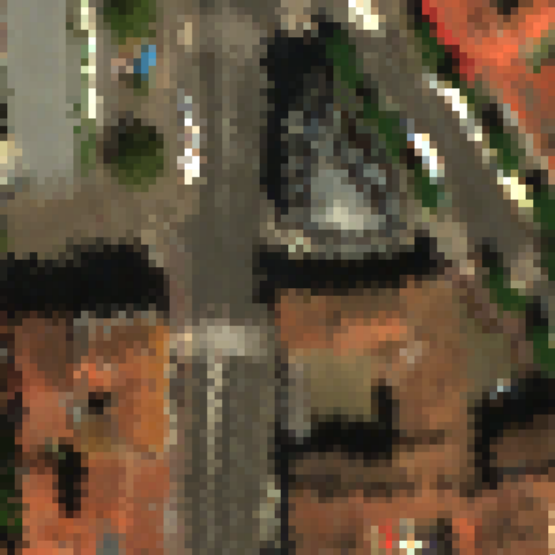}}
    \hspace{0.05cm}
    \subfloat[]{\includegraphics[width=0.18\textwidth]{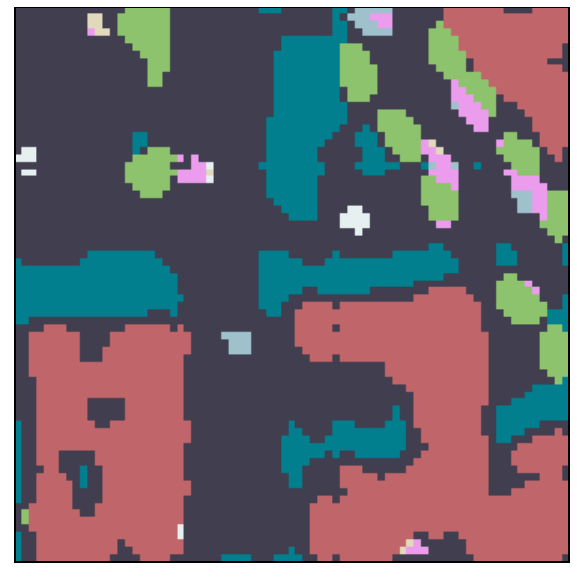}}
    \hspace{0.05cm}
    \subfloat[]{\includegraphics[width=0.18\textwidth]{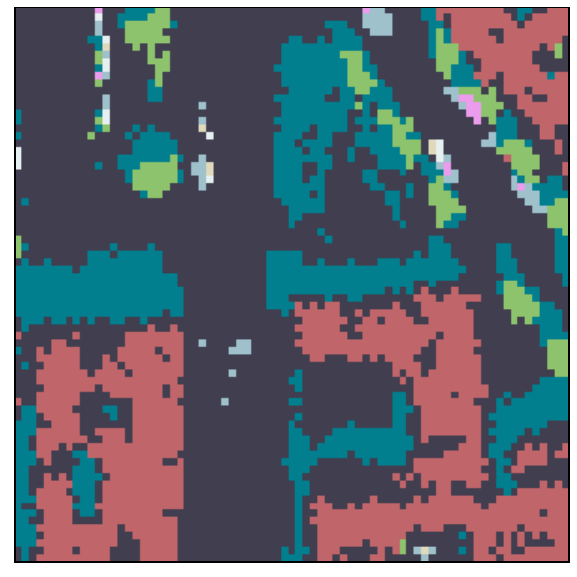}}
    \hspace{0.05cm}
    \subfloat[]{\includegraphics[width=0.18\textwidth]{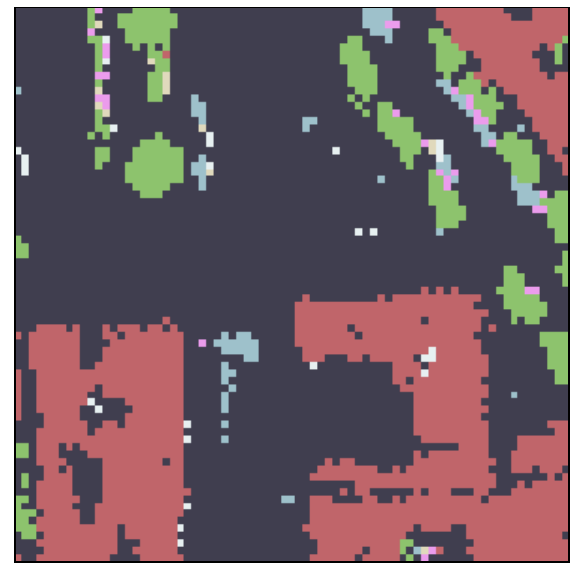}}
    \hspace{0.05cm}
    \subfloat[]{\includegraphics[width=0.18\textwidth]{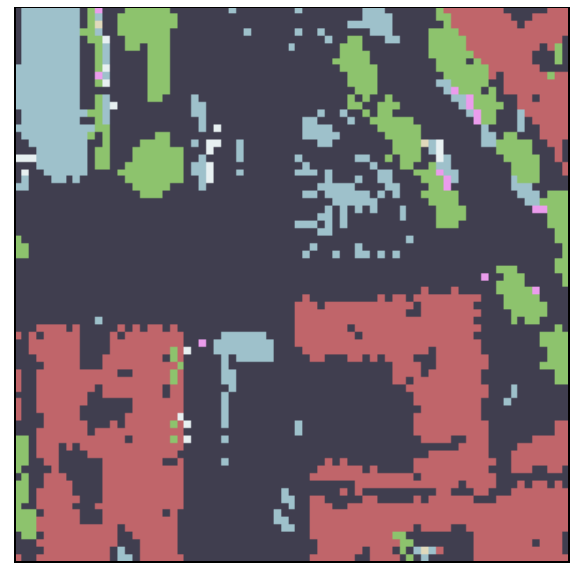}}
    
    \subfloat[]{\includegraphics[width=0.18\textwidth]{test_fib_img.pdf}}
    \hspace{0.05cm}
    \subfloat[]{\includegraphics[width=0.18\textwidth]{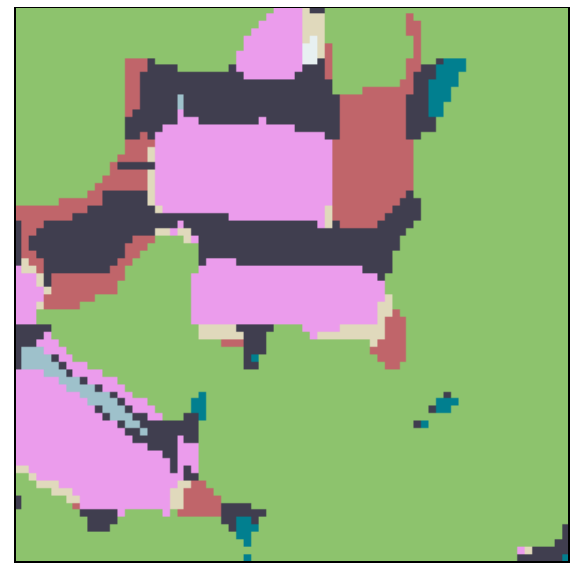}}
    \hspace{0.05cm}
    \subfloat[]{\includegraphics[width=0.18\textwidth]{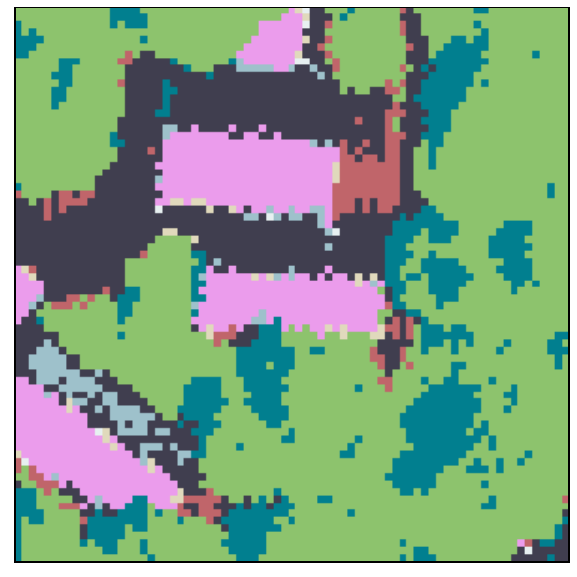}}
    \hspace{0.05cm}
    \subfloat[]{\includegraphics[width=0.18\textwidth]{gaussian_fib_test_pred.pdf}}
    \hspace{0.05cm}
    \subfloat[]{\includegraphics[width=0.18\textwidth]{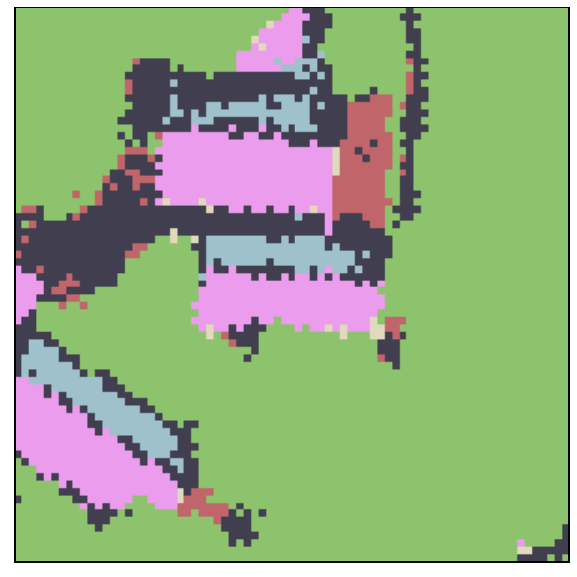}}
    
    \subfloat[]{\includegraphics[width=\textwidth]{legende.pdf}}

    \caption{(a)-(f)-(k) False color composition of subsets of the AI4GEO hyperspectral image, (b)-(g)-(l) the predictions of FG-UNET, (c)-(h)-(m) the predictions of ssInfoGAN, (d)-(i)-(n) the predictions of the gaussian VAE, (e)-(j)-(o) the predictions of the physics-guided VAE. FG-UNET predictions are spatially consistent but confuses asphalt in the shadows with water, as ssInfoGAN.  \label{fig:land_cover_maps_other}}
\end{figure*}

Fig. \ref{fig:land_cover_maps_other} shows additional land cover maps, demonstrating the benefits of p$^3$VAE against competing methods.   

\section{Methane plume inversion \label{sec:appendix_methane}}

In this section, we explain how we simulated the methane plume, and derived the physical prior model $f_E$.
We follow \cite{nesme2021joint} and approximate the ground-level radiance $R(\lambda)$, at wavelength $\lambda$, as follows:
\begin{equation}
	R(\lambda) = \frac{1}{\pi} r(\lambda) I_{{tot}}(\lambda) \tau_{atm}(\lambda) \label{eq:radiance_methane_eq} 
\end{equation}
where $r(\lambda)$ is the ground-level reflectance, $I_{{tot}}(\lambda)$ is the total solar irradiance and $\tau_{atm}(\lambda)$ is the atmospheric transmission. The transmission of methane through the plume $\tau_{CH_4}$ is defined as follows:
\begin{equation}
	\tau_{CH_4}(\lambda) = \mbox{exp} \big(- \rho_{CH_4} A_{CH_4}(\lambda) (1 + \frac{1}{\mbox{cos } \Theta})\big)
\end{equation} 
where $ \rho_{CH_4}$ is the methane path-integrated concentration trough the plume (ppm.m), $A_{CH_4}$ is the methane monochromatic absorption (ppm$^{-1}$.m$^{-1}$), and $\Theta$ is the solar zenith angle. The atmospheric transmission in the presence of the plume becomes $\tau_{atm}(\lambda) \tau_{CH_4}(\lambda)$. Therefore, given PRISMA reflectance data, irradiance data, and methane plume concentrations, we used eq. \ref{eq:radiance_methane_eq} to simulate the methane plume.
Besides, we also defined the prior physical knowledge $f_E : [0, 1]^B \times \mathbb{R}_+ \longmapsto \mathbb{R}_+^B$ from eq. \ref{eq:radiance_methane_eq}:
\begin{align*}
f_E(\x, z_E) = \x\mbox{exp} \big(- z_E A_{CH_4} (1 + \frac{1}{\mbox{cos } \Theta})\big)
\end{align*}
where $\x$ is meant to be an approximation of the $\frac{r I_{tot} \tau_{atm}}{\pi}$, computed by the neural network $f_I^{\param}$. \\

\noindent
The latent variable $\z_I$, encoding the land cover, is modeled with Gaussian distributions. The latent variable $z_E$, encoding the methane concentration, is modeled by Beta distributions, and scaled between the minimum and maximum admissible concentration values.

\section{Quantitative disentanglement evaluation \label{sec:appendix_mig}}

%Assessing whether a representation is disentangled is still an active research topic. Nevertheless, \cite{DBLP:journals/corr/abs-2012-09276} made a review of state-of-the art metrics to measure disentanglement, that they decompose in modularity, compactness and explicitness. \textbf{Modularity} guarantees that a variation in one factor only affects a subspace of the latent space, and that this subspace is only affected by one factor. \textbf{Compactness} relates to the size of the subspace affected by the variation in one factor. Finally, \textbf{explicitness} expresses how explicit is the relation between the latent code and the factors.
 
A latent space is said to be disentangled if its variables independently capture true underlying factors that explain the data \cite{DBLP:journals/corr/abs-2012-09276}. In this paper, we reported the Mutual Information Gap (MIG) introduced in \cite{NEURIPS2018_1ee3dfcd}. The MIG is computed based on a data set of factors (\textit{e.g.} categorical classes, local irradiance conditions, etc.) and latent codes (the latent representation inferred by the generative models). In the following, we denote $\vv$ as the factors and $\z$ as the latent code. MIG estimates the mutual information between factors and latent codes. Intuitively, the mutual information $I(\vv,\z)$ between $\vv$ and $\z$ indicates how much information we have about $\z$ when we know $\vv$. Therefore, the mutual information between a factor $\vv_i$ (\textit{e.g.} $\delta_{dir} cos \: \Theta$) and its related subspace $\z_\mathcal{K}$ (\textit{e.g.} $z_P$) should be high while the mutual information between $\vv_i$ and other subspaces should be low. MIG computes the mutual information between each code and factor $I(\vv_i, \z_j)$ and retains the two latent variables $\z_\ast$ and $\z_\circ$ with the highest mutual information $I(\vv_i, \z_\ast)$ and $I(\vv_i, \z_\circ)$. The difference between $I(\vv_i, \z_\ast)$ and $I(\vv_i, \z_\circ)$ is the mutual information gap. It reflects how much information related to the factor $\vv_i$ is expressed by $\z_\ast$ only. The gap is normalized by the entropy of the factor: 
\begin{equation}
    \mbox{MIG} := \frac{I(v_i, z_\ast) - I(v_i, z_\circ)}{H(\vv_i)}
\end{equation}
The higher the MIG, the better.

\begin{table}[h]
\begin{center}
    \caption{Architecture of the CNN \label{tab:cnn_arch}}
    \begin{tabular}{|p{7cm}|}
    \hline 
        \textbf{CNN}  \\ \hline \hline 
        \textit{CNN feature extractor} \\ \hline \hline 
        Input $1 \times n_{bands}$ \\ \hline 
        For each continuous spectral segment of length $n$: \\
        $2 \times ((\floor{n / 5})$ 1D conv - ReLU) \\ \hline 
        Skip connection \\ \hline 
        $(1 \times 2)$ max-pooling \\ \hline \hline 
        \textit{CNN dense classifier} \\ \hline \hline 
        2 $\times$ (FC 256 hidden neurons - ReLU) \\ \hline 
        FC output layer \\
        \hline 
    \end{tabular}
\end{center}
\end{table}

\begin{table} [h!]
	\begin{center}
    \caption{VAE-like encoders and decoders architectures \label{tab:VAE_arch}}
    \begin{tabular}{|p{7cm}|}
    \hline 
        \textbf{Gaussian VAE encoder} \\ \hline 
        Input $1 \times n_{bands}$ \\ \hline 
        $(1 \times 11)$ 1D conv - ReLU \\ \hline 
        $(1 \times 9)$ 1D conv - ReLU \\ \hline 
        $(1 \times 7)$ 1D conv - ReLU \\ \hline 
        $(1 \times 5)$ 1D conv - ReLU \\ \hline 
        Concatenation of the class $y$ of size $n_{classes}$ \\ \hline 
        2 $\times$ (FC 256 hidden neurons - ReLU) \\ \hline 
        FC variational posterior approximation output\\ \hline 
    \end{tabular}
    
    \vspace{0.2cm}
    
    \begin{tabular}{|p{7cm}|}
    \hline 
        \textbf{Gaussian VAE decoder} \\ \hline 
        Input $1 \times (dim(z) + n_{classes})$ \\ \hline 
        2 $\times$ (FC 256 hidden neurons - ReLU) \\ \hline 
        FC likelihood output \\ \hline 
    \end{tabular}
    
    \vspace{0.2cm}
    
    \begin{tabular}{|p{7cm}|}
    \hline 
        \textbf{Physics encoder} \\ \hline \hline 
        \textit{$z_P$ branch} \\ \hline \hline 
        Input $1 \times (n_{bands} + n_{classes})$ \\ \hline 
        2 $\times$ (FC 256 hidden neurons - ReLU) \\ \hline 
        2 $\times$ (FC 256 hidden neurons - ReLU) \\ \hline 
        FC latent posterior output \\ \hline \hline 
        \textit{$z_A$ branch} \\ \hline \hline 
        Input $1 \times n_{bands}$ \\ \hline 
        $(1 \times 11)$ 1D conv - ReLU \\ \hline 
        $(1 \times 9)$ 1D conv - ReLU \\ \hline 
        $(1 \times 7)$ 1D conv - ReLU \\ \hline 
        $(1 \times 5)$ 1D conv - ReLU \\ \hline 
        Concatenation of the class $y$ of size $n_{classes}$ \\ \hline 
        2 $\times$ (FC 256 hidden neurons - ReLU) \\ \hline 
        FC latent posterior output \\ \hline 
    \end{tabular}
    
    \vspace{0.2cm}
    
    \begin{tabular}{|p{3cm}|p{3cm}|}
        \hline 
        \multicolumn{2}{|p{7cm}|}{\textbf{Physics-guided VAE decoder}} \\ \hline 
        \multicolumn{2}{|p{7cm}|}{Input $1 \times n_{classes}$} \\ \hline 
        FC 256 hidden neurons - ReLU & FC 256 hidden neurons - ReLU \\ \hline 
        FC $n_{bands}$ hidden neurons - Sigmoid & FC $4 \times n_{bands}$ hidden neurons - Sigmoid \\ \hline 
        \multicolumn{2}{|p{7cm}|}{Element-wise multiplication}\\ \hline 
        \multicolumn{2}{|p{7cm}|}{Concatenation of $z_P$} \\ \hline 
        \multicolumn{2}{|p{7cm}|}{2 $\times$ (FC 256 hidden neurons - ReLU)} \\ \hline 
        \multicolumn{2}{|p{7cm}|}{FC likelihood output} \\ \hline 
    \end{tabular}
    
    \vspace{0.2cm}
    
    \begin{tabular}{|p{3cm}|p{3cm}|}
        \hline 
        \multicolumn{2}{|p{7cm}|}{\textbf{p$^3$VAE decoder}} \\ \hline 
        \multicolumn{2}{|p{7cm}|}{Input $1 \times n_{classes}$} \\ \hline 
        FC 256 hidden neurons - ReLU & FC 256 hidden neurons - ReLU \\ \hline 
        FC $n_{bands}$ hidden neurons - Sigmoid & FC $4 \times n_{bands}$ hidden neurons - Sigmoid \\ \hline 
        \multicolumn{2}{|p{7cm}|}{Element-wise multiplication}\\ \hline 
        \multicolumn{2}{|p{7cm}|}{Concatenation of $z_P$} \\ \hline 
        \multicolumn{2}{|p{7cm}|}{Evaluation of $f_P$} \\ \hline 
    \end{tabular}
    \end{center}
\end{table}

\begin{table}[h!]
	\begin{center}
    \caption{ssInfoGAN architecture \label{tab:info_arch}}
    \begin{tabular}{|p{7cm}|}
        \hline 
        \textbf{GAN discriminator head} \\ \hline 
        Input $1 \times n_{bands}$ \\ \hline 
        3 $\times$ (FC 256 hidden neurons - ReLU) \\ \hline 
        \textbf{GAN discriminator} \\ \hline
        Input $1 \times 256$ \\ \hline 
        Concatenate the smooth feature \\ \hline 
        FC 256 hidden neurons \\ \hline 
    \end{tabular}
    
    \vspace{0.2cm}
    
    \begin{tabular}{|p{3cm}|p{3cm}|}
        \hline 
        \multicolumn{2}{|p{7cm}|}{\textbf{GAN recognition network}} \\ \hline 
        Input $1 \times 256$ \\ \hline  
        \multicolumn{2}{|p{7cm}|}{FC 256 hidden neurons - ReLU} \\ \hline 
        FC continuous posterior & CNN dense classifier \\ \hline 
    \end{tabular}
    
    \vspace{0.2cm}
    
    \begin{tabular}{|p{7cm}|}
        \hline 
        \textbf{GAN generator} \\ \hline 
        Input $1 \times (dim(z) + dim(z_P) + dim(z_A) + n_{classes})$ \\ \hline 
        3 $\times$ (FC 256 hidden neurons - ReLU) \\ \hline 
        FC output layer - Sigmoid \\ \hline 
    \end{tabular}
    \end{center}
\end{table}

\begin{table}[h!]
	\begin{center}
    \caption{FG-UNET architecture }\label{tab:nowork}
    \begin{tabular}{|p{7cm}|}
        \hline 
        \textbf{Encoder} \\ \hline 
        Input $32 \times 32 \times n_{bands}$ \\ \hline 
        2 $\times$ ($(3 \times 3 \times 7)$ 3D conv - stride $(1 \times 1 \times 3)$ - ReLU) \\ \hline
        $\rightarrow$ Features 1 \\ \hline 
        $(2 \times 2 \times 1)$ max-pooling \\ \hline 
        2 $\times (3 \times 3 \times 7)$ 3D conv - stride 1 - ReLU) \\ \hline 
        $\rightarrow$ Features 2 \\ \hline 
        $(2 \times 2 \times 2)$ max-pooling \\ \hline 
        ($(3 \times 3 \times 7)$ 3D conv - stride 1 - ReLU) \\ \hline
    \end{tabular}
    
    \vspace{0.2cm}
    
    \begin{tabular}{|p{7cm}|}
        \hline 
        \textbf{Decoder} \\ \hline 
        Input $4 \times 4 \times 4$ \\ \hline
        $(3 \times 3 \times 7)$ 3D transposed conv - ReLU \\ \hline
        $(2 \times 2 \times 2)$ up-sampling \\ \hline 
        Concatenation of features 2 \\ \hline 
        $2 \times ((3 \times 3 \times 7)$ 3D transposed conv - ReLU) \\ \hline
        $(2 \times 2 \times 1)$ up-sampling \\ \hline
        Concatenation of features 1 \\ \hline \hline
        $\rightarrow$ Features 3 \\ \hline 
        \textit{Unlabeled branch} \\ \hline \hline
        Input: Features 3 \\ \hline 
        $2 \times ((3 \times 3 \times 7)$ 3D transposed conv - stride $(1 \times 1 \times 3)$ - ReLU) \\ \hline \hline 
        \textit{Labeled branch} \\ \hline \hline
        Input: Features 3 \\ \hline
        $(1 \times 1 \times 32)$ 3D conv - ReLU \\ \hline 
        Concatenation with pixel-wise output \\ \hline 
        $(1 \times 1)$ 2D conv - Softmax \\ \hline 
    \end{tabular}
    
    \vspace{0.2cm}
    
    \begin{tabular}{|p{7cm}|}
        \hline 
        \textbf{Pixel-wise fully connected path} \\ \hline 
        Input $32 \times 32 \times n_{bands}$ \\ \hline 
        $(1 \times 1 \times n_{bands})$ 3D conv - ReLU \\ \hline
        $2 \times ((1 \times 1)$ 2D conv - ReLU) \\ \hline
    \end{tabular}
    \end{center}
\end{table}

\end{appendices}

\end{document}